\documentclass{article}

\usepackage[main, final]{neurips_2026}
\usepackage{booktabs}
\usepackage{graphicx}
\usepackage{pifont}
\usepackage{xcolor}

\usepackage[T1]{fontenc}    
\usepackage{nicefrac}       
\usepackage{microtype}      

\definecolor{tableStrong}{HTML}{B7CAE8}
\definecolor{tableMid}{HTML}{E0E9F6}
\definecolor{tableNeutral}{HTML}{F3F4F6}
\definecolor{tableRow}{HTML}{EEF2F9}

\usepackage{float}
\usepackage[dvipsnames]{xcolor}
\usepackage{array}
\usepackage{colortbl}
\usepackage{graphicx}
\usepackage{epigraph}
\usepackage{subcaption}
\usepackage[autolanguage]{numprint}
\usepackage[utf8]{inputenc}
\usepackage{csquotes}
\usepackage{soul}
\usepackage{amssymb}
\usepackage{booktabs}
\usepackage{amsmath}
\usepackage{multirow}
\usepackage{siunitx}
\usepackage{xcolor}
\usepackage{graphicx} 
\usepackage{algorithm}
\usepackage{algpseudocode}
\usepackage{amsmath}
\usepackage{listings}
\usepackage{overpic}
\usepackage{dashrule}
\usepackage{enumitem} 
\definecolor{cgreen}{RGB}{35,140,70}
\definecolor{corange}{RGB}{210,120,20}
\definecolor{cgray}{RGB}{120,120,120}

\newcommand{\cmark}{\ding{51}}
\newcommand{\pmark}{\(\triangle\)}
\newcommand{\xmark}{\textendash}

\usepackage{tikz}
\definecolor{cvprblue}{rgb}{0.21,0.49,0.74}
\definecolor{yellow}{rgb}{1, 1, 0.7}
\definecolor{orange}{rgb}{1, 0.85, 0.7}
\definecolor{tablered}{rgb}{1, 0.7, 0.7}
\definecolor{red}{rgb}{1, 0, 0}
\definecolor{wincolor}{rgb}{0.85, 0.0, 0.0}
\definecolor{darkyellow}{rgb}{0.8, 0.8, 0.5}
\definecolor{darkred}{rgb}{0.7, 0.3, 0.3}
\definecolor{darkgreen}{rgb}{0.3, 0.7, 0.3}
\definecolor{green}{rgb}{0, 1.0, 0}
\definecolor{pink}{rgb}{1, 0.4, 0.7}
\definecolor{MyCustomBlue}{HTML}{150CA6}
\usepackage[
    pagebackref,
    breaklinks,
    colorlinks=true,
    citecolor=MyCustomBlue,
    pdftitle={GenSfM},
]{hyperref}
\usepackage{xspace}

\newcommand{\ours}{\textbf{\textsc{GenSfM}}\xspace}

\newcommand{\Nview}{N}                          
\newcommand{\numpts}{L}                         

\newcommand{\object}{\mathcal{O}}               
\newcommand{\image}{\mathbf{I}}                 
\newcommand{\imageset}{\mathcal{I}}             

\newcommand{\projmat}{\mathbf{P}}               
\newcommand{\poseset}{\mathcal{P}}              
\newcommand{\camK}{\mathbf{K}}                  
\newcommand{\camR}{\mathbf{R}}                  
\newcommand{\camt}{\mathbf{t}}                  

\newcommand{\point}{\mathbf{x}}                 
\newcommand{\pixcoord}{\mathbf{u}}              

\newcommand{\Slat}{\mathcal{S}}                 
\newcommand{\feat}{\mathbf{f}}                  

\newcommand{\timestep}{t}
\newcommand{\latentt}{\mathbf{z}_{\timestep}}
\newcommand{\latentzero}{\mathbf{z}_{0}}
\newcommand{\latentss}{\mathbf{z}^{s}}          
\newcommand{\latentpose}{\mathbf{z}^{p}}        
\newcommand{\latentreg}{\mathbf{z}^{r}}         
\newcommand{\latentcond}{\mathbf{z}^{I}}        
\newcommand{\condfeat}{\mathbf{c}}              
\newcommand{\sem}{\mathrm{sem}}                 
\newcommand{\tex}{\mathrm{tex}}                 
\newcommand{\noise}{\boldsymbol{\epsilon}}
\newcommand{\velocity}{\mathbf{v}}
\newcommand{\parameter}{\phi}                   

\newcommand{\slatenc}{\operatorname{Enc}_{\mathrm{L}}}

\newcommand{\trellis}{{\textsc{TRELLIS}}\xspace}
\newcommand{\cupid}{{\textsc{CUPID}}\xspace}
\newcommand{\dino}{\textsc{DINOv2}\xspace}
\newcommand{\interp}{\operatorname{BilinearInterp}}

\usepackage{amsmath,amsfonts,bm}

\def\Secref#1{Sec.~\ref{#1}}
\def\Appref#1{App.~\ref{#1}}
\def\Figref#1{Fig.~\ref{#1}}
\def\Eqref#1{Eq.~\ref{#1}}

\def\eqref#1{equation~\ref{#1}}

\def\1{\bm{1}}

\DeclareMathAlphabet{\mathsfit}{\encodingdefault}{\sfdefault}{m}{sl}
\SetMathAlphabet{\mathsfit}{bold}{\encodingdefault}{\sfdefault}{bx}{n}

\DeclareMathOperator*{\argmin}{arg\,min}

\title{
Generative Structure from Motion \\ 
with Native 3D Diffusion
}

\author{%
  \textbf{Haobin Duan}$^{1,2}$\thanks{Equal contribution.}\quad
  \textbf{Binbin Huang}$^{1,2}$\footnotemark[1]\quad
  \textbf{Mochu Xiang}$^{1,2}$\quad
  \textbf{Yiqun Zhao}$^{1,2}$ \\
  \textbf{Zibo Zhao}$^{3}$\quad
  \textbf{Yi Ma}$^{1,2}$\quad
  \textbf{Shenghua Gao}$^{1,2}$\thanks{Corresponding author.} \\[0.4em]
  $^{1}$The University of Hong Kong\quad
  $^{2}$Transengram\quad
  $^{3}$Tencent
}
\hypersetup{pdfauthor={Haobin Duan, Binbin Huang, Mochu Xiang, Yiqun Zhao, Zibo Zhao, Yi Ma, Shenghua Gao}}

\begin{document}

\maketitle

\begin{figure*}[h]
    \centering
    \includegraphics[width=\textwidth]{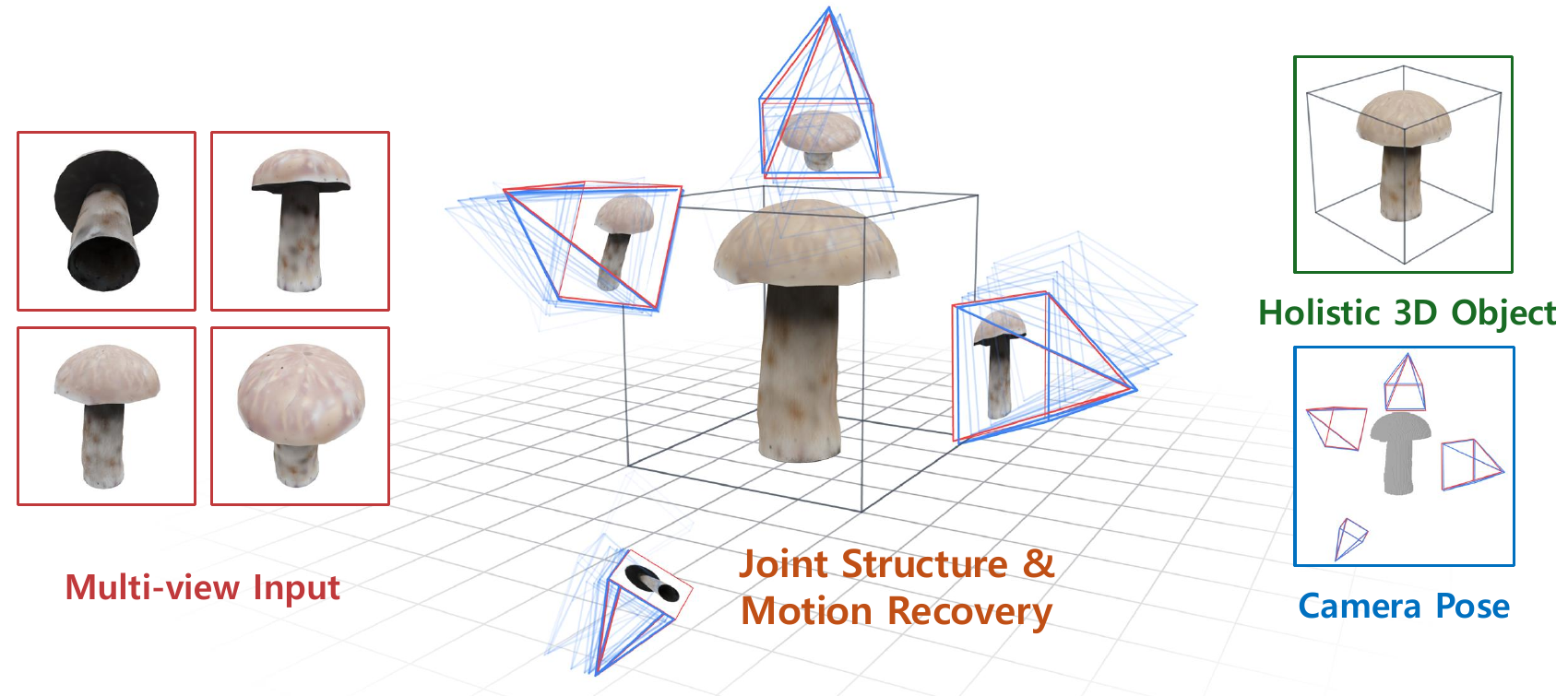}
    \caption{
    \textbf{Generative object shape and camera motion recovery.} Given multiview input images, \ours{} recovers the object shape and camera poses in a unified 3D space. A diffusion transformer performs multi-view reasoning through iterative structure-and-pose denoising (the trajectory of the camera), emulating an end-to-end SfM process. The denoised cameras ({\color{blue}blue}) align accurately with the ground-truth poses ({\color{red}red}) and the recovered 3D structure gives a holistic object representation.
    }
    \label{fig:teaser}
\end{figure*}
\vspace{3mm}

\begin{abstract}

Recovering an object's 3D structure and camera motion from a few
uncalibrated views is a fundamental problem in computer vision. Classical structure from motion methods recover 3D geometry consistent with the input images but cannot recover unseen regions; native generation methods produce plausible 3D objects which are related but not necessarily faithful to the input images. Methods simply concatenating reconstruction and generation allow error to cascade across the two stages, instead of producing a 3D object consistent with both the distribution and the input. We propose \ours, which integrates image-based 3D reconstruction with a native 3D diffusion model. In particular, we use a multi-modal diffusion transformer to jointly denoise the shape latent and per-view pose latents under shared self-attention, grounding generated geometry
in the observed views while using object-level priors to infer
unobserved regions. A compact 3D-native UV-volume tokenization keeps
the joint sequence small as views are added, and a pose-aware stage refines geometry and texture by sampling per-voxel image features at the recovered camera poses. On Toys4k and GSO, for $4$ uncalibrated views, \ours improves
novel-view PSNR by $3.2$--$4.5$~dB and reduces median Chamfer distance
by $2.8$--$3.8\times$ over a strong native-3D generation baseline, while
recovering input-view poses over $96\%$ Acc@$5^\circ$ without any
external pose-estimation backbone, post-hoc optimization, or known intrinsics.
Project page: \url{https://dotb1ue.github.io/GenSfm/}.

\end{abstract}


%


\section{Introduction}
\label{sec:intro}

Recovering 3D structure and camera motion from images is a foundational problem in computer vision. Classical Structure-from-Motion and Multi-View Stereo pipelines~\cite{schoenberger2016sfm, schoenberger2016mvs, furukawa2009accurate} estimate camera poses from multi-view correspondences and reconstruct geometry through photometric or geometric consistency, often regularized by hand-crafted priors. While effective when input views provide sufficient texture, overlap, and coverage, this geometric paradigm becomes brittle in sparse object-centric settings, where only a few uncalibrated views are available and large object regions go unrecovered.

Such settings are common in practice, from robotic manipulation to AR capture and content-creation pipelines, all of which need a holistic 3D representation to reason about occluded surfaces and contact regions. Sparse-view 3D object recovery therefore requires more than triangulating visible evidence: it must infer a complete 3D explanation that is both consistent with the input images and plausible as an object, jointly answering \textbf{\emph{what}} the object is and \textbf{\emph{where}} each view was taken from.

Recent learning-based methods relax the texture and overlap assumptions of classical pipelines by introducing large-scale data-driven priors. Feed-forward reconstructors~\cite{wang2024dust3r, leroy2024grounding, wang2025vggt} predict camera parameters and pixel-aligned 3D representations jointly, while pose-centric works~\cite{zhang2022relpose, wang2023posediffusion, sinha2023sparsepose, zhang2024cameras} estimate relative camera motion under limited overlap and pose ambiguity. Both threads remain primarily observation-driven: they recover what the input views support but do not explicitly model the distribution of complete object shapes, leaving outputs incomplete in unobserved regions when input coverage is poor.

A complementary line of work introduces generative priors learned from large-scale 3D datasets~\cite{deitke2023objaverse} into sparse reconstruction. View-based generative SfM~\cite{zhao2025diffusionsfm, zhang2024world, xu2024sparp, zhao2024sparseags} casts structure and motion as a generative problem in view space, while reconstruct-then-generate pipelines~\cite{ardelean2025gen3dsr, huang2026unirecgen, chang2025reconviagen, sam3dteam2025sam3d3dfyimages, siddiqui2026shaper} pair feed-forward reconstructors with native 3D generators~\cite{TripoSR2024, xiang2025structured, zhao2025hunyuan3d}. The latter family in particular has shown that 3D generative priors can complete hidden geometry that observation-driven methods cannot. These systems remain cascaded: an external module predicts poses or image-aligned representations and the 3D generator only completes the result, propagating errors across stages and preventing joint optimization of a single 3D hypothesis that explains all input views.

In this paper, we invert this pipeline: instead of using a generative prior to complete the output of a feed-forward reconstructor, we treat the 3D generative prior itself as the primary reasoning mechanism for sparse-view reconstruction. We propose \ours, an end-to-end framework that formulates joint recovery of complete object structure and per-view camera motion as a single conditional inference problem in a native 3D diffusion model, as shown in Fig.~\ref{fig:teaser}. The diffusion prior operates directly on a native 3D representation throughout inference, simultaneously denoising shape and per-view pose under joint self-attention. This grounds the generated object in the observed views while using object-level regularities to infer unobserved geometry.

Realizing this formulation requires a single diffusion transformer that jointly denoises a shared 3D shape with a per-view pose latent bound to its input image, while keeping the per-view pose-token overhead small as views are added. We adopt \cupid's~\cite{huang2025cupid} 3D-native UV-volume pose tokenization, but compress each view's pose latent to a compact token grid, since the dense volume over-parameterizes a low-DoF perspective projection. A learned view embedding tags each view's pose latent and image tokens, binding pose to its source image. Since neither stream natively carries cross-view information, we prepend a small set of register tokens to the shape stream as a shared channel for all views to communicate through self-attention.

Instead of relying on back-projection to recover surface texture like feed-forward methods~\citep{wang2024dust3r}, we add a second stage of native 3D refinement and texturing~\citep{xiang2025structured} so that hidden regions stay coherent with the observed views. We design a pose-aware fine stage that, using the cameras predicted in the coarse stage, projects every active voxel into the input views and conditions denoising on per-voxel semantic and texture features sampled and averaged across views.

On the Toys4k~\cite{stojanov2021using} and GSO~\cite{downs2022google} sparse-view object benchmarks under uncalibrated 4-view inputs, \ours improves novel-view PSNR by 3.2--4.5~dB, reduces median Chamfer distance by $2.8$--$3.8\times$, and lifts $\mathrm{F@}.01$ from $0.56$ to $0.83$ on Toys4k over the strong native-3D generation baseline \trellis-M~\cite{xiang2025structured}, while recovering input-view camera poses to over $96\%$ Acc@$5^\circ$ (a $5\times$ rotation-error reduction over the strongest pose baseline VGGT~\cite{wang2025vggt}) without requiring known intrinsics.
Our contributions are summarized as follows:
\begin{itemize}[leftmargin=2em]
    \item We propose \ours, a joint native-3D architecture that recasts a diffusion model from a post-hoc completion module into the primary inference mechanism for joint sparse-view structure and motion recovery, eliminating any external feed-forward reconstruction backbone.
    \item A 3D-native multi-view pose tokenization that places per-view UV-volume pose latents and the shape latent in a single multi-modal sequence, jointly denoised by a diffusion transformer under shared self-attention.
    \item A pose-aware fine stage that, with cameras recovered from the coarse stage, anchors semantic and texture features at every active voxel and averages them across views, sharpening surface detail beyond global, pose-agnostic conditioning.
\end{itemize}

\section{Related Work}
\label{sec:related}
Sparse-view object reconstruction must jointly answer two coupled questions: \textbf{\emph{what}} is the 3D object, and \textbf{\emph{where}} was each view taken from? \textbf{\emph{What}} requires both visible and hidden object structure, while \textbf{\emph{where}} requires the camera motion relating the input views. Existing methods typically excel on one side or address the two with separate models stacked in series (Tab.~\ref{tab:comparison}).

\begin{table}[tb]
\centering
\renewcommand{\arraystretch}{1.12}
\setlength{\tabcolsep}{5pt}
\caption{
Comparison of paradigms for sparse-view object structure-and-motion recovery.
``Where'' denotes camera-motion recovery and ``what'' denotes object-structure recovery, with observed and hidden geometry referring to visible and unobserved object regions, respectively.
``Unified joint inference'' means jointly recovering complete object structure and camera motion in one model, while ``native 3D prior'' denotes a generative prior over complete 3D shapes.
\cmark: explicit support; \pmark: partial or indirect support; \xmark: not a primary objective.
}
\label{tab:comparison}
\colorlet{paradigmYes}{tableStrong}
\colorlet{paradigmMid}{tableMid}
\colorlet{paradigmNo}{tableNeutral}
\providecommand{\cYes}{\cellcolor{paradigmYes}\cmark}
\providecommand{\cMid}{\cellcolor{paradigmMid}\pmark}
\providecommand{\cNo}{\cellcolor{paradigmNo}\xmark}
\resizebox{\linewidth}{!}{
\begin{tabular}{lccccc}
\toprule
\textbf{Method family}
& \textbf{Where:}
& \multicolumn{2}{c}{\textbf{What: object structure}}
& \textbf{Unified joint}
& \textbf{Native} \\
\cmidrule(lr){3-4}
& \textbf{camera motion}
& \textbf{observed}
& \textbf{hidden}
& \textbf{inference}
& \textbf{3D prior} \\
\midrule
Classical SfM/MVS~\cite{schoenberger2016sfm, schoenberger2016mvs, furukawa2009accurate}
& \cYes & \cYes & \cNo  & \cMid & \cNo  \\
Pose-centric methods~\cite{zhang2022relpose, wang2023posediffusion, sinha2023sparsepose, zhang2024cameras}
& \cYes & \cNo  & \cNo  & \cNo  & \cNo  \\
Feed-forward reconstruction~\cite{wang2024dust3r, leroy2024grounding, wang2025vggt}
& \cYes & \cYes & \cMid & \cMid & \cNo  \\
Native 3D generation~\cite{TripoSR2024, xiang2025structured, zhao2025hunyuan3d}
& \cNo  & \cMid & \cYes & \cNo  & \cYes \\
View-based generative SfM~\cite{zhao2025diffusionsfm, zhang2024world, xu2024sparp}
& \cYes & \cMid & \cMid & \cMid & \cNo  \\
Recon. with 3D Gen. priors~\cite{chang2025reconviagen, huang2026unirecgen, sam3dteam2025sam3d3dfyimages}
& \cYes & \cYes & \cYes & \cMid & \cYes \\
\midrule
\ours (Ours)
& \cYes & \cYes & \cYes & \cYes & \cYes \\
\bottomrule
\end{tabular}
}
\end{table}

\paragraph{Feed-forward pose and 3D reconstruction.}
Pose-centric approaches~\cite{zhang2022relpose, wang2023posediffusion, sinha2023sparsepose, zhang2024cameras} target the \emph{where} side, regressing relative camera motion under limited overlap and pose ambiguity, but leave \emph{what} largely untouched. More general feed-forward reconstructors~\cite{wang2024dust3r, leroy2024grounding, wang2025vggt} predict dense pointmaps, correspondences, and camera parameters jointly, addressing camera motion and visible geometry; hidden geometry, however, is recovered only indirectly through learned priors.

\paragraph{Native 3D generation.}
Native 3D generators trained on large 3D corpora~\cite{deitke2023objaverse}, such as TripoSR~\cite{TripoSR2024}, \trellis~\cite{xiang2025structured}, and Hunyuan3D~\cite{zhao2025hunyuan3d}, are strong on \emph{what}: complete object structure is modeled explicitly and hidden geometry is hallucinated from the prior. They do not target \emph{where}: camera motion is typically assumed and only weakly encoded through cross-attention, so applied directly to uncalibrated sparse views they can produce plausible but misaligned geometry.

\paragraph{View-based generative SfM.}
View-based generative SfM~\cite{zhao2025diffusionsfm, zhang2024world, xu2024sparp, zhao2024sparseags} casts \emph{what} and \emph{where} as a joint generative problem in view space; DiffusionSfM, for instance, diffuses ray origins and endpoints to recover per-view geometry and camera motion jointly. Because diffusion operates over view-based quantities (rays, pointmaps, depths), hidden \emph{what} is recovered only insofar as it can be back-projected from individual views.

\paragraph{Reconstruction with 3D generative priors.}
Reconstruct-then-generate pipelines~\cite{ardelean2025gen3dsr, huang2026unirecgen, chang2025reconviagen, sam3dteam2025sam3d3dfyimages, siddiqui2026shaper, yao2025castcomponentaligned3dscene} couple feed-forward reconstruction with native 3D generators: e.g., ReconViaGen~\cite{chang2025reconviagen} injects VGGT~\cite{wang2025vggt} geometry tokens into a 3D generator. These pipelines recover hidden geometry through the generative prior while grounding camera motion and visible structure with feed-forward geometry, but solve \emph{what} and \emph{where} in separate stages, so errors from the reconstruction backbone propagate into a 3D hypothesis that need not remain coherent with all input views.

\section{Preliminaries}
\label{sec:method_prelim}

\paragraph{Overview.}
Given $\Nview$ uncalibrated multi-view images $\imageset = \{\image_i\}_{i=1}^{\Nview}$ of an object $\object$, our goal is to recover the holistic 3D structure and per-view camera poses $\poseset = \{\projmat_i\}_{i=1}^{\Nview}$ as a coarse-to-fine generative problem in a native 3D diffusion model. We build on two native 3D generators reviewed below: \trellis~\cite{xiang2025structured} provides the two-stage backbone and CFM objective we inherit, and \cupid~\cite{huang2025cupid} contributes the 3D-native single-image pose tokenization we lift to multi-view.

\paragraph{Coarse-to-fine 3D generation.}
Our backbone is \trellis~\cite{xiang2025structured}, which generates a 3D object in two coarse-to-fine stages under a shared conditional flow matching~\cite{lipman2022flow} (CFM) objective. The first stage produces a sparse-structure (SS) latent $\latentss$ on a $16^3$ grid, decoded to a $64^3$ occupancy grid whose active voxels define positions $\{\point_j\}_{j=1}^{\numpts}$. The second stage produces the structured-latent (SLat) set $\Slat = \{\point_j, \feat_j\}_{j=1}^{\numpts}$ with per-voxel features $\feat_j$, which separate decoders~\cite{shen2023flexible,kerbl20233d} map to 3D Gaussians, a radiance field, or a mesh. Both stages train with
\begin{equation}
    \mathcal{L}_{\mathrm{CFM}}(\parameter) = \mathbb{E}_{\timestep, \latentzero, \noise} \left\| \velocity_{\parameter}(\latentt, \image, \timestep) - (\noise - \latentzero) \right\|_2^2,
    \label{eq:cfm}
\end{equation}
where $\latentt = (1{-}\timestep)\latentzero + \timestep\,\noise$ is the noisy latent at $\timestep \in [0,1]$ for clean target $\latentzero$ ($\latentss$ for SS, $\{\feat_j\}$ for SLat) and noise $\noise \sim \mathcal{N}(\mathbf{0}, \mathbf{I})$; $\velocity_{\parameter}$ is the DiT velocity prediction conditioned on $\image$.

\paragraph{3D-native pose representation.}
Our pose tokenization derives from \cupid~\cite{huang2025cupid}, which extends \trellis's SS stage to jointly recover the camera pose for a single input image: the SS latent gains channels encoding per-voxel UV coordinates, decoded to a $64^3$ UV volume that gives each active voxel's projection $\pixcoord_j \in [0,1]^2$. The $3{\times}4$ projection matrix $\projmat$ is thus a dense 3D-2D field on the active voxels, recovered from the per-voxel correspondences $\{\point_j, \pixcoord_j\}_{j=1}^{\numpts}$ by Direct Linear Transform (DLT) least-squares~\citep{abdel2015direct}:
\begin{equation}
\projmat^{*}
= \argmin_{\projmat} \sum_{j=1}^{\numpts} \big\Vert\pi(\projmat, \point_j) - \pixcoord_j\big\Vert^2,
\label{eq:dlt}
\end{equation}
where $\pi$ is the perspective projection.
Finally, camera intrinsic $\camK$ and extrinsic $[\camR \mid \camt]$ are recovered from $\projmat^{*}$ via RQ decomposition.

\section{Method}
\label{sec:method}

\begin{figure*}[!t]
    \centering
    \begin{overpic}[width=\textwidth]{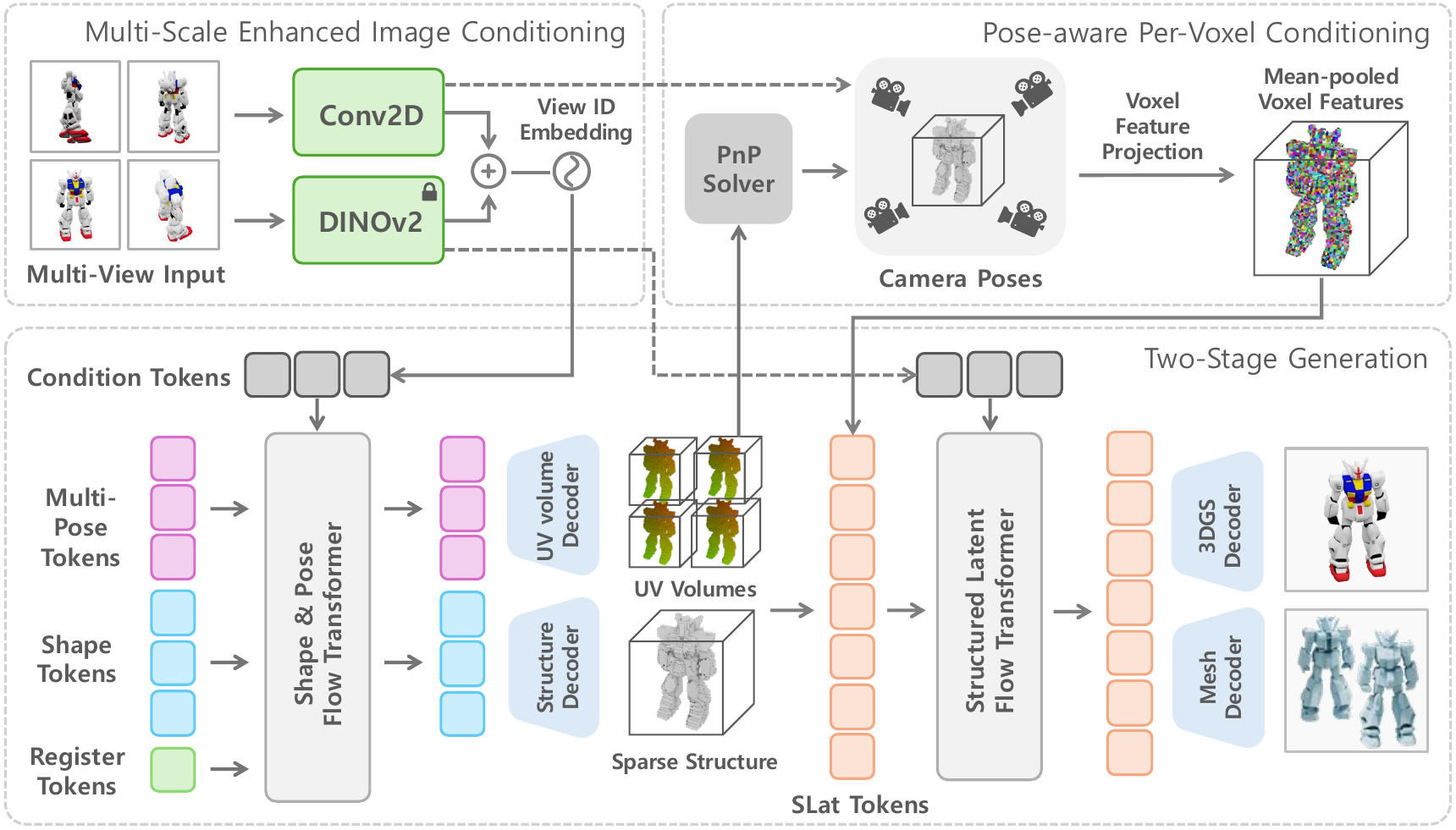}
        \put(48.9,80.2){%
            \begingroup
            \setlength{\fboxsep}{1pt}%
            \colorbox{gray!20}{\makebox[2.4em][c]{\textbf{DLT}}}%
            \endgroup
        }
    \end{overpic}
    \caption{Overview of \ours. We cast pose-free multi-view 3D recovery as a single coarse-to-fine generative problem. The coarse stage (\Secref{sec:method_coarse}) jointly denoises the sparse-structure latent and per-view 3D-native pose tokens with a multi-modal shape-pose DiT, recovering shape and multi-view camera poses together. The fine stage (\Secref{sec:method_fine}) refines geometry and appearance with per-voxel pose-sampled image conditioning, producing the final structured-latent set.}
    \label{fig:method_overview}
\end{figure*}

We frame this joint recovery as a coarse-to-fine generative problem in the latent space of a native 3D diffusion model (Fig.~\ref{fig:method_overview}). Building on \trellis's two-stage architecture and \cupid's 3D-native pose representation (\Secref{sec:method_prelim}), \Secref{sec:method_coarse} jointly generates 3D structure and multi-view poses with a multi-modal DiT over per-view pose tokens, and \Secref{sec:method_fine} refines geometry and appearance via per-voxel pose-sampled image conditioning.

\subsection{3D structure and pose recovery from multi-view images}
\label{sec:method_coarse}

The coarse stage casts joint recovery of 3D structure and per-view camera poses as a single denoising problem: each camera is encoded as a per-view 3D-native pose token, and shape and pose are denoised jointly by a multi-modal DiT with joint self-attention over both streams.

\paragraph{Tokenizing multi-view camera poses.}
For each view $i$, every active voxel $\point_j$ projects to $\pixcoord_{j,i} = \pi(\projmat_i, \point_j)$, yielding a UV volume $\pixcoord_i \in [0,1]^{2 \times 64^3}$. A shared pose VAE compresses each $\pixcoord_i$ into a latent grid $\latentpose_i \in \mathbb{R}^{c \times r \times r \times r}$ with $c{=}8, r{=}4$. To represent multiple input views, we form a unified token sequence with the shape latent $\latentss$ followed by the per-view pose latents $\{\latentpose_i\}_{i=1}^{\Nview}$ — in contrast to \cupid~\citep{huang2025cupid}, whose channel-wise concatenation of shape and pose into a single SS latent is restricted to one view. We choose $r{=}4$ (vs \cupid's $r{=}16$), giving $64\times$ fewer tokens per pose at negligible accuracy cost since the UV volume is heavily over-parameterized for a low-DoF projection. At inference, the pose VAE decodes each generated $\latentpose_i$ back to a UV volume, from which $\projmat_i^{*}$ is recovered by solving \Eqref{eq:dlt}. See \Appref{sec:suppl_pose_vae} for details.

\paragraph{Multi-view conditioned shape-pose DiT.}
A shared transformer backbone denoises shape and pose jointly, with stream-specific projections and grid-aligned positional embeddings. Learnable register tokens $\latentreg$ are prepended to the shape stream to capture global context. We extend the CFM interpolation of \Eqref{eq:cfm} to a shape timestep $\timestep^{s} \in [0,1]$ and per-view pose timesteps $\mathbf{t}^{p} \in [0,1]^{\Nview}$, giving $\latentzero = [\latentss;\, \latentpose]$ and $\latentt = [\latentt^{s};\, \latentt^{p}]$. Stream-specific in-projections then lift each noisy latent to the shared transformer width, where positional and view embeddings are added in to form the layer-$0$ hidden state of the DiT:
\begin{equation}
    \mathbf{h}_{0} = [\,\latentreg;\;\; W_{\mathrm{in}}^{s}\,\latentt^{s} + \mathbf{e}^{s};\;\; W_{\mathrm{in}}^{p}\,\latentt^{p} + \mathbf{e}^{p} + \mathbf{e}^{v}\,],
    \label{eq:dit_input}
\end{equation}
where $\mathbf{e}^{s}, \mathbf{e}^{p}$ are absolute 3D PEs over the $16^3$ and $4^3$ grids that anchor each stream to its native voxel layout, and $\mathbf{e}^{v}$ is a per-view ID embedding shared between each pose latent and its source view's image-condition tokens (\Eqref{eq:image_cond}), binding each pose to its view through cross-attention. We then stack $L$ DiT blocks; each block injects the multi-view condition tokens $\latentcond = \{\latentcond_i\}_{i=1}^{\Nview}$,
\begin{equation}
    \mathbf{h}_{l+1} = \operatorname{CrossAttn}\!\big(Q(\mathbf{h}_{l}),\, K(\latentcond),\, V(\latentcond)\big),
\end{equation}
where $Q, K, V$ are the query/key/value projections, applied jointly over all $\Nview$ views without per-view masking. Each block also runs a joint self-attention over the entire sequence and a feed-forward sub-layer, both modulated by per-modality AdaLN values conditioned on $\timestep^{s}$ for shape tokens and on $\timestep^{p}_{i}$ for view-$i$ pose tokens. After the final block, stream-specific out-projections read out the velocities (register outputs are discarded), $\velocity_{\parameter}(\latentt, \latentcond, \timestep^{s}, \mathbf{t}^{p}) = [\,W_{\mathrm{out}}^{s}\,\mathbf{h}_{L}^{s};\; W_{\mathrm{out}}^{p}\,\mathbf{h}_{L}^{p}\,]$, and the whole network is trained with a stream-wise conditional flow matching loss:
\begin{equation}
    \mathcal{L}_{\mathrm{coarse}}(\parameter) = \mathbb{E}_{\timestep^{s},\, \mathbf{t}^{p},\, \latentzero,\, \noise}\, \big\| \velocity_{\parameter}(\latentt, \latentcond, \timestep^{s}, \mathbf{t}^{p}) - (\noise - \latentzero) \big\|_{2}^{2}.
    \label{eq:coarse_cfm}
\end{equation}

\paragraph{Multi-scale enhanced image condition.}
\dino patch features capture rich semantics but are spatially coarse, providing limited cues for the fine appearance and accurate 2D-3D correspondences our coarse stage must recover. We therefore augment them with multi-scale local features extracted by a small convolutional pyramid on the same image: for each view $i$, the condition tokens are
\begin{equation}
    \latentcond_i = \big(\dino(\image_i) \,\oplus\, \Phi^{s}_{\mathrm{cnn}}(\image_i)\big) + \mathbf{e}^{v},
    \label{eq:image_cond}
\end{equation}
where $\dino(\image_i)$ are the frozen patch tokens, $\Phi^{s}_{\mathrm{cnn}}(\image_i)$ stacks features from a 4-level convolutional pyramid where each level patchified to the \dino patch grid and channel-concatenated across levels, $\oplus$ denotes channel-wise concatenation per token, and $\mathbf{e}^{v}$ is the view-tag embedding from \Eqref{eq:dit_input}. The resulting $\latentcond = \{\latentcond_i\}_{i=1}^{\Nview}$ serves as the cross-attention key/value sequence in every DiT block. See \Appref{sec:suppl_image_cond} for details.

\subsection{Pose-aware generation of fine geometry and appearance}
\label{sec:method_fine}

The fine stage generates geometry and appearance with pose-aware per-voxel image conditioning: each voxel receives a semantic feature from \dino and a texture feature from a multi-scale CNN pyramid, with both features average-pooled over pose-aligned projections across views.

\paragraph{Pose-aware per-voxel image conditioning.}
\trellis's SLat stage conditions on the input image only through global cross-attention with \dino patch tokens, leaving fine-appearance cues unanchored from any specific voxel and producing blurry surface detail. Once $\projmat^{*}$ is recovered by the coarse stage (\Secref{sec:method_coarse}), every active voxel $\point_j$ has a known image projection $\pixcoord_j = \pi(\projmat^{*}, \point_j)$, so image features can be sampled directly per voxel. At each $\point_j$ we form two complementary features: a semantic feature $\condfeat^{\sem}_j$ from pose-sampled \dino patches re-encoded through the frozen SLat encoder $\slatenc$ (placing it on the same latent manifold as the SLat target $\feat_j$), and a texture feature $\condfeat^{\tex}_j$ from a multi-scale CNN pyramid $\Phi^{l}_{\mathrm{cnn}}$ that supplies the low-level cues that \dino lacks:
\begin{equation}
\begin{aligned}
\{\condfeat^{\sem}_j\}_{j=1}^{\numpts} &= \slatenc\big(\{\point_j,\, \interp(\pixcoord_{j},\, \dino(\image))\}_{j=1}^{\numpts}\big), \\
\condfeat^{\tex}_j &= \interp\big(\pixcoord_{j},\, \Phi^{l}_{\mathrm{cnn}}(\image)\big).
\end{aligned}
\label{eq:slat_pose_cond}
\end{equation}
where $\interp$ denotes bilinear interpolation. We channel-wise concatenate $\condfeat^{\sem}_j$ and $\condfeat^{\tex}_j$ with the noisy voxel feature $\feat_j^{\timestep}$ (the SLat target $\feat_j$ noised via \Eqref{eq:cfm}) and linearly project to form the per-voxel layer-$0$ DiT token $\mathbf{h}_0(\point_j) = W^{l}_{\mathrm{in}}[\,\feat_j^{\timestep} \,\oplus\, \condfeat^{\sem}_j \,\oplus\, \condfeat^{\tex}_j\,]$; the remainder of the DiT follows \trellis. See \Appref{sec:suppl_image_cond} for the visual pyramid architecture.

\paragraph{Pooling per-voxel features across views.}
With $\Nview$ input views, every active voxel has $\Nview$ pose-aligned projections $\pixcoord_{j,i} = \pi(\projmat_i^{*}, \point_j)$. We extend $\interp$ to per-view sampling followed by averaging, $\interp(\pixcoord_j,\, \Phi(\imageset)) := \tfrac{1}{\Nview}\sum_{i=1}^{\Nview} \interp(\pixcoord_{j,i},\, \Phi(\image_i))$, so \Eqref{eq:slat_pose_cond} applies unchanged with $\image$ replaced by $\imageset$. The cross-attention condition is likewise the per-view mean of \dino patch tokens. The mean operator itself is parameter-free and view-count-agnostic; the model in this work is trained with $\Nview{=}4$.

\section{Experiments}
\label{sec:experiments}

\begin{table}[t]
\renewcommand{\arraystretch}{1.08}
\setlength{\tabcolsep}{3pt}
\centering
\caption{
\textbf{Comparison with feed-forward reconstruction methods.}
We align predictions to the ground truth via a similarity transform fit by least squares between predicted and ground-truth input-view pointmaps, then evaluate photometric quality on novel views and 3D geometric accuracy in the ground-truth coordinate frame. CD is bidirectional symmetric (squared L2, $\times 10^{-3}$), reported as the median across objects.
}
\label{tab:main_recon}
\resizebox{\linewidth}{!}{
\begin{tabular}{lcccccccccccc}
\toprule
\multirow{2}{*}{\textbf{Method}} &
\multicolumn{6}{c}{\textbf{Toys4k}} &
\multicolumn{6}{c}{\textbf{GSO}} \\
\cmidrule(lr){2-7}\cmidrule(lr){8-13}
&
\textbf{PSNR}$\uparrow$ & \textbf{SSIM}$\uparrow$ & \textbf{LPIPS}$\downarrow$ & \textbf{CD-med}$\downarrow$ & \textbf{F@.01}$\uparrow$ & \textbf{F@.05}$\uparrow$ &
\textbf{PSNR}$\uparrow$ & \textbf{SSIM}$\uparrow$ & \textbf{LPIPS}$\downarrow$ & \textbf{CD-med}$\downarrow$ & \textbf{F@.01}$\uparrow$ & \textbf{F@.05}$\uparrow$ \\
\midrule
FreeSplatter~\cite{xu2024freesplatter} &
22.00 & 0.9183 & 0.0913 & 2.7160 & 0.3163 & 0.7112 &
20.07 & 0.8948 & 0.1135 & 3.2859 & 0.2713 & 0.6736 \\
LucidFusion~\cite{he2025lucidfusion} &
{22.24} & {0.9251} & 0.0827 & 4.1596 & 0.1527 & 0.6670 &
{20.94} & {0.9097} & 0.0943 & 3.8702 & 0.1755 & 0.7164 \\
ReconViaGen~\cite{chang2025reconviagen} &
21.88 & 0.9216 & {0.0721} & {1.1237} & {0.3369} & {0.8434} &
20.88 & 0.9021 & {0.0856} & {1.0495} & {0.3524} & {0.8789} \\
\rowcolor{tableRow}
\ours &
{28.61} & {0.9609} & {0.0308} & {0.2184} & {0.6701} & 
{0.9823} &
{26.36} & {0.9411} & {0.0472} & {0.2801} & {0.6095} & {0.9842} \\
\bottomrule
\end{tabular}
}
\\[2pt]
\end{table}
\newlength{\comprimsize}
\setlength{\comprimsize}{2.25cm}

\newcommand{\comprbp}{{\color{black}\rule{\comprimsize}{\comprimsize}}}
\newcommand{\comprhdr}[1]{\small #1}

\begin{figure*}[!t]
\centering
\setlength{\tabcolsep}{1pt}

\begin{tabular}{*{6}{>{\centering\arraybackslash}m{\comprimsize}}}
\comprhdr{Input} & \comprhdr{FreeSplatter} & \comprhdr{LucidFusion} & \comprhdr{ReconViaGen} & \comprhdr{\textbf{Ours}} & \comprhdr{GT} \\[2pt]
\includegraphics[width=\comprimsize,height=\comprimsize]{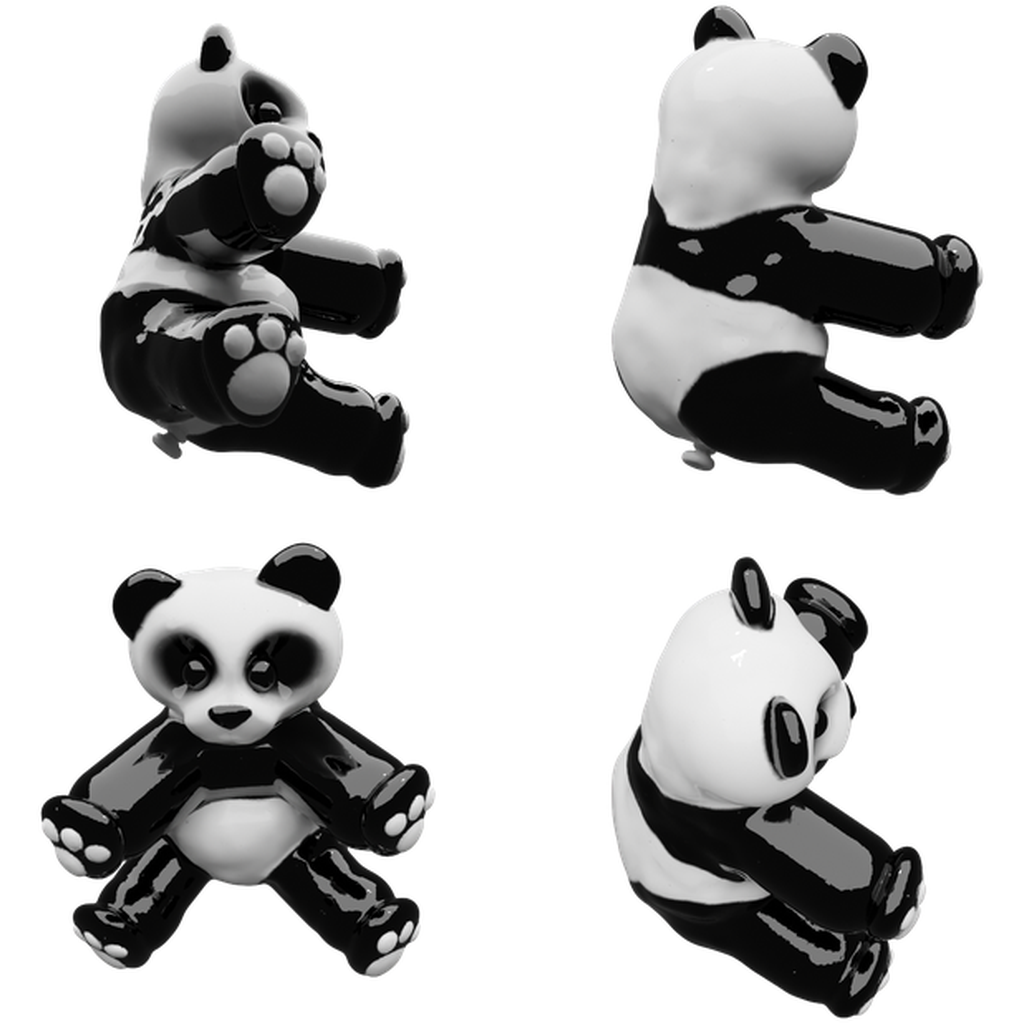} & \includegraphics[width=\comprimsize,height=\comprimsize]{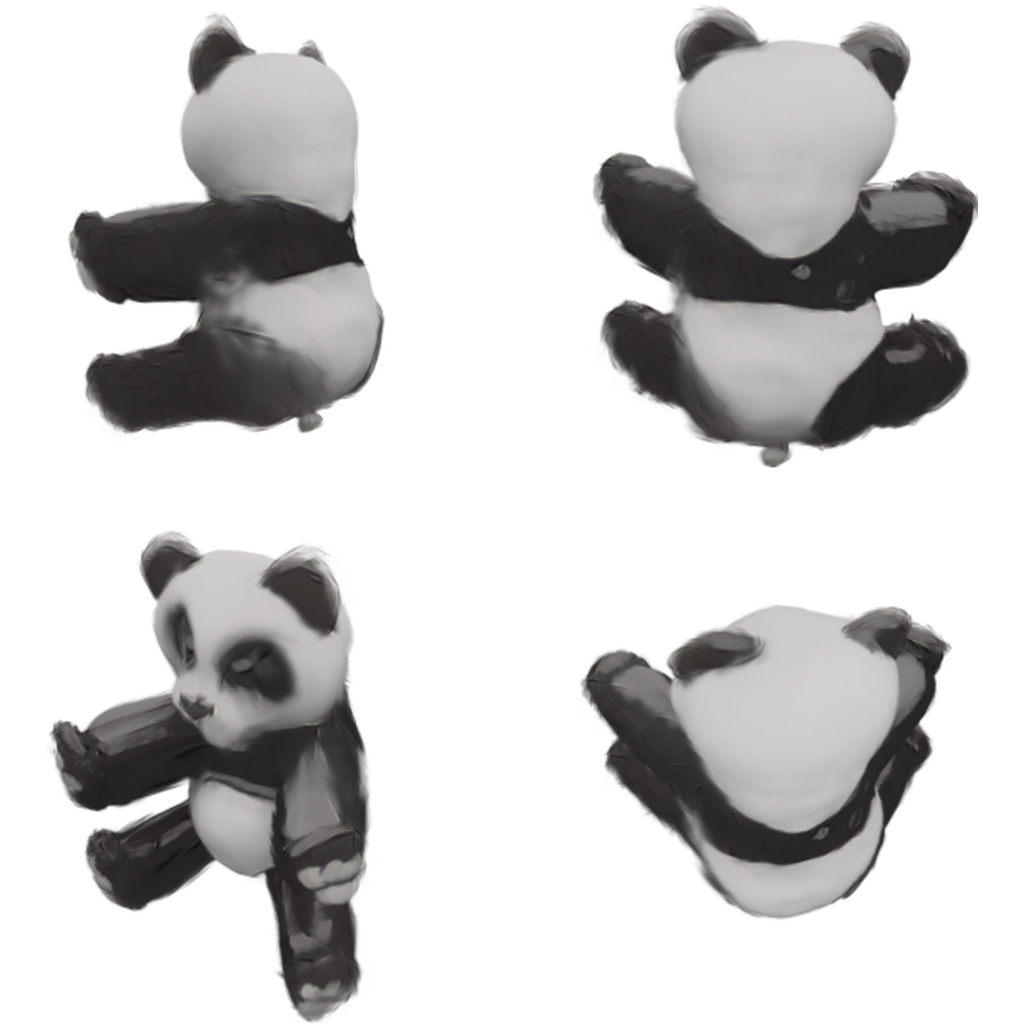} & \includegraphics[width=\comprimsize,height=\comprimsize]{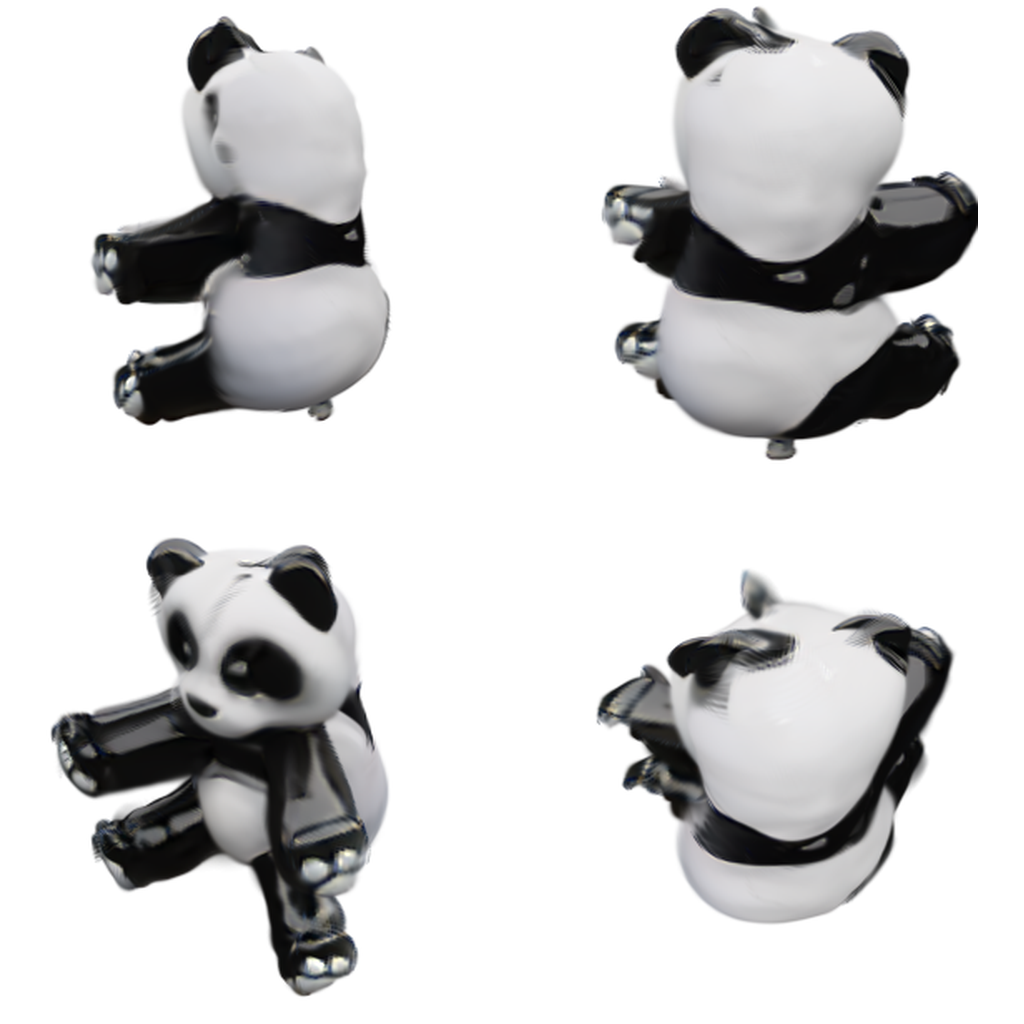} & \includegraphics[width=\comprimsize,height=\comprimsize]{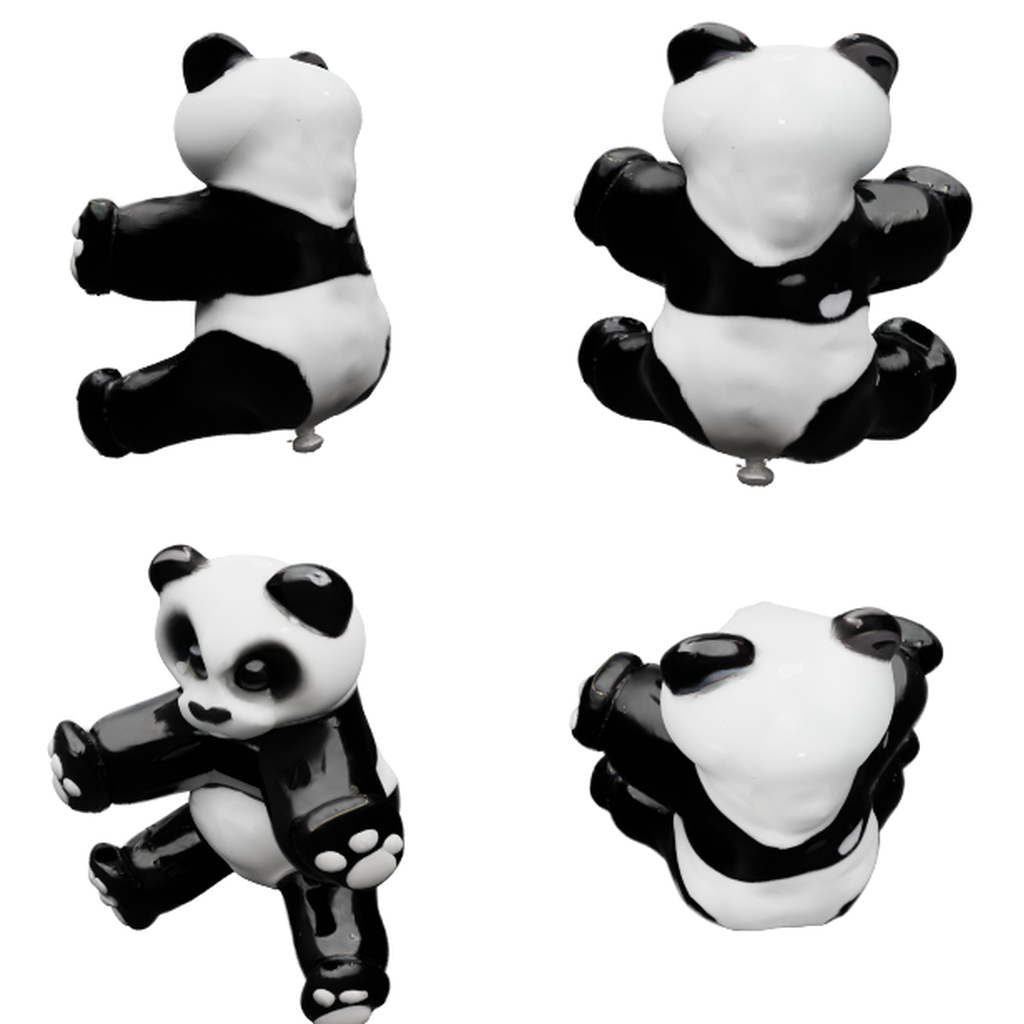} & \includegraphics[width=\comprimsize,height=\comprimsize]{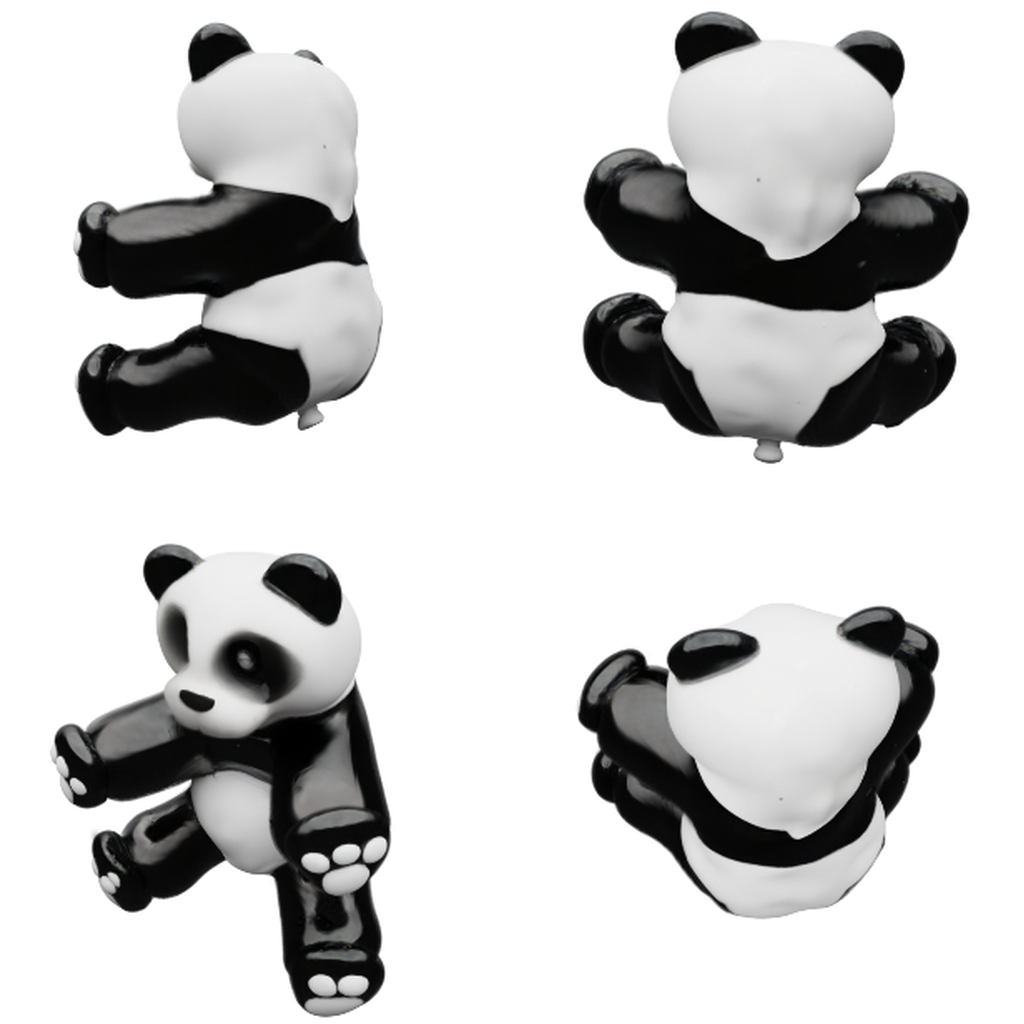} & \includegraphics[width=\comprimsize,height=\comprimsize]{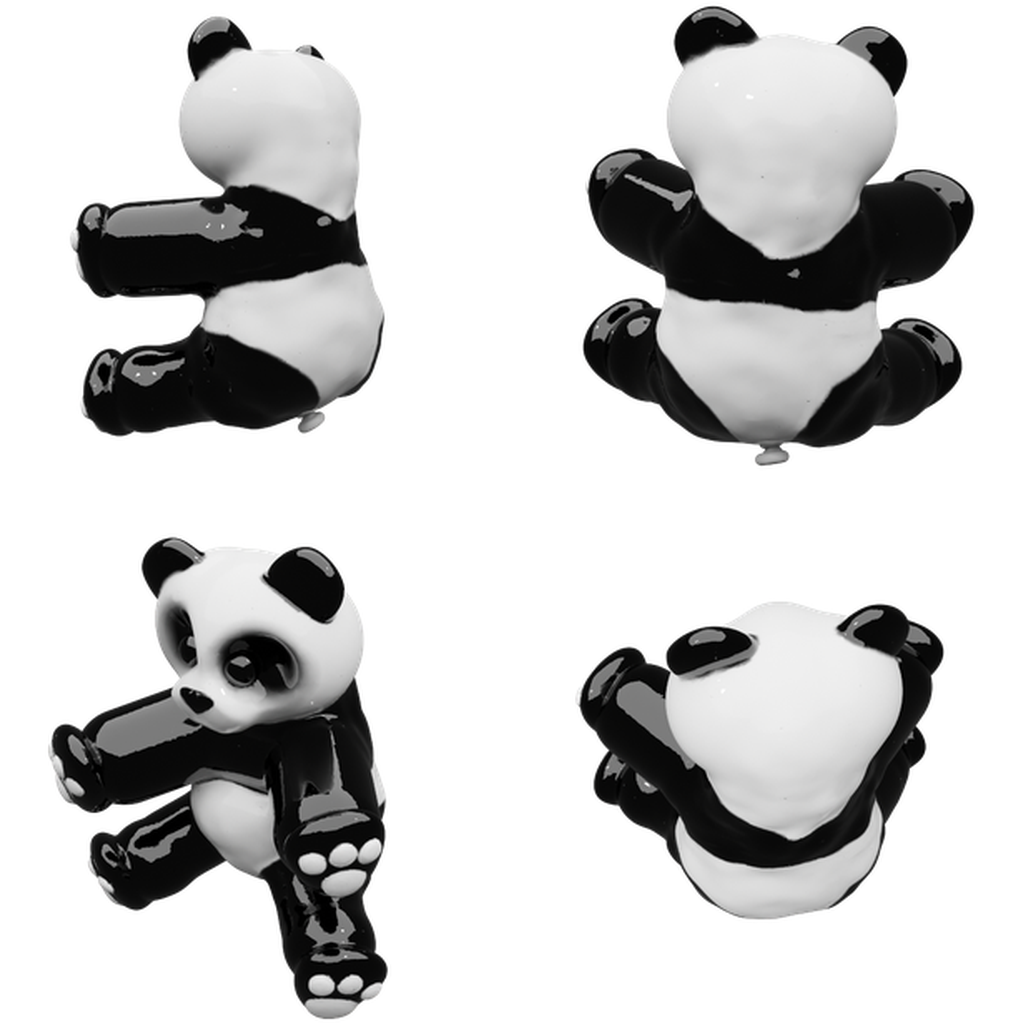} \\
\includegraphics[width=\comprimsize,height=\comprimsize]{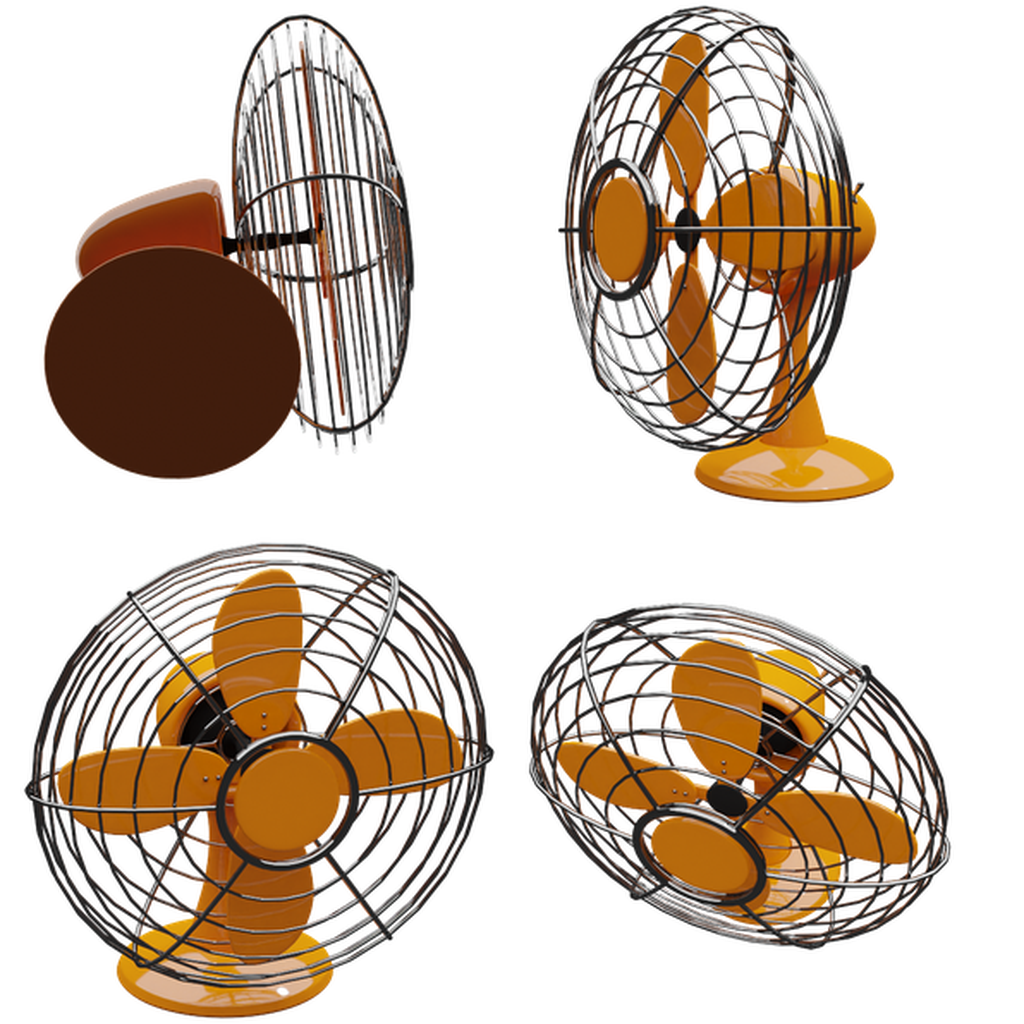} & \includegraphics[width=\comprimsize,height=\comprimsize]{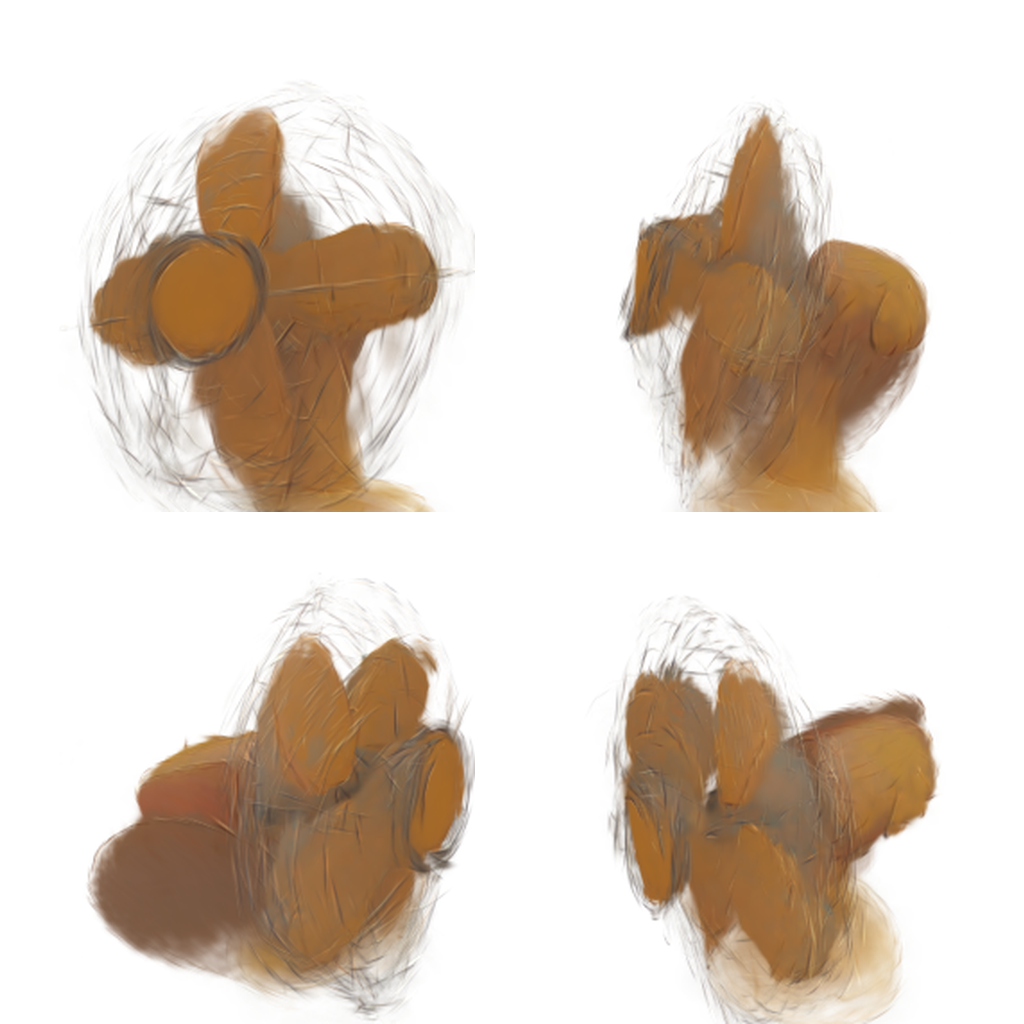} & \includegraphics[width=\comprimsize,height=\comprimsize]{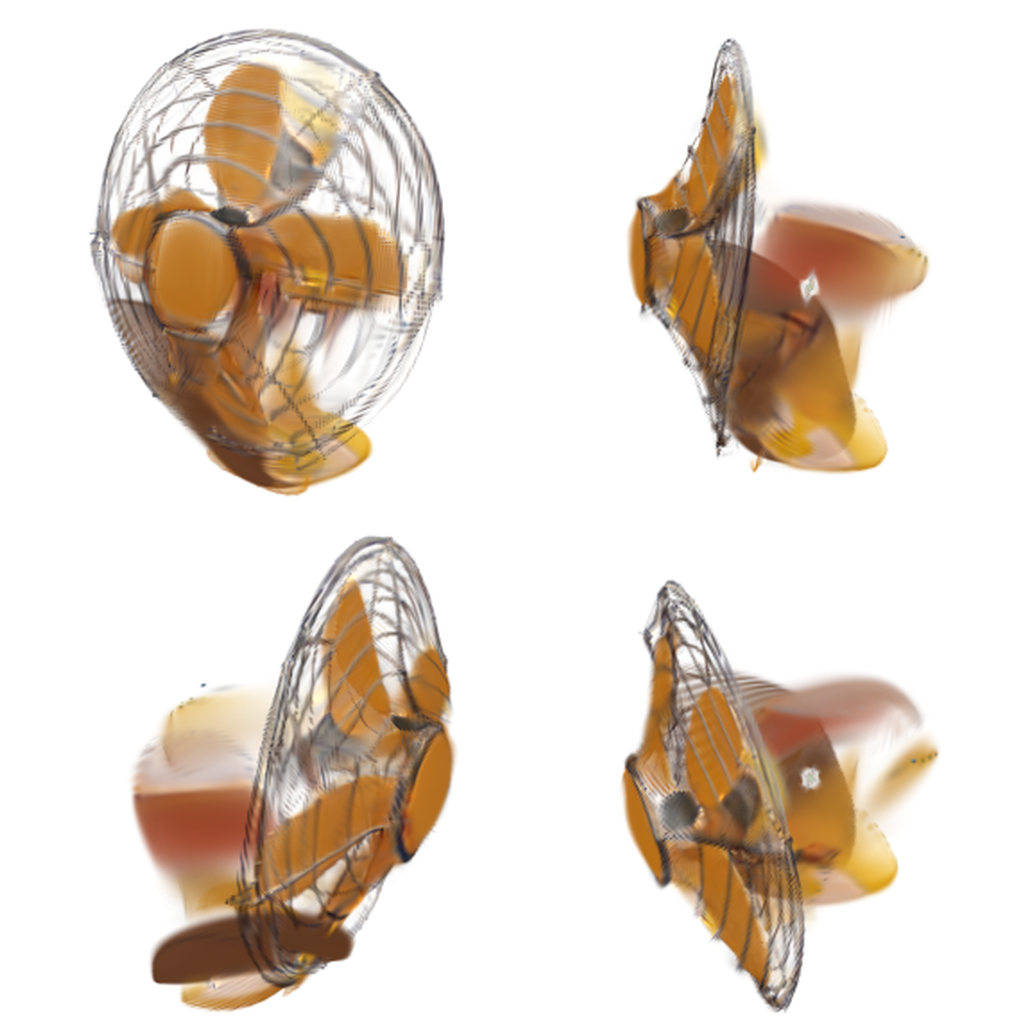} & \includegraphics[width=\comprimsize,height=\comprimsize]{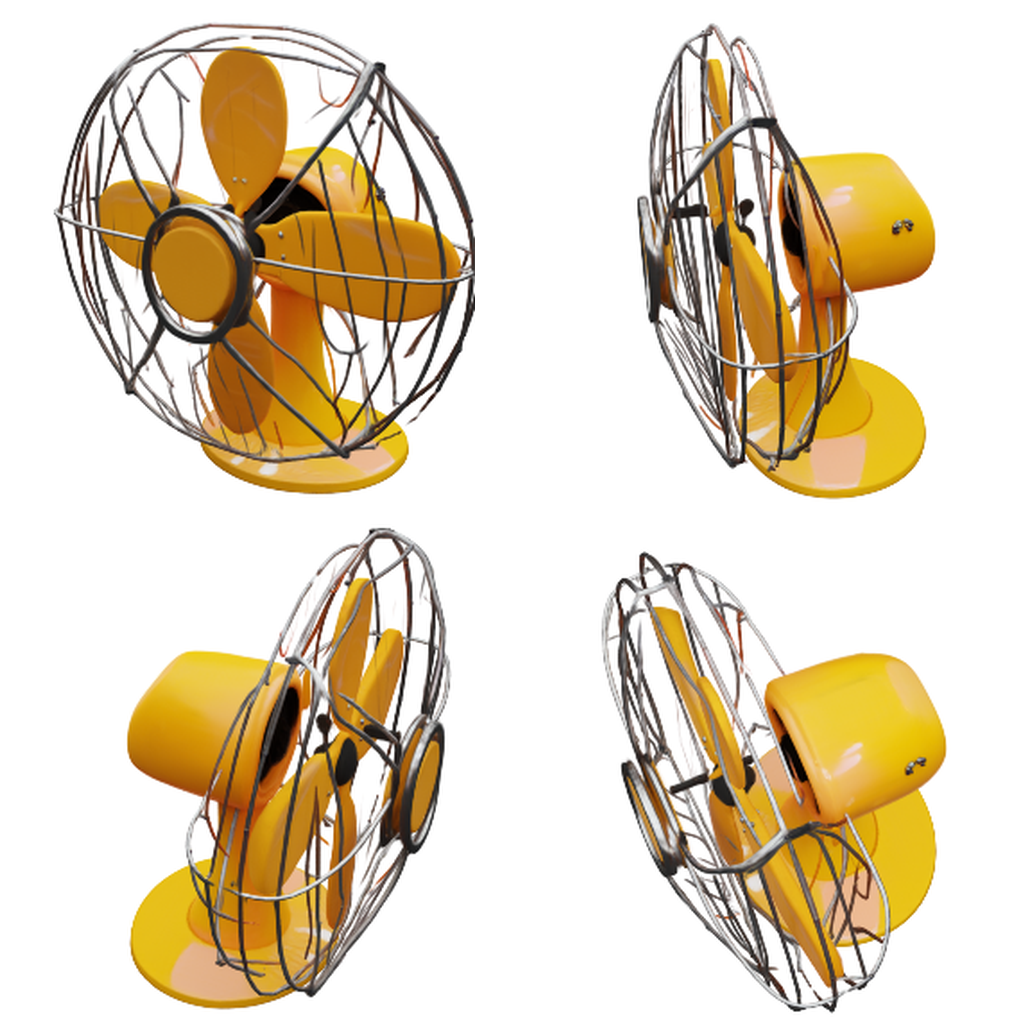} & \includegraphics[width=\comprimsize,height=\comprimsize]{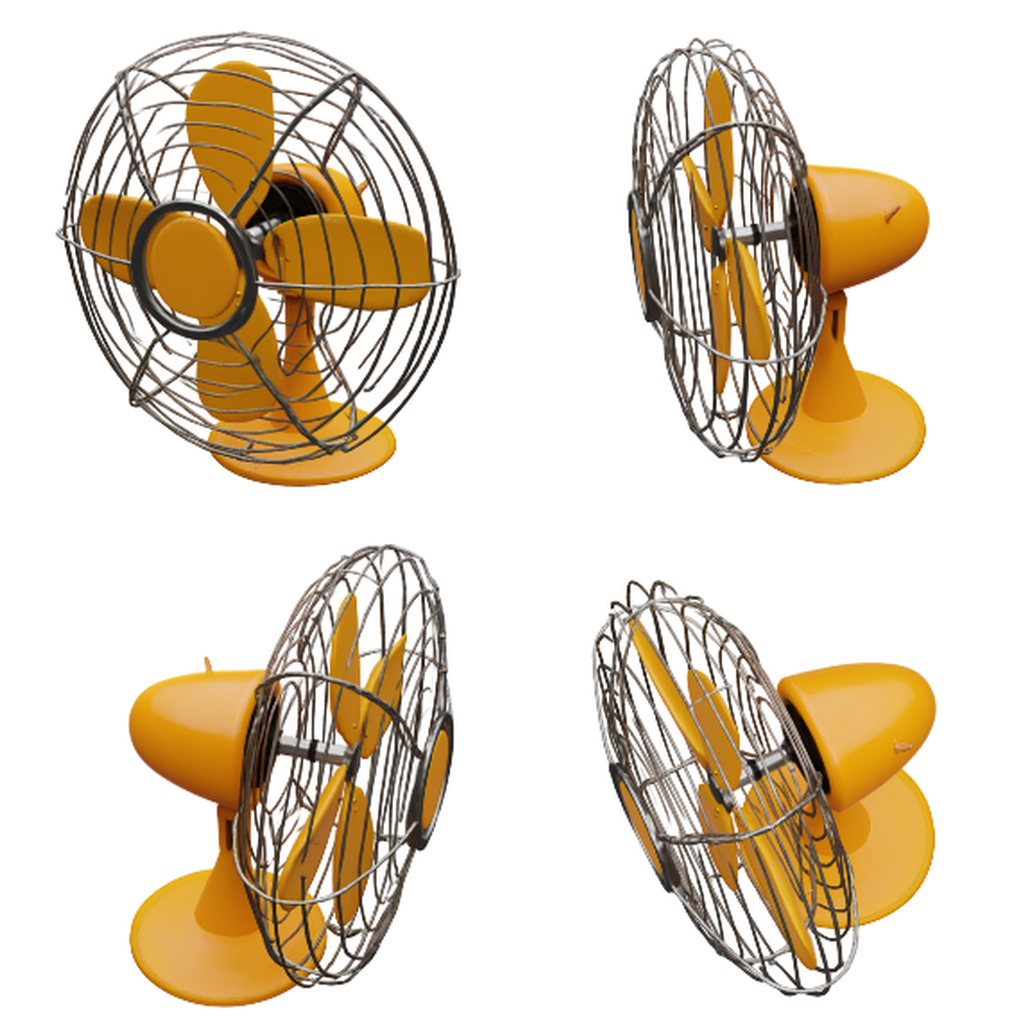} & \includegraphics[width=\comprimsize,height=\comprimsize]{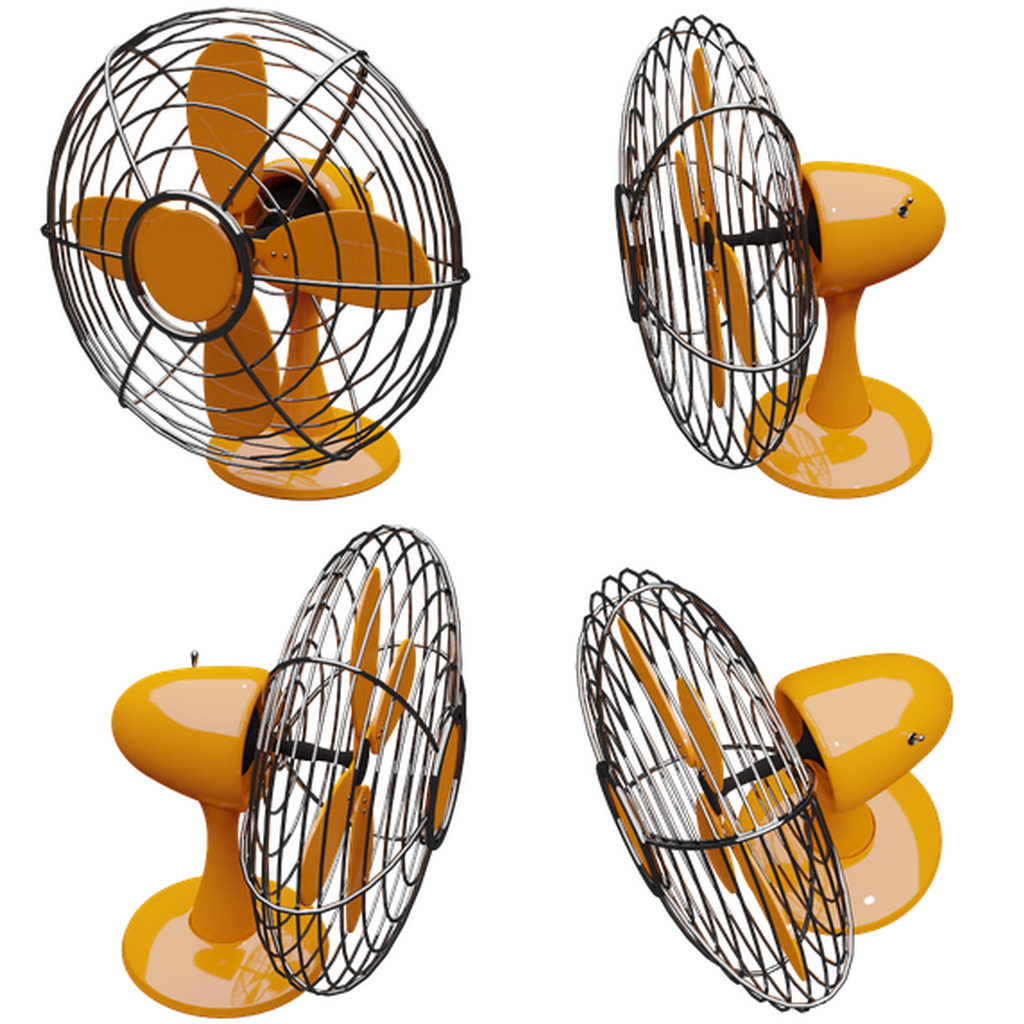} \\
\includegraphics[width=\comprimsize,height=\comprimsize]{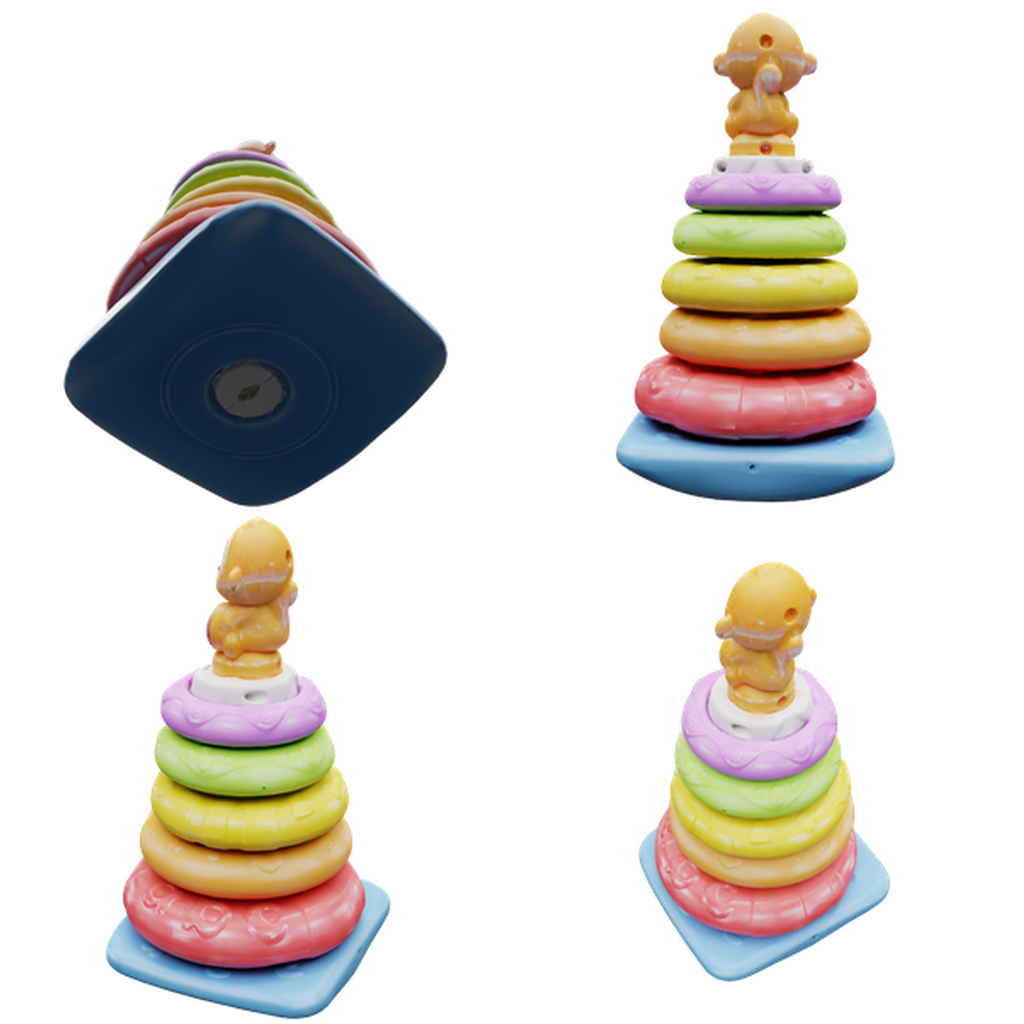} & \includegraphics[width=\comprimsize,height=\comprimsize]{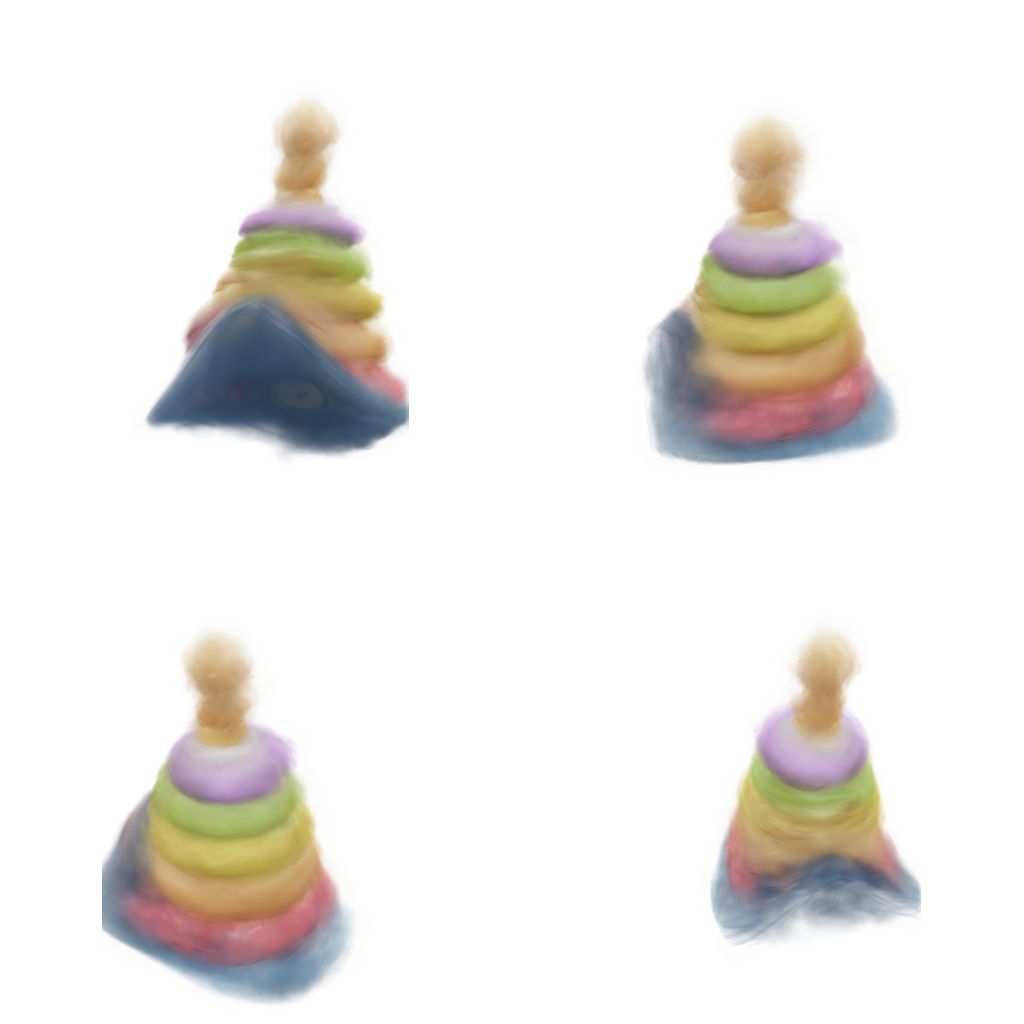} & \includegraphics[width=\comprimsize,height=\comprimsize]{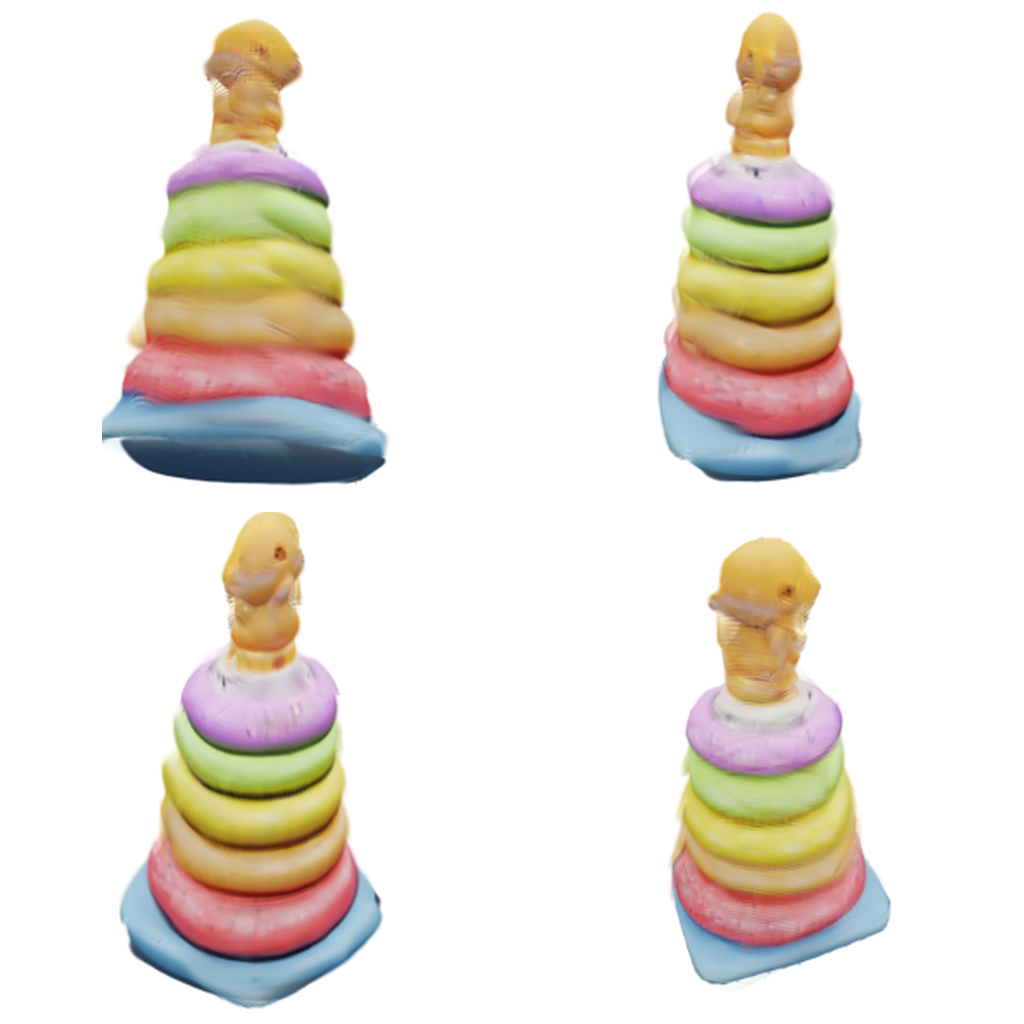} & \includegraphics[width=\comprimsize,height=\comprimsize]{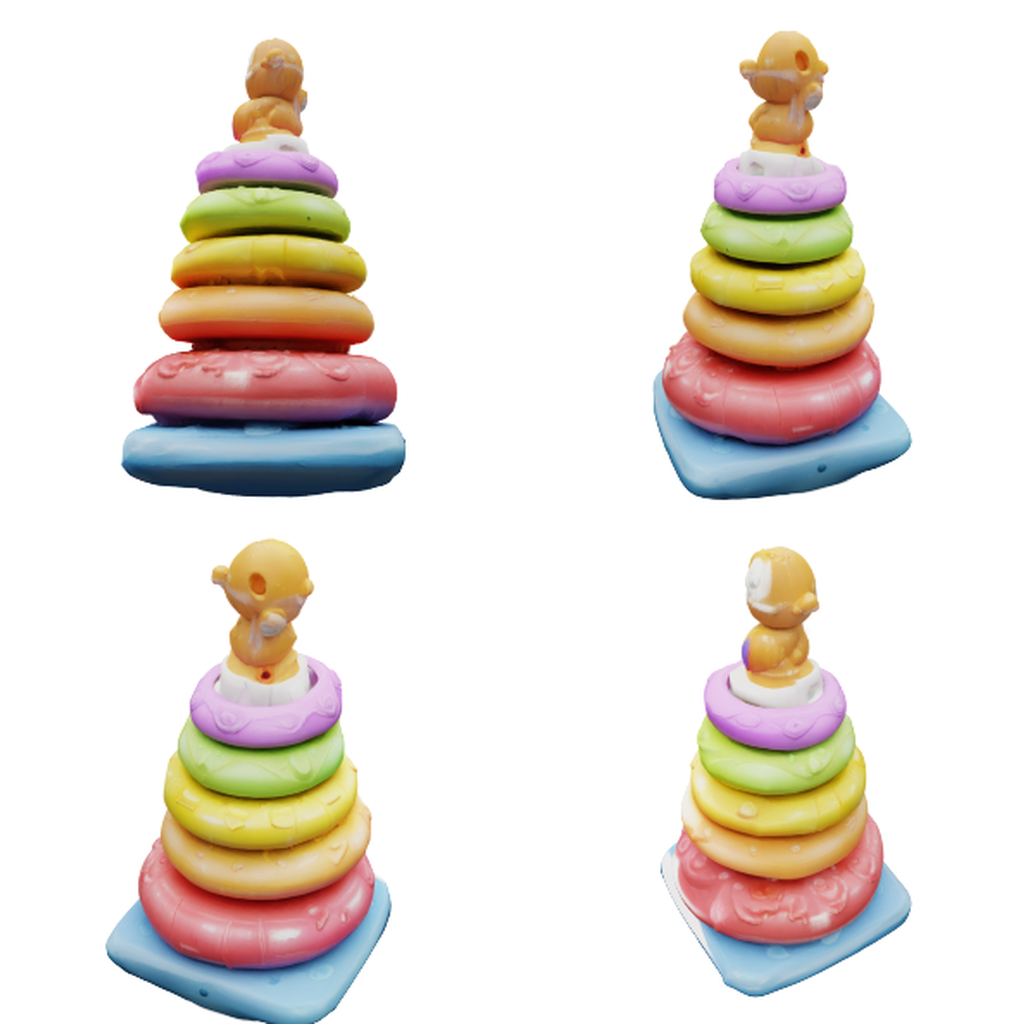} & \includegraphics[width=\comprimsize,height=\comprimsize]{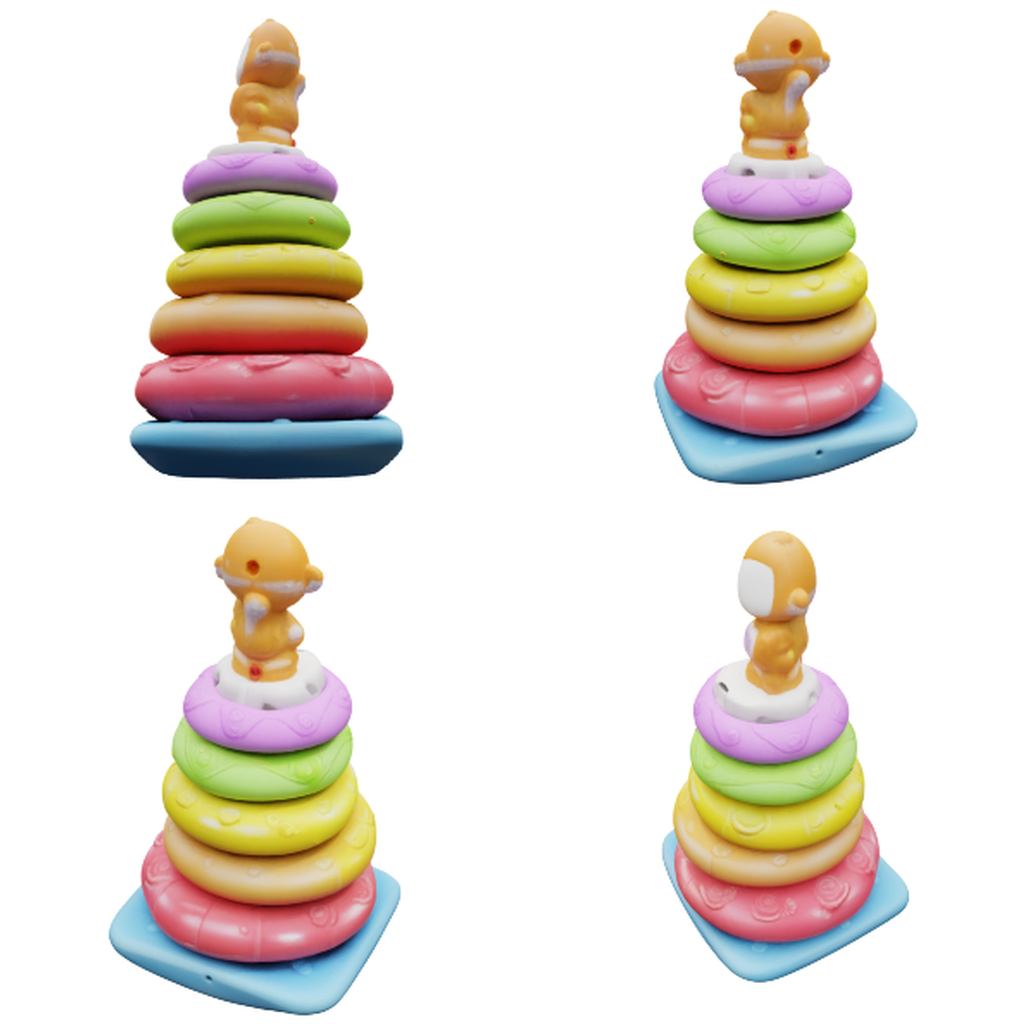} & \includegraphics[width=\comprimsize,height=\comprimsize]{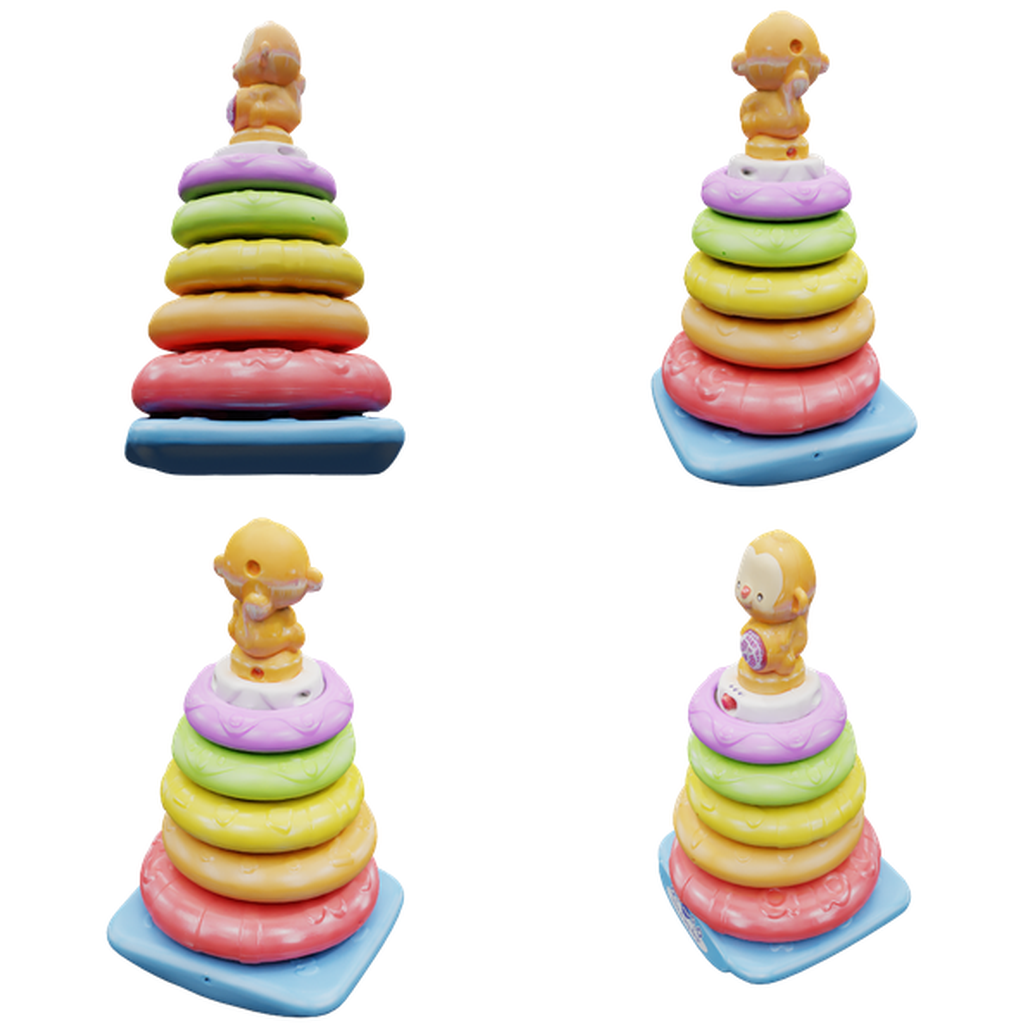} \\
\includegraphics[width=\comprimsize,height=\comprimsize]{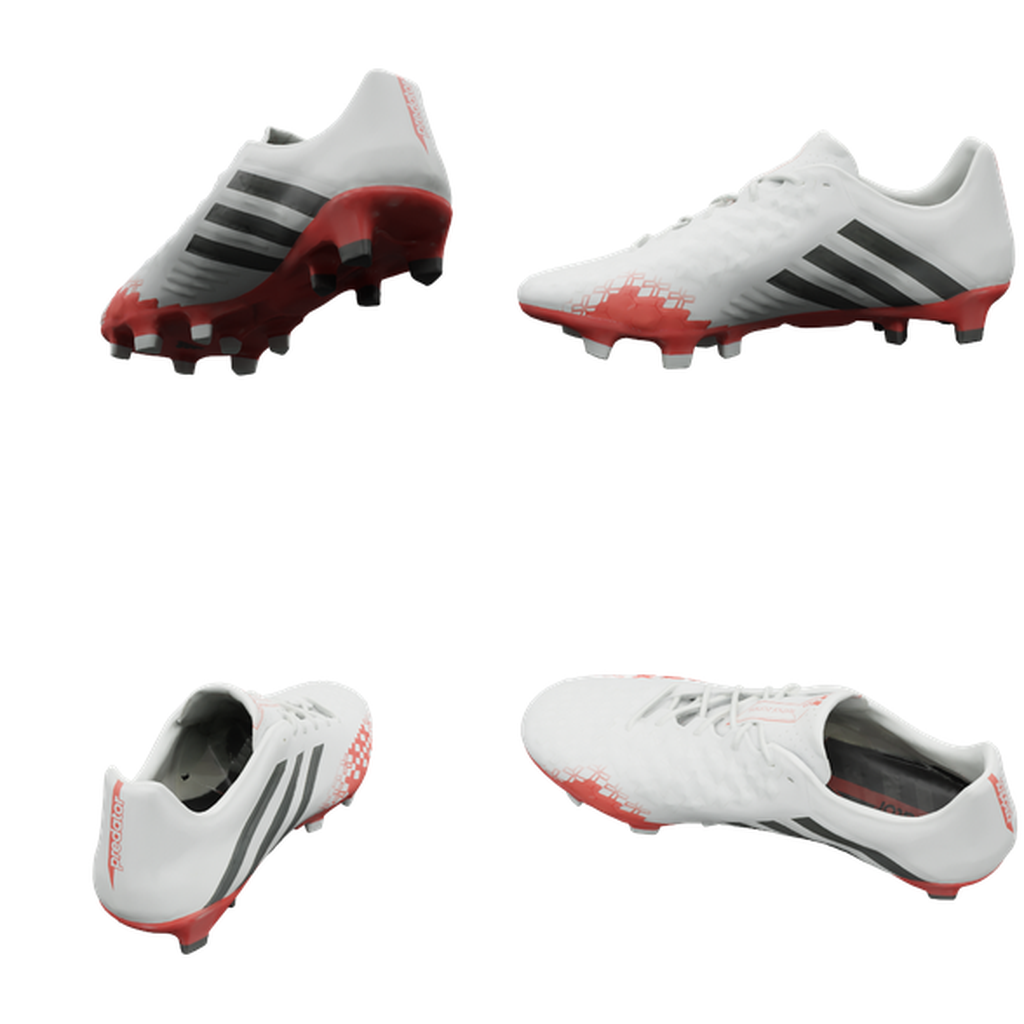} & \includegraphics[width=\comprimsize,height=\comprimsize]{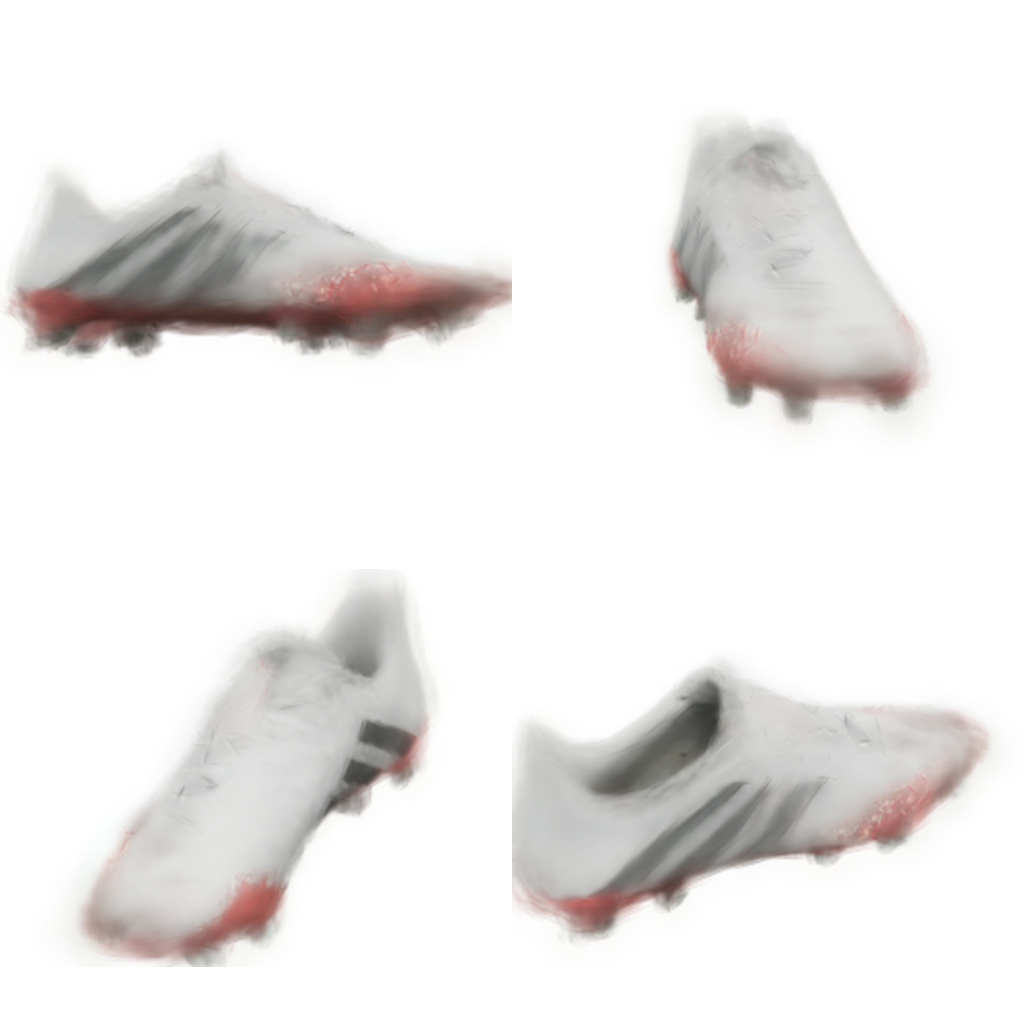} & \includegraphics[width=\comprimsize,height=\comprimsize]{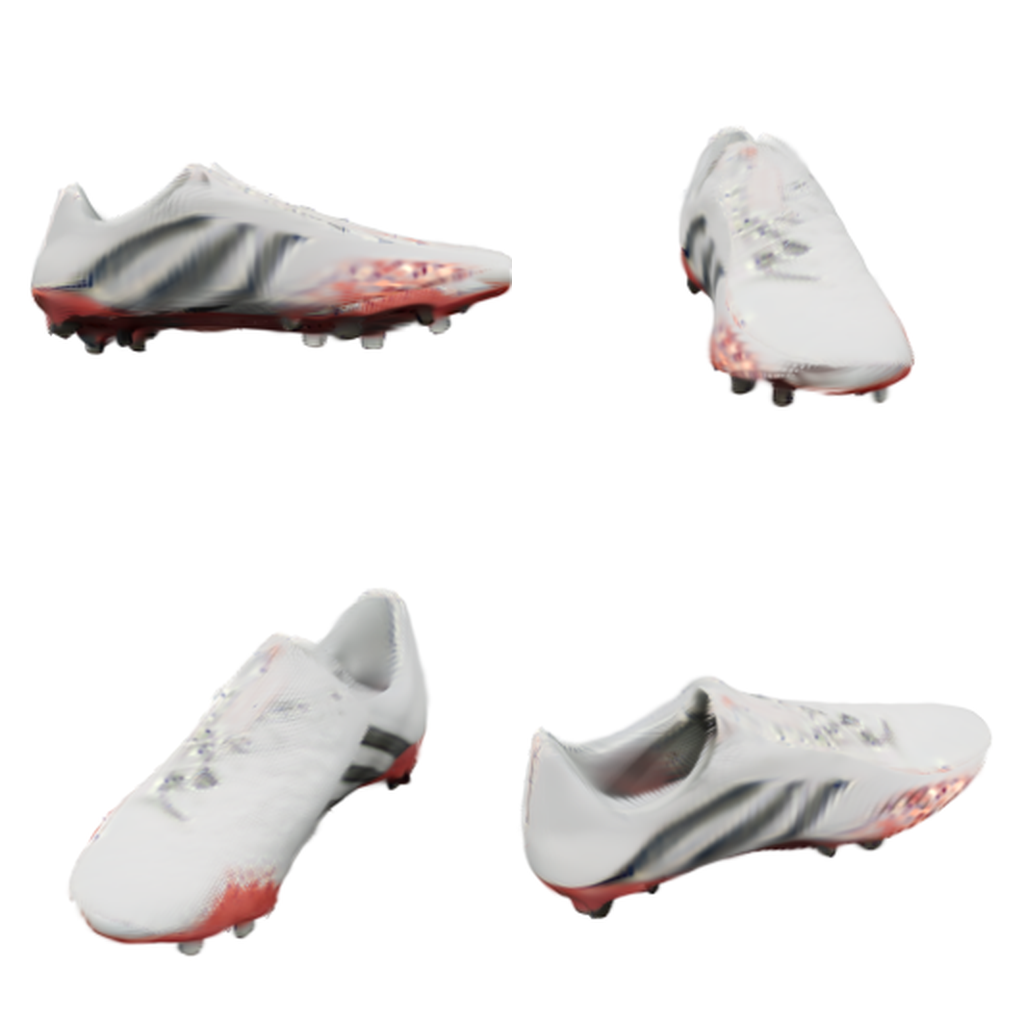} & \includegraphics[width=\comprimsize,height=\comprimsize]{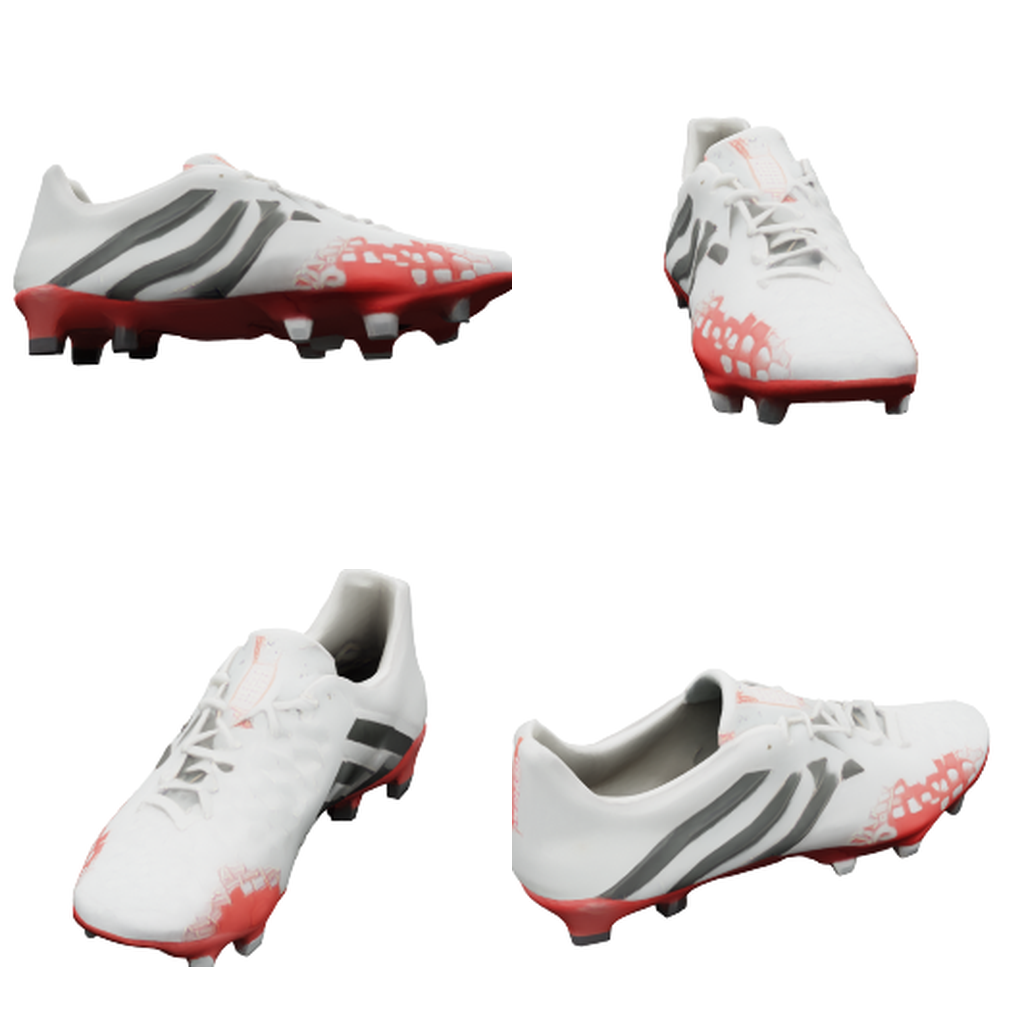} & \includegraphics[width=\comprimsize,height=\comprimsize]{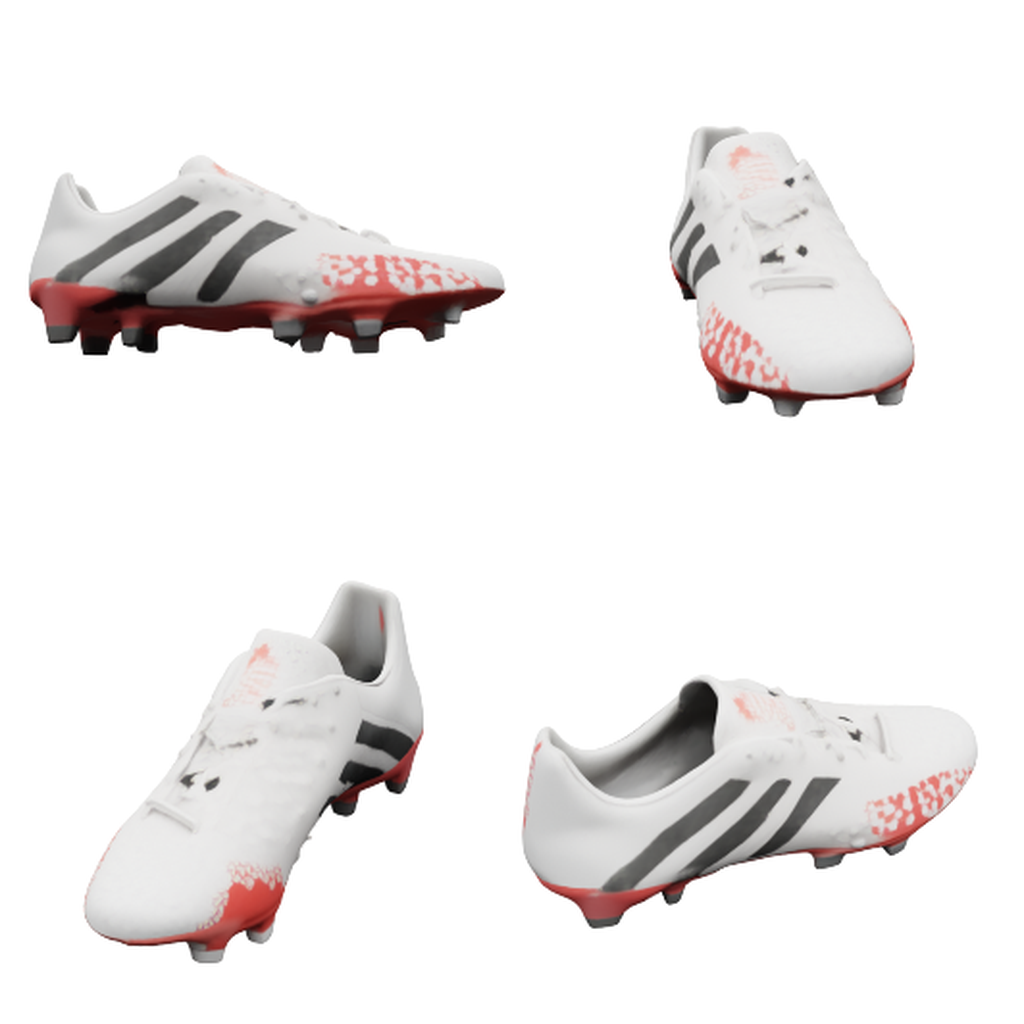} & \includegraphics[width=\comprimsize,height=\comprimsize]{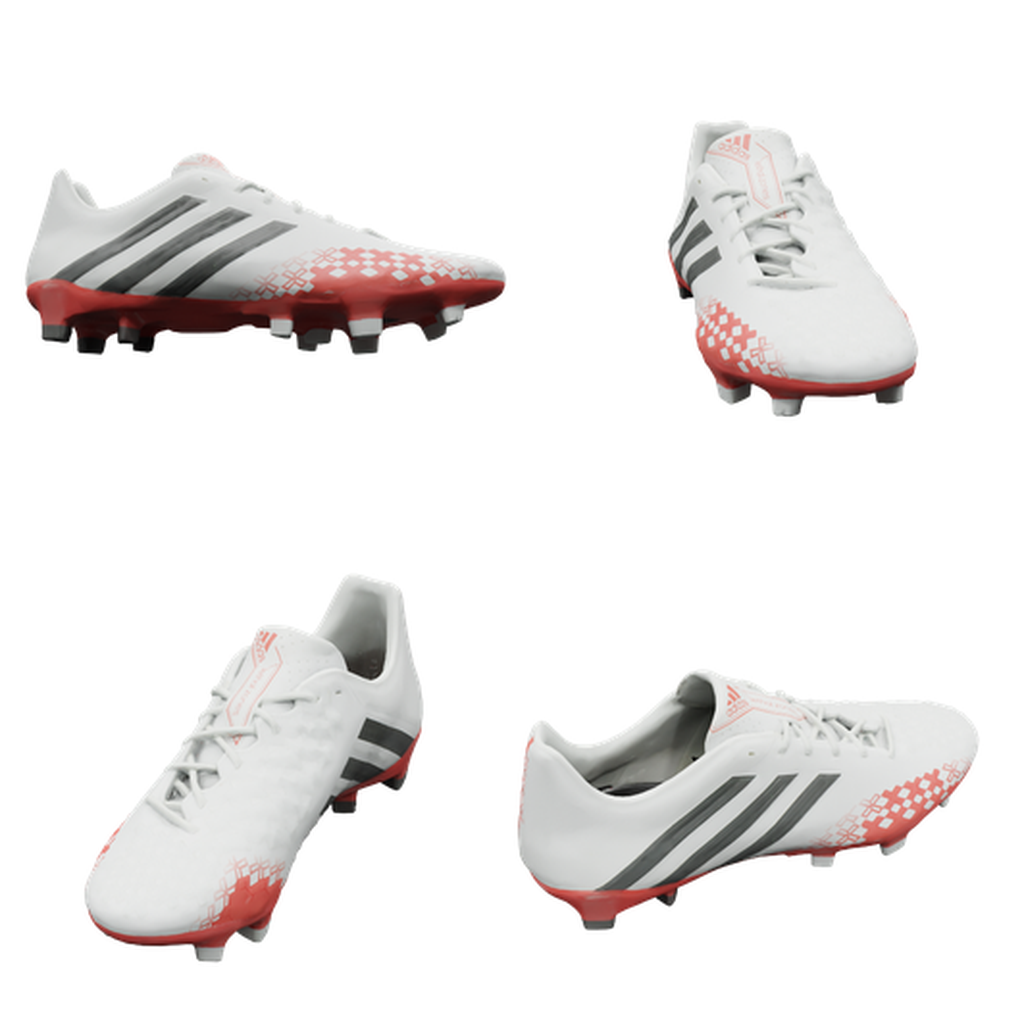} \\
\end{tabular}

\caption{
\textbf{Qualitative comparison with feed-forward reconstruction methods} on Toys4k (top two rows) and GSO (bottom two). All non-input columns render the same held-out novel view in the GT camera frame after the input-view pointmap LSQ alignment of Table~\ref{tab:main_recon}.
}
\label{fig:qualitative_recon}
\end{figure*}
\vspace{1mm}

\subsection{Experimental Setup}
\label{sec:exp_setup}

\paragraph{Datasets.}
We train our model on four public 3D-asset datasets: Objaverse-XL~\cite{deitke2023objaverse} (GitHub and Sketchfab partitions), HSSD~\cite{khanna2024habitat}, 3D-FUTURE~\cite{fu20213d}, and ABO~\cite{collins2022abo}. After \trellis's aesthetic filter ($\ge 4.5$), about $430$K instances remain. Each is rendered following \trellis's protocol, with $24$ condition views at $1024{\times}1024$ from cameras uniformly sampled on the upper hemisphere. We evaluate on two held-out test sets, Toys4k~\cite{stojanov2021using} (synthetic) and GSO~\cite{downs2022google} (real-captured), neither overlapping the training corpus.

\paragraph{Evaluation protocol.}
We assess photometric fidelity, 3D geometry, and camera-pose accuracy. Each test set uses its first $400$ objects with two disjoint $4$-view sets, $\{0,6,12,18\}$ and $\{3,9,15,21\}$, yielding $800$ object--viewset instances for aggregation. As baselines fall into three capability families, we apply the most suitable evaluation protocol to each: (i) least squares on predicted vs ground-truth input-view pointmaps for feed-forward reconstruction, tight only with accurate predicted geometry and poses; (ii) similarity ICP on sampled point clouds for 3D generation; and (iii) gauge-invariant pairwise relative-pose comparison on unordered input-view pairs for pose-predicting methods. \ours{} is reevaluated under each applicable protocol; comparisons are restricted to methods sharing a table and protocol. We report PSNR, SSIM, and LPIPS~\cite{zhang2018unreasonable} on $512{\times}512$ novel views at the 8 held-out condition-view indices $\{2,4,8,10,14,16,20,22\}$; bidirectional Chamfer Distance (CD; reported as the median across objects, since a few severe failures dominate the mean) and F-scores at $\tau\in\{0.01,\,0.05\}$; and Relative Rotation Error (RRE) with Acc$@\{5^{\circ}, 15^{\circ}\}$, Relative Translation Error (RTE), and mean absolute FOV error (RFOV; only for methods that predict intrinsics). See \Appref{sec:suppl_alignment} for the full alignment protocol and \Appref{sec:suppl_pose_metrics} for the pose metric definitions.

\paragraph{Baseline methods.}
We compare against a representative set of multi-view image-to-3D baselines: (a) \textbf{\trellis-S}~\cite{xiang2025structured}: \trellis run as a single-image generator on each input view independently, with the mean over the four outputs reported; (b) \textbf{\trellis-M}~\cite{xiang2025structured}: \trellis with multi-view input in stochastic mode, where each denoising step is conditioned on one randomly chosen input view; (c) \textbf{SAM 3D}~\cite{sam3dteam2025sam3d3dfyimages}: a single-image-to-3D generator run once per input view, with the mean over the four outputs reported; (d) \textbf{FreeSplatter}~\cite{xu2024freesplatter}: predicts per-view pixel-aligned 3D Gaussians together with input camera parameters in a pose-free setting; (e) \textbf{LucidFusion}~\cite{he2025lucidfusion}: predicts per-view relative coordinate maps that are fused into a 3D Gaussian output; (f) \textbf{ReconViaGen}~\cite{chang2025reconviagen}: combines feed-forward reconstruction with generative refinement, using VGGT-derived geometry tokens to condition a 3D generator; (g) \textbf{VGGT}~\cite{wang2025vggt}: a transformer that jointly regresses per-view pointmaps and camera extrinsics/intrinsics from uncalibrated images. Our table appear in three families: 3D generation (a,b,c,f), feed-forward reconstruction (d,e,f), and pose prediction (d,f,g).

\subsection{Main Results}
\label{sec:exp_main}

\begin{table}[t]
\renewcommand{\arraystretch}{1.08}
\setlength{\tabcolsep}{3pt}
\centering
\caption{
\textbf{Comparison with 3D generation methods.}
For methods that produce a canonical 3D output without input camera poses, we align predictions to the ground truth via similarity ICP on $100$K points sampled from the rendered surface of the output mesh of each side, then evaluate photometric quality on novel views and 3D geometric accuracy in the ground-truth coordinate frame. 
}
\label{tab:main_gen}
\resizebox{\linewidth}{!}{
\begin{tabular}{lcccccccccccc}
\toprule
\multirow{2}{*}{\textbf{Method}} &
\multicolumn{6}{c}{\textbf{Toys4k}} &
\multicolumn{6}{c}{\textbf{GSO}} \\
\cmidrule(lr){2-7}\cmidrule(lr){8-13}
&
\textbf{PSNR}$\uparrow$ & \textbf{SSIM}$\uparrow$ & \textbf{LPIPS}$\downarrow$ & \textbf{CD-med}$\downarrow$ & \textbf{F@.01}$\uparrow$ & \textbf{F@.05}$\uparrow$ &
\textbf{PSNR}$\uparrow$ & \textbf{SSIM}$\uparrow$ & \textbf{LPIPS}$\downarrow$ & \textbf{CD-med}$\downarrow$ & \textbf{F@.01}$\uparrow$ & \textbf{F@.05}$\uparrow$ \\
\midrule
SAM 3D~\cite{sam3dteam2025sam3d3dfyimages} &
20.49 & 0.9155 & 0.0926 & 1.7935 & 0.3292 & 0.8250 &
19.49 & 0.8995 & 0.1112 & 1.3716 & 0.3387 & 0.8505 \\
\trellis-S~\cite{xiang2025structured} &
21.78 & 0.9205 & 0.0874 & 1.2032 & 0.4064 & 0.8881 &
20.00 & 0.8943 & 0.1127 & 1.2418 & 0.3663 & 0.8695 \\
\trellis-M~\cite{xiang2025structured} &
24.01 & 0.9355 & 0.0658 & 0.4329 & 0.5570 & 0.9604 &
22.28 & 0.9138 & 0.0867 & 0.5200 & 0.5079 & 0.9655 \\
ReconViaGen~\cite{chang2025reconviagen} &
23.26 & 0.9316 & 0.0587 & 0.6955 & 0.4374 & 0.9555 &
21.98 & 0.9108 & 0.0778 & 0.6500 & 0.4428 & 0.9434 \\
\rowcolor{tableRow}
\ours &
28.54 & 0.9599 & 0.0315 & 0.1133 & 0.8279 & 0.9970 &
25.50 & 0.9349 & 0.0533 & 0.1882 & 0.7283 & 0.9914 \\
\bottomrule
\end{tabular}
}
\end{table}

\newlength{\compgimsize}
\setlength{\compgimsize}{2.25cm}

\newcommand{\compgbp}{{\color{black}\rule{\compgimsize}{\compgimsize}}}
\newcommand{\compghdr}[1]{\small #1}

\begin{figure*}[!t]
\centering
\setlength{\tabcolsep}{1pt}

\begin{tabular}{*{6}{>{\centering\arraybackslash}m{\compgimsize}}}
\compghdr{Input} & \compghdr{SAM\,3D} & \compghdr{\trellis-S} & \compghdr{\trellis-M} & \compghdr{\textbf{Ours}} & \compghdr{GT} \\[2pt]
\includegraphics[width=\compgimsize,height=\compgimsize]{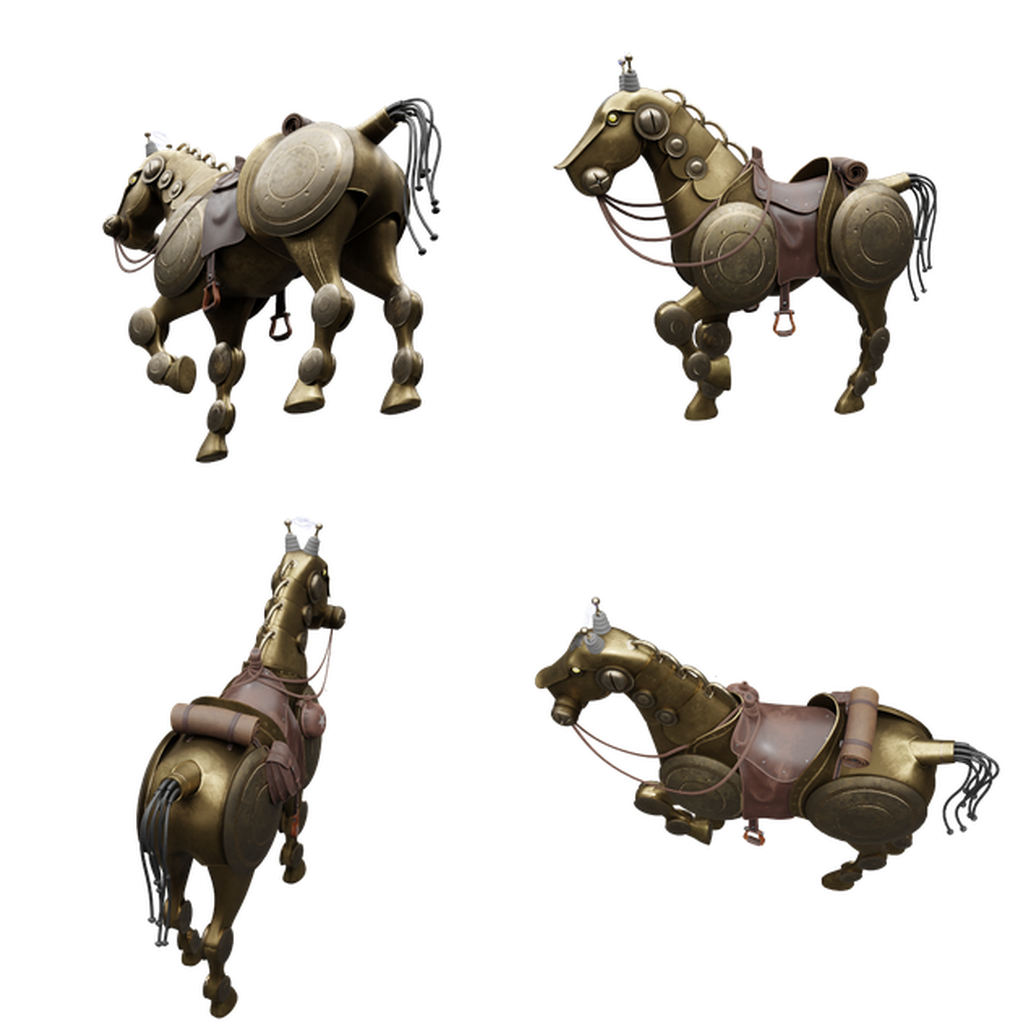} & \includegraphics[width=\compgimsize,height=\compgimsize]{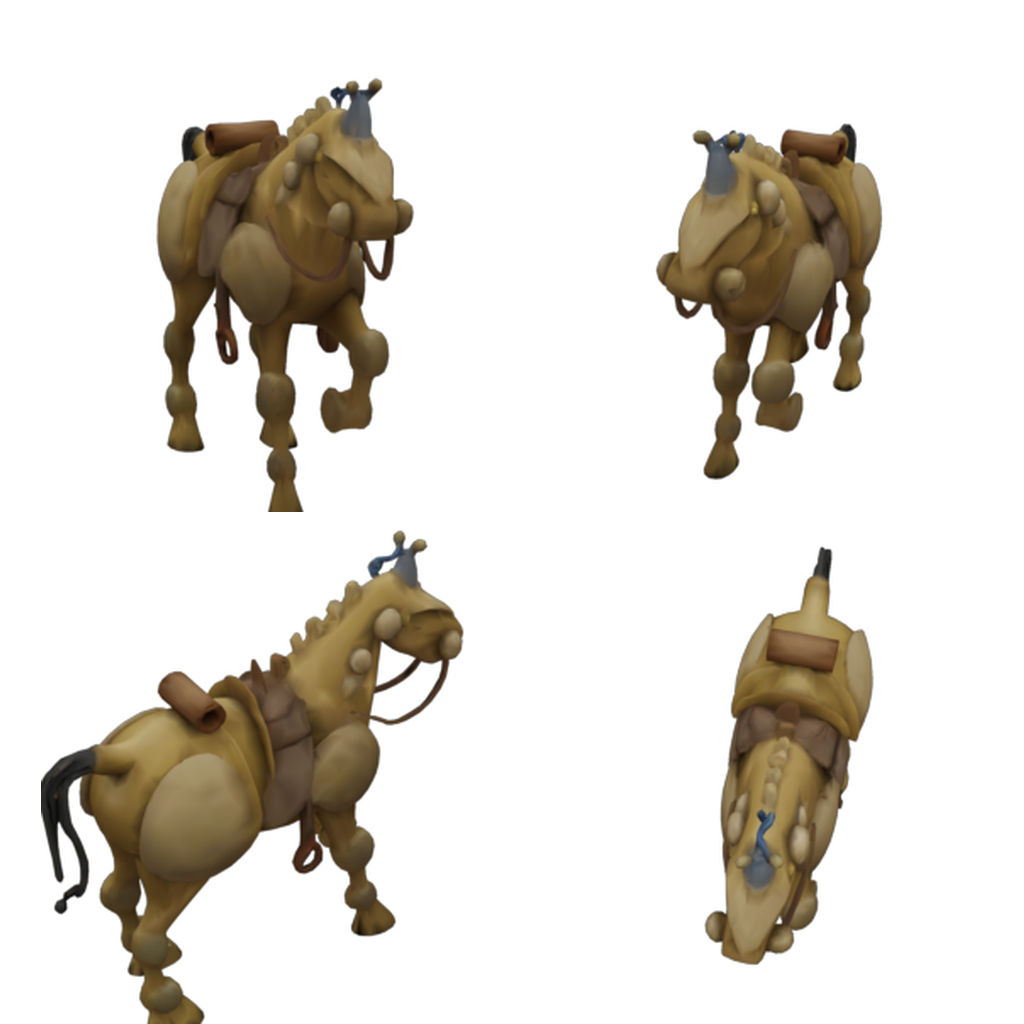} & \includegraphics[width=\compgimsize,height=\compgimsize]{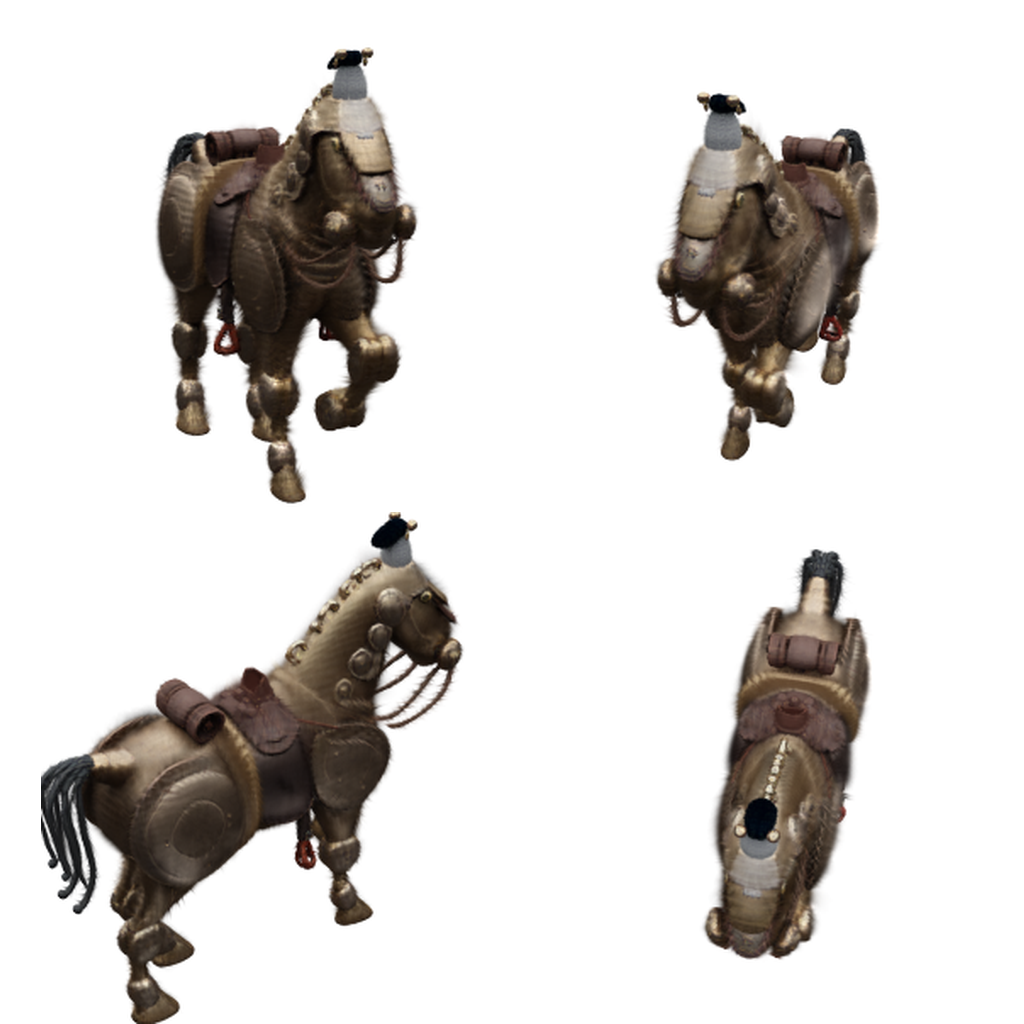} & \includegraphics[width=\compgimsize,height=\compgimsize]{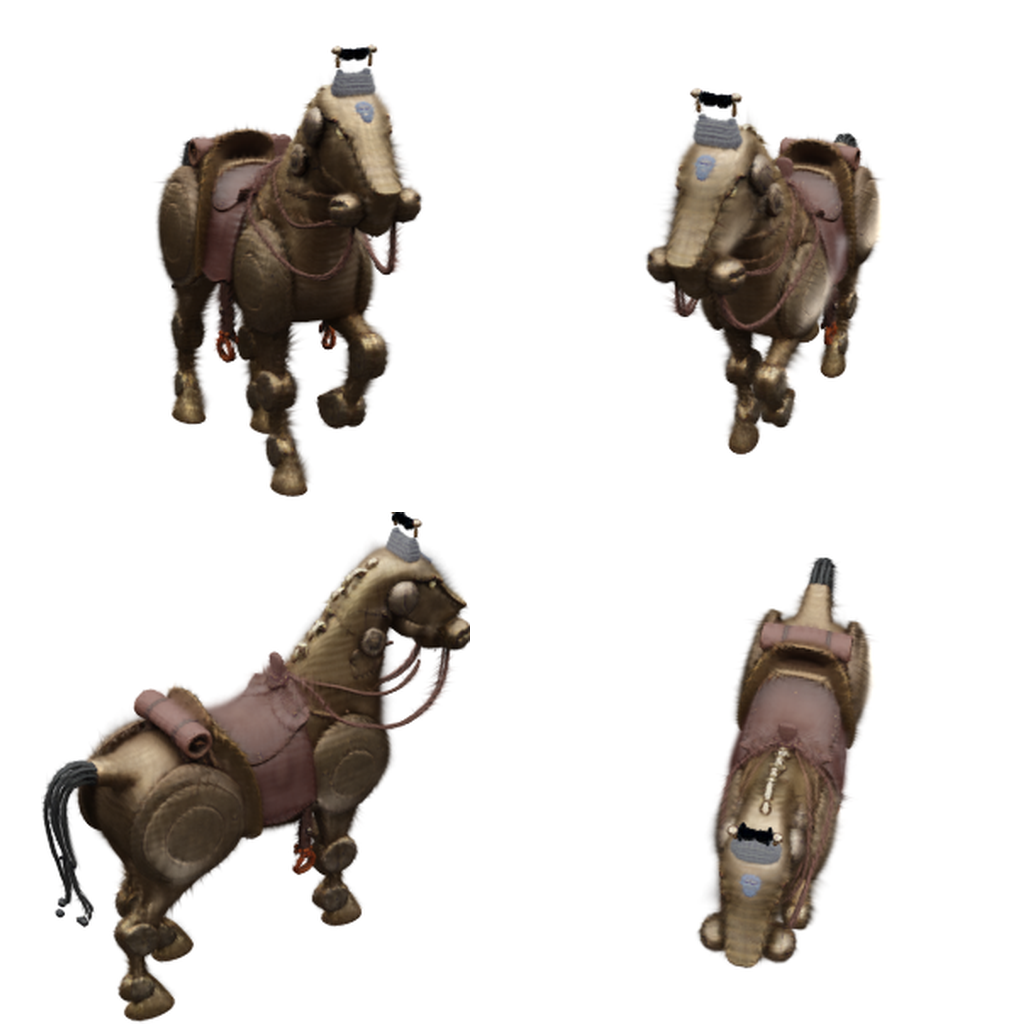} & \includegraphics[width=\compgimsize,height=\compgimsize]{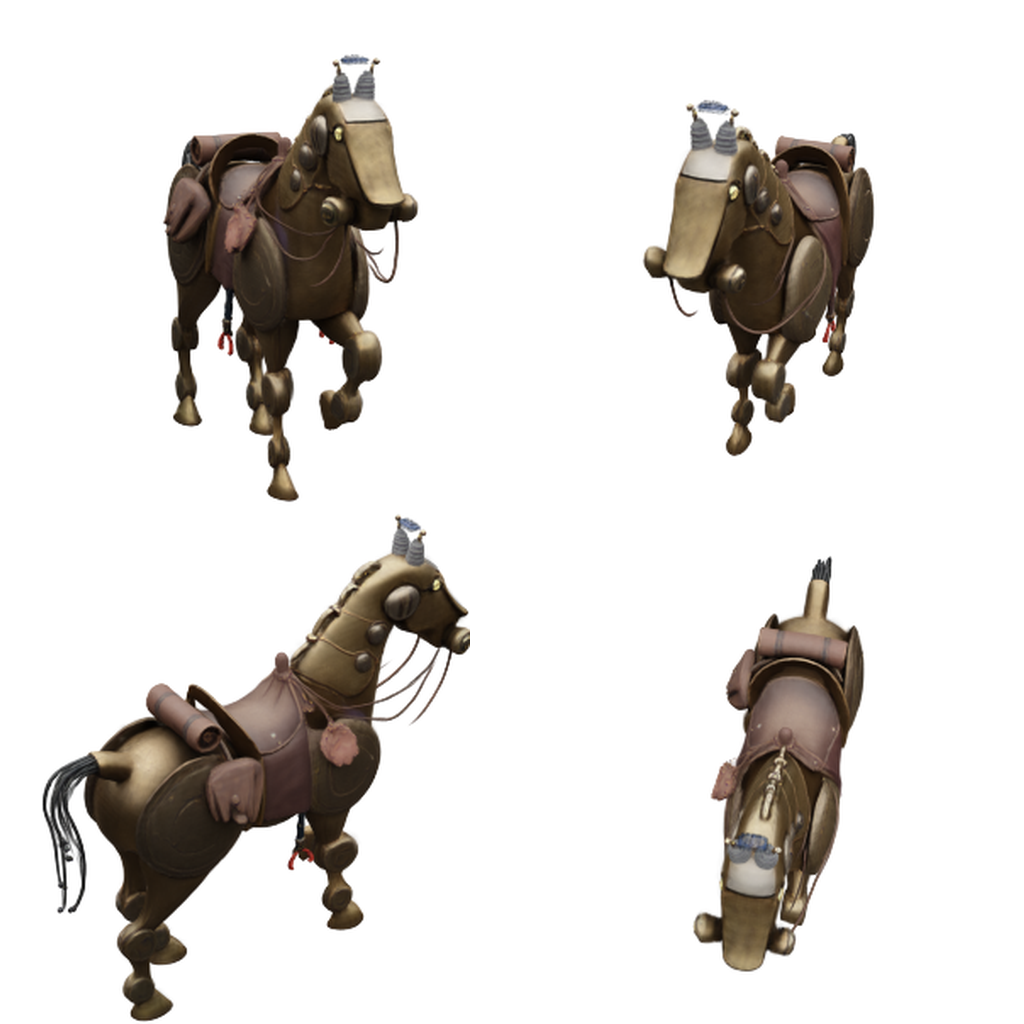} & \includegraphics[width=\compgimsize,height=\compgimsize]{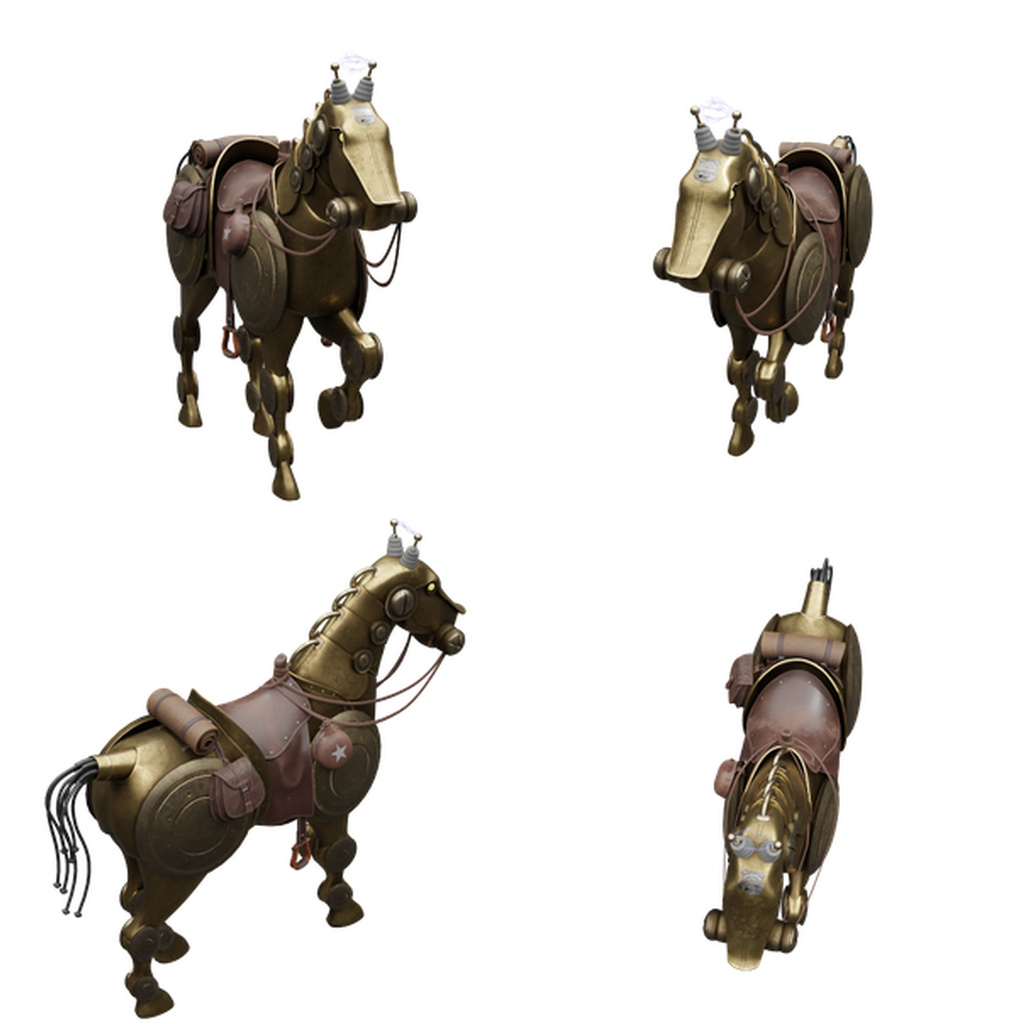} \\
\includegraphics[width=\compgimsize,height=\compgimsize]{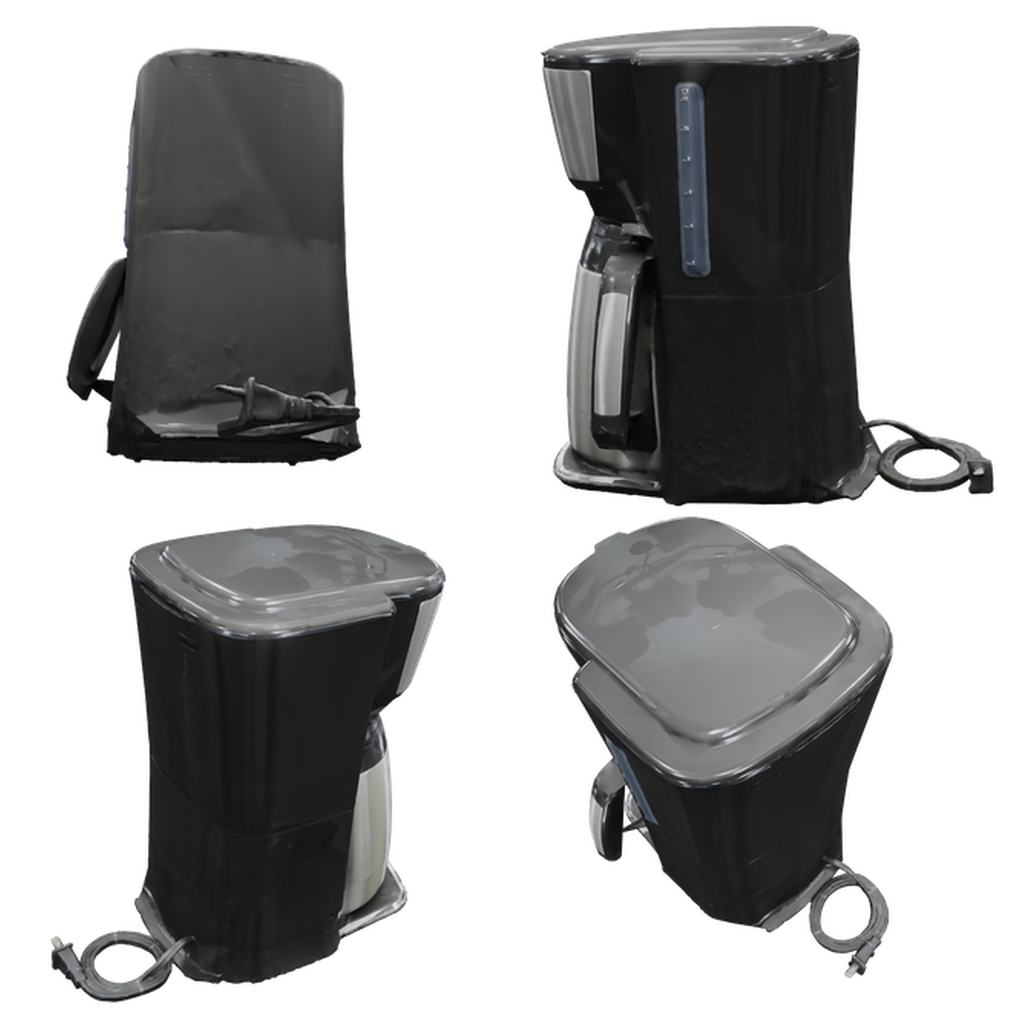} & \includegraphics[width=\compgimsize,height=\compgimsize]{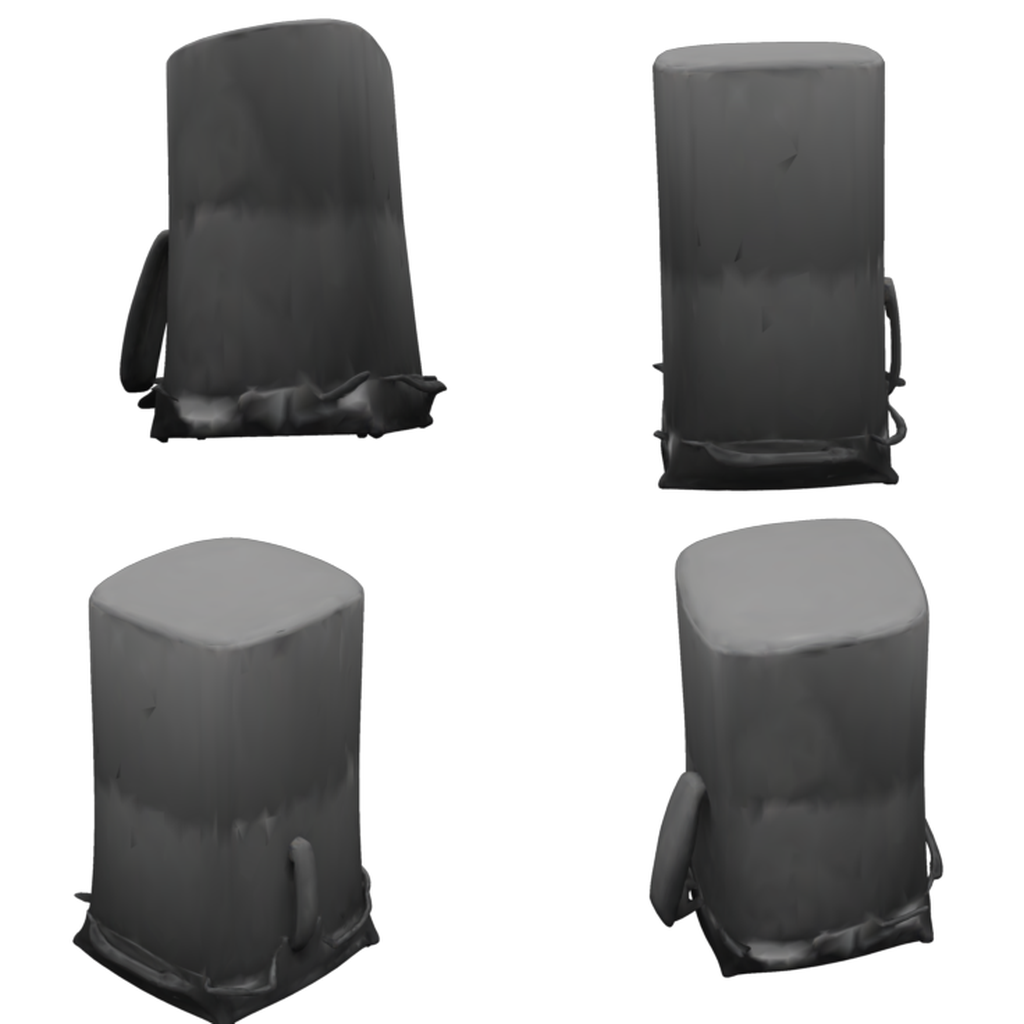} & \includegraphics[width=\compgimsize,height=\compgimsize]{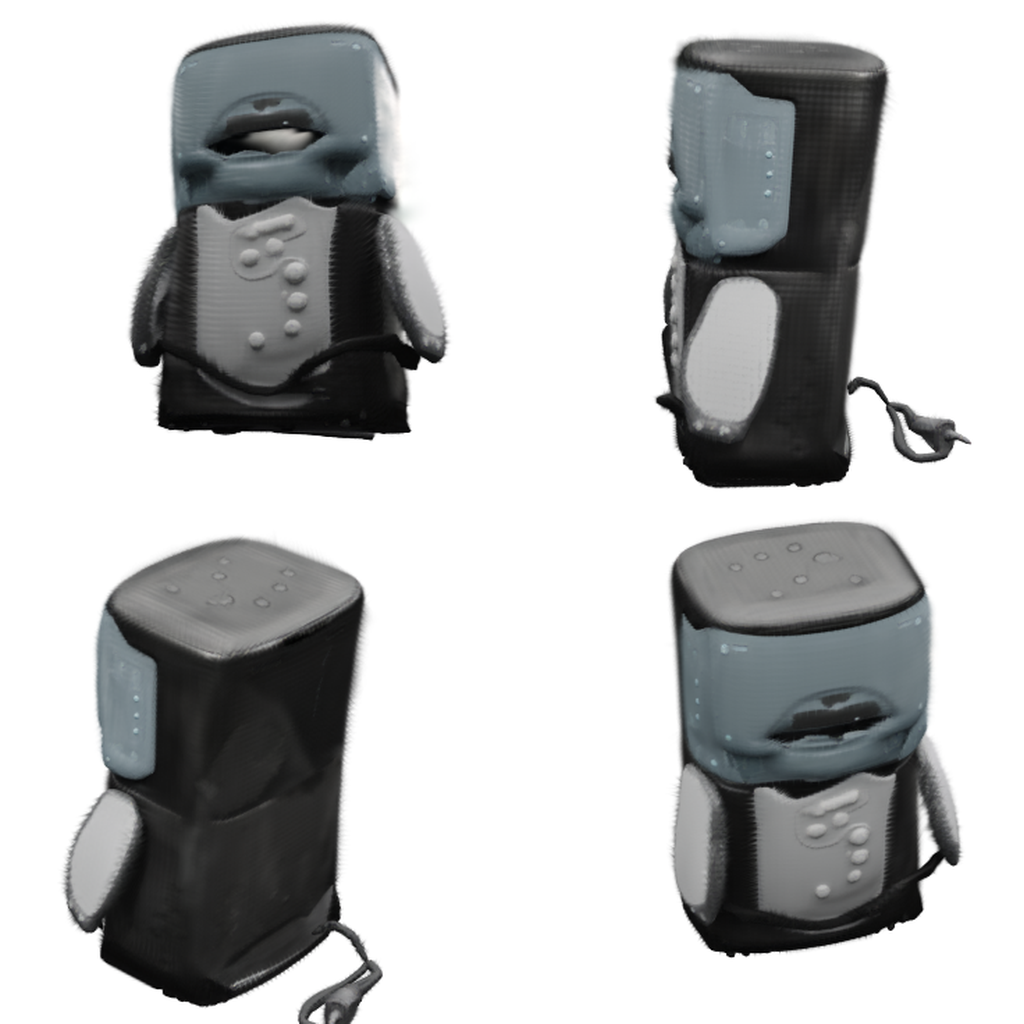} & \includegraphics[width=\compgimsize,height=\compgimsize]{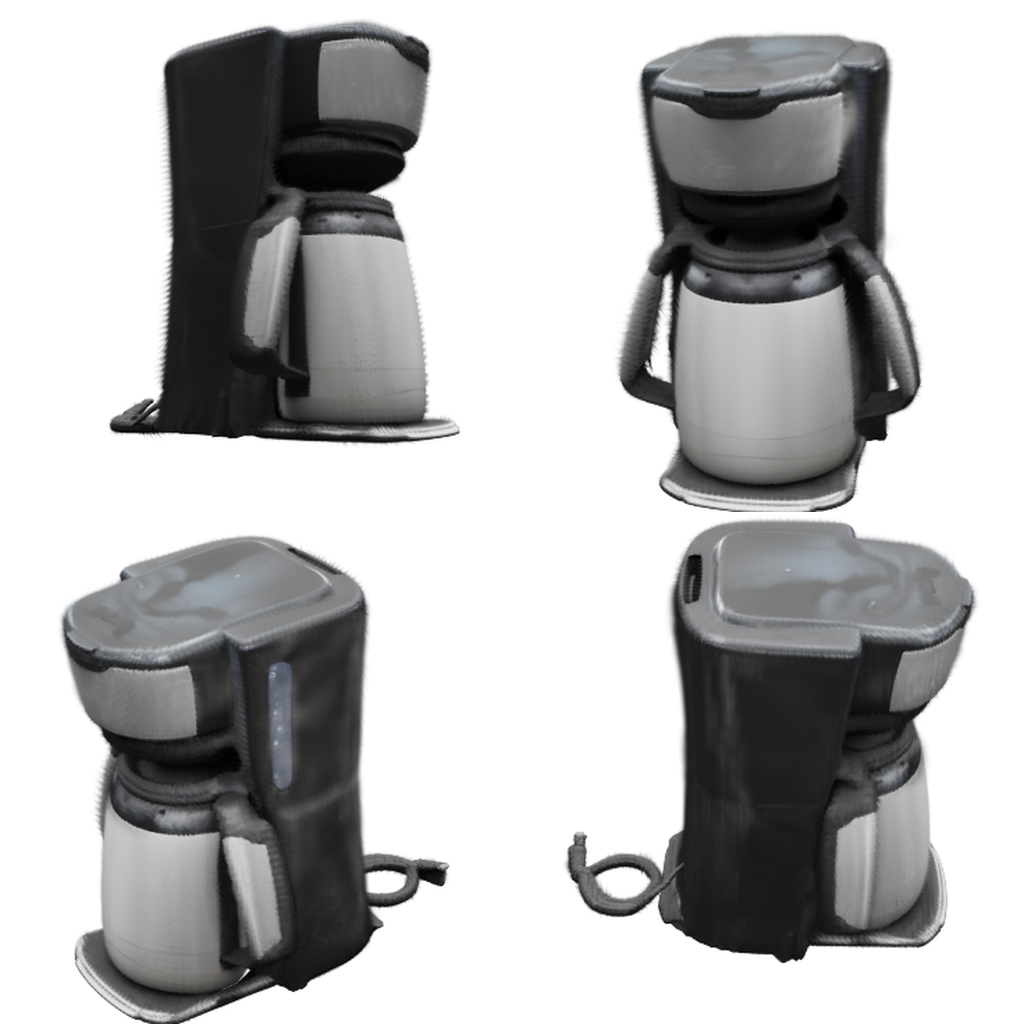} & \includegraphics[width=\compgimsize,height=\compgimsize]{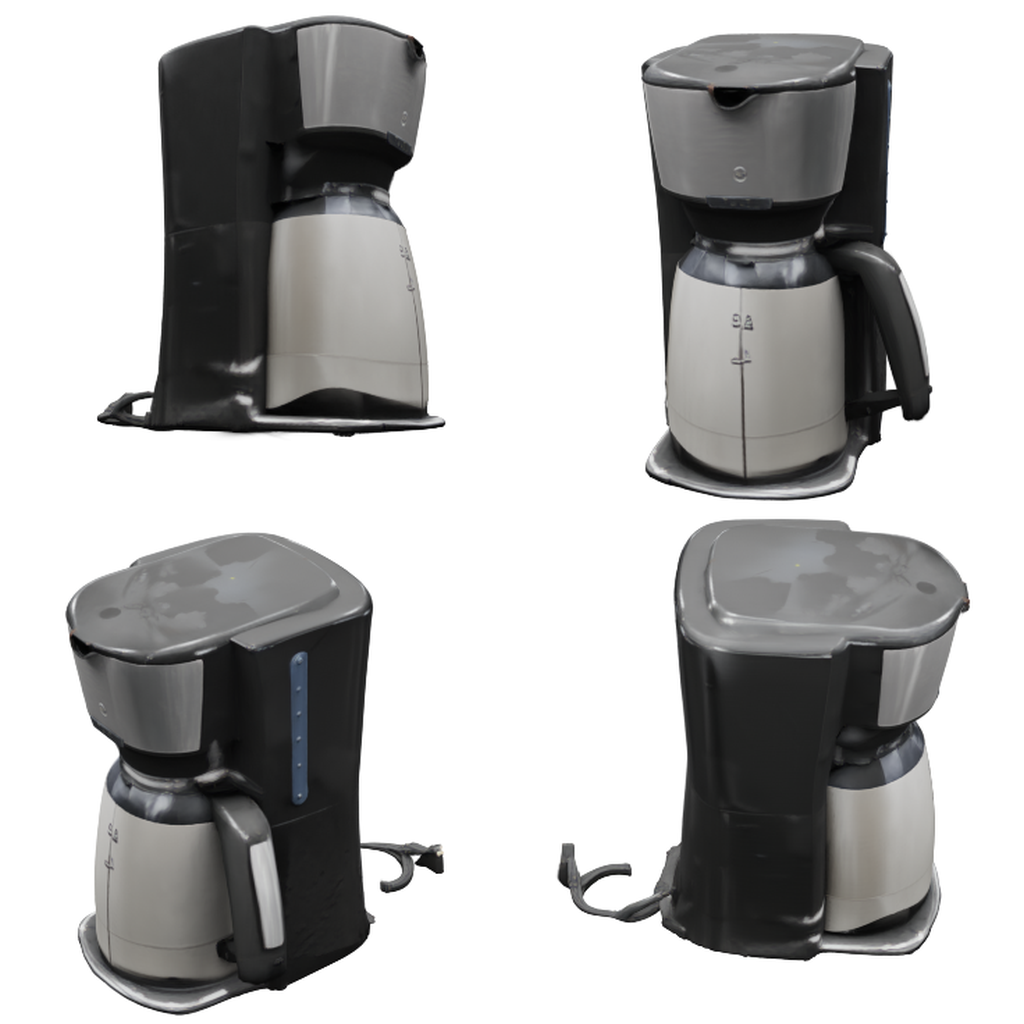} & \includegraphics[width=\compgimsize,height=\compgimsize]{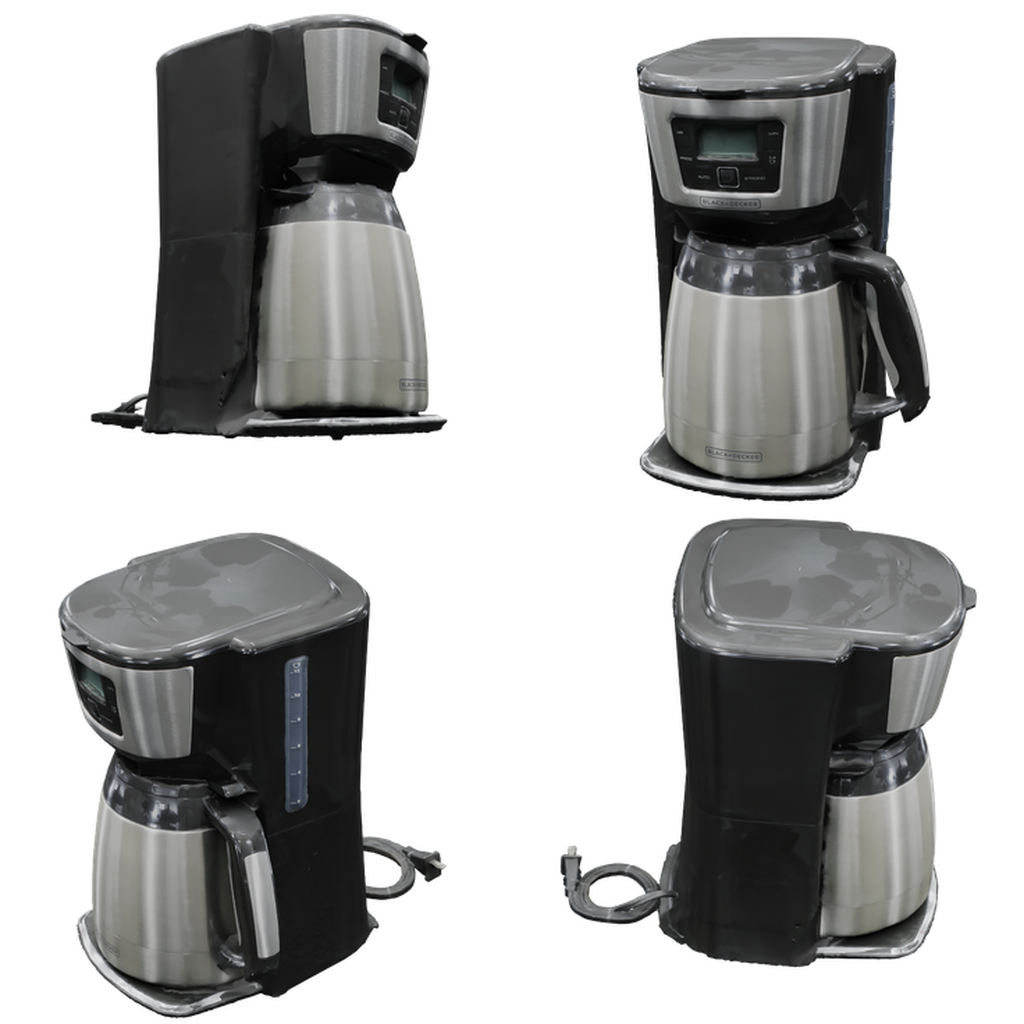} \\
\includegraphics[width=\compgimsize,height=\compgimsize]{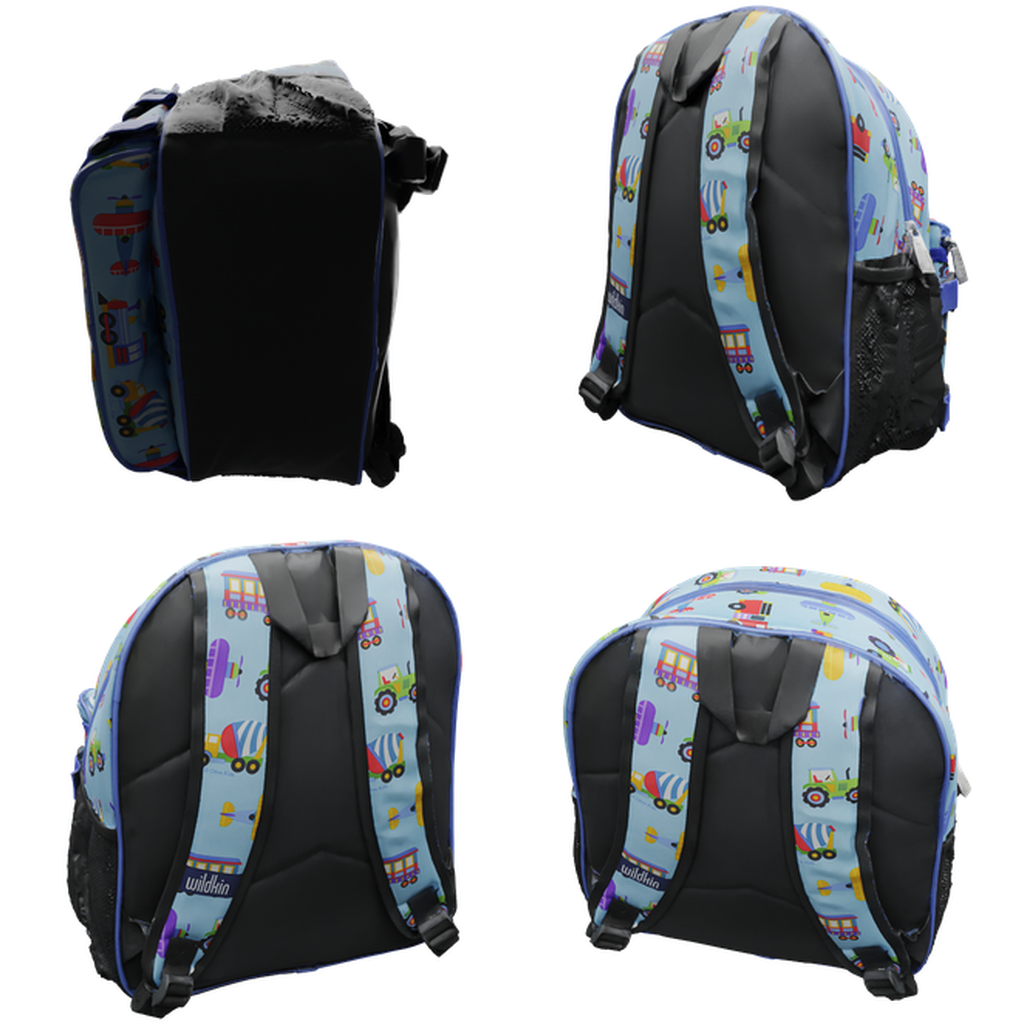} & \includegraphics[width=\compgimsize,height=\compgimsize]{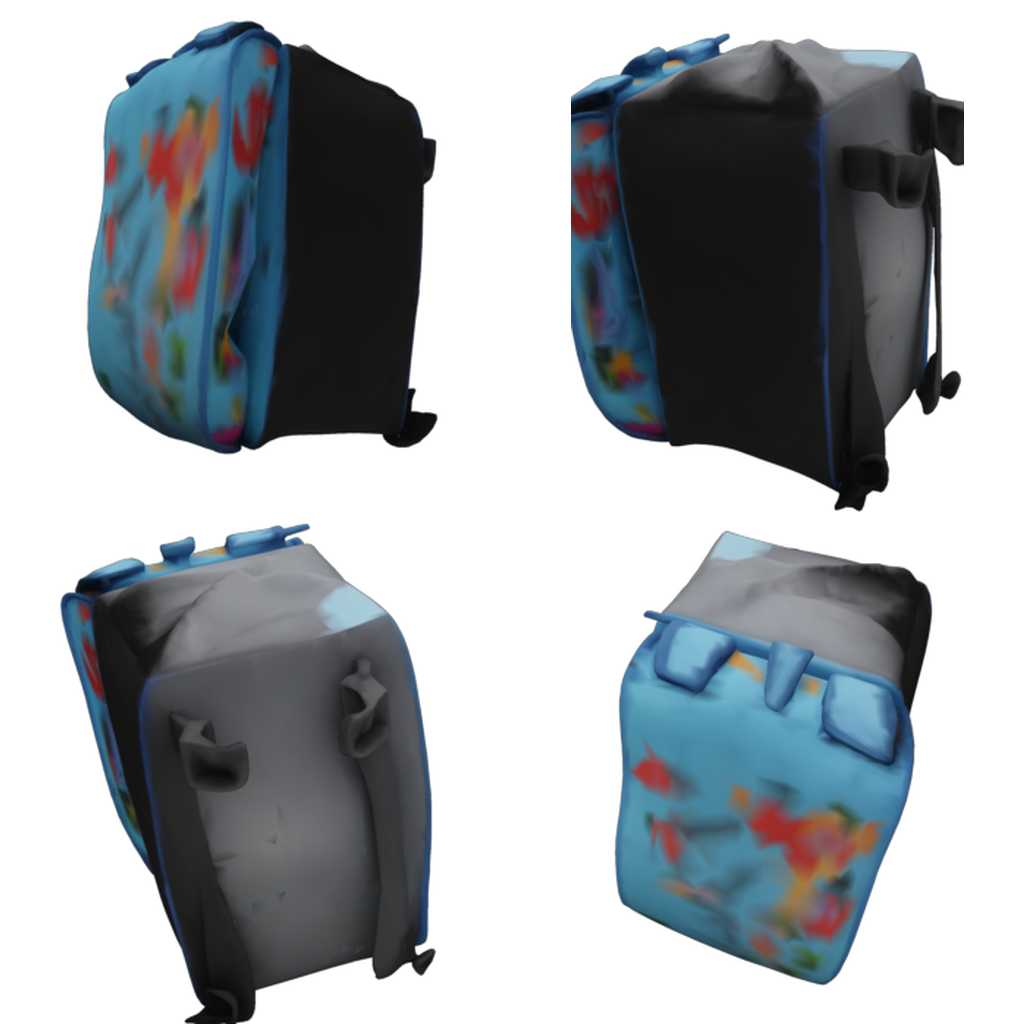} & \includegraphics[width=\compgimsize,height=\compgimsize]{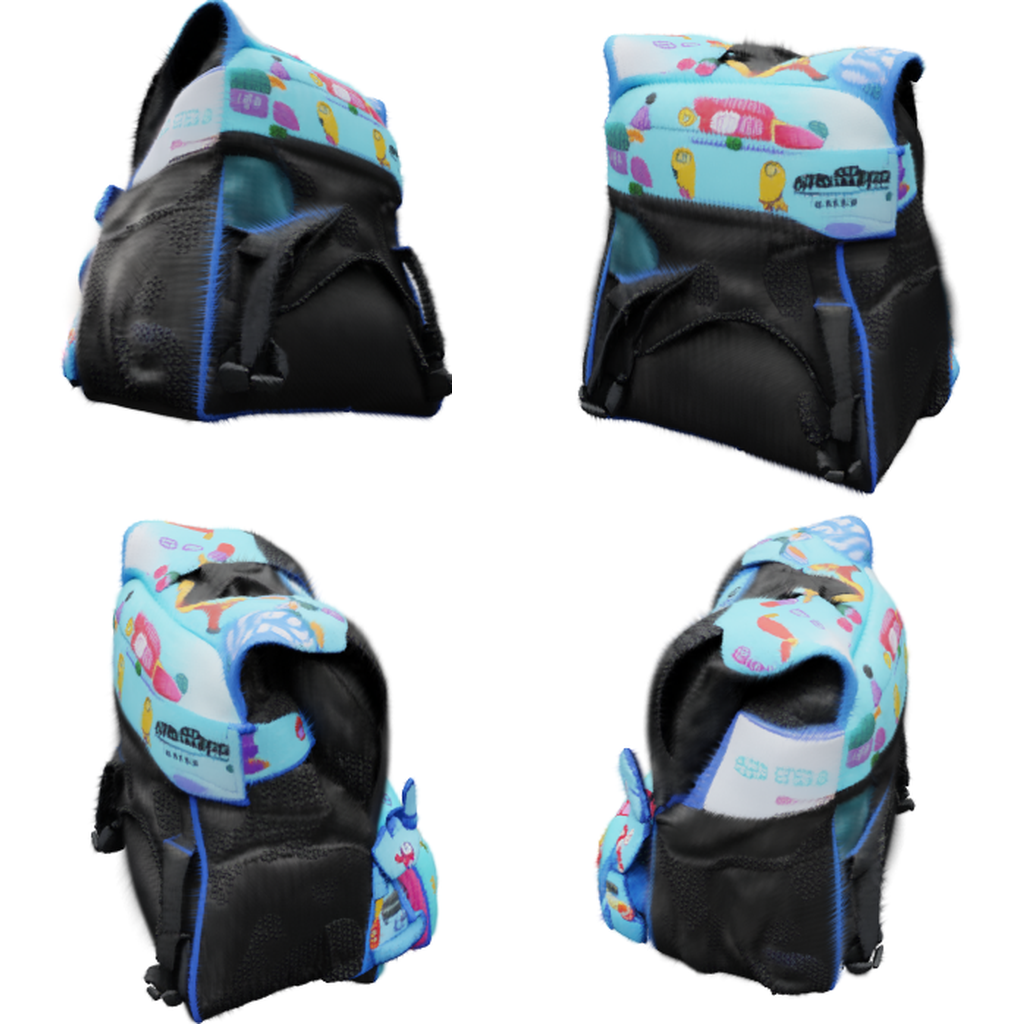} & \includegraphics[width=\compgimsize,height=\compgimsize]{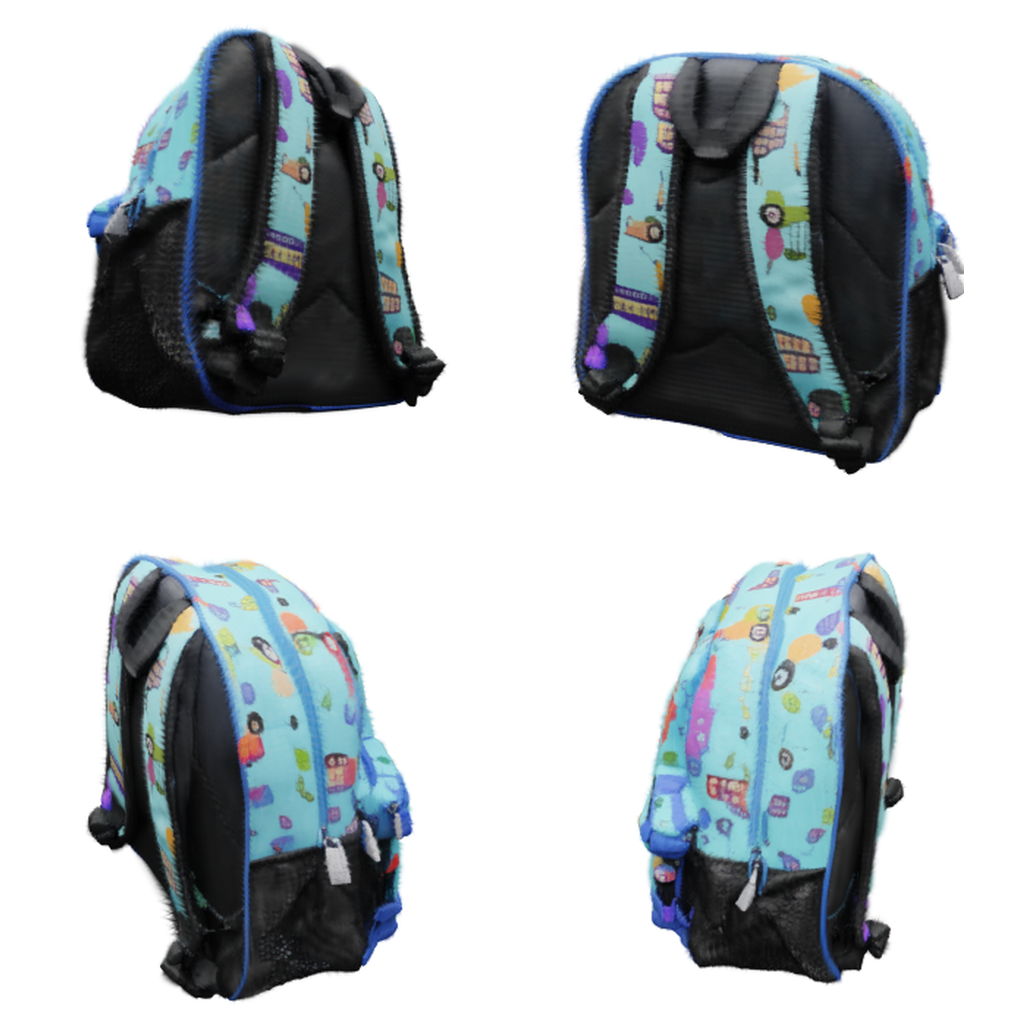} & \includegraphics[width=\compgimsize,height=\compgimsize]{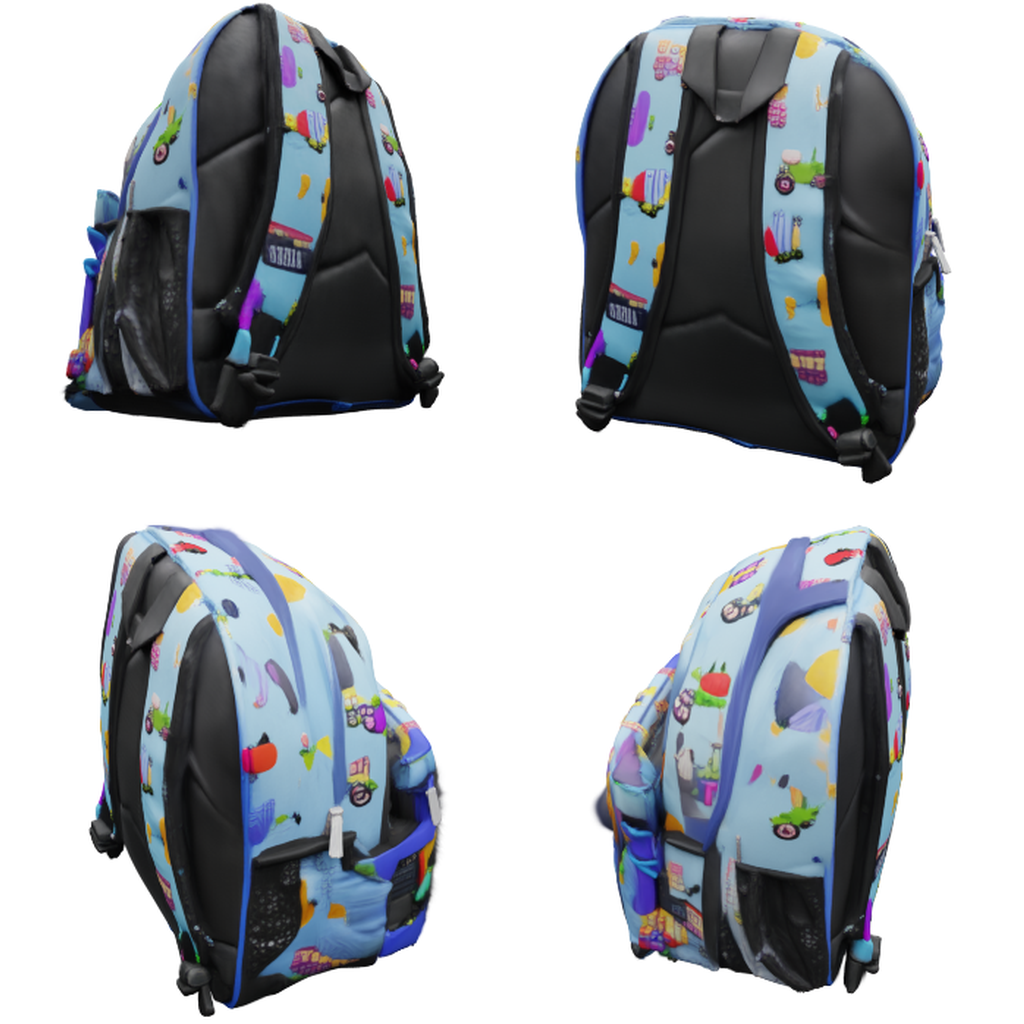} & \includegraphics[width=\compgimsize,height=\compgimsize]{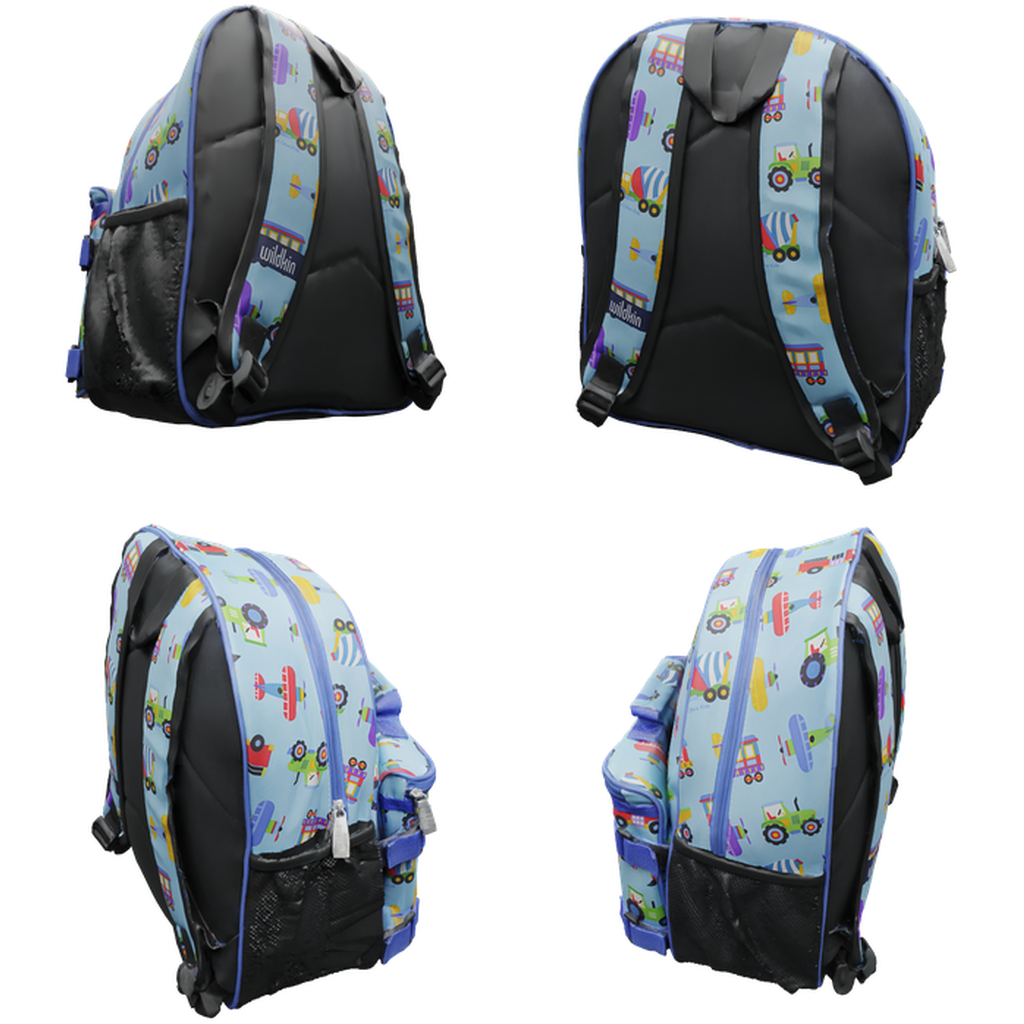} \\
\includegraphics[width=\compgimsize,height=\compgimsize]{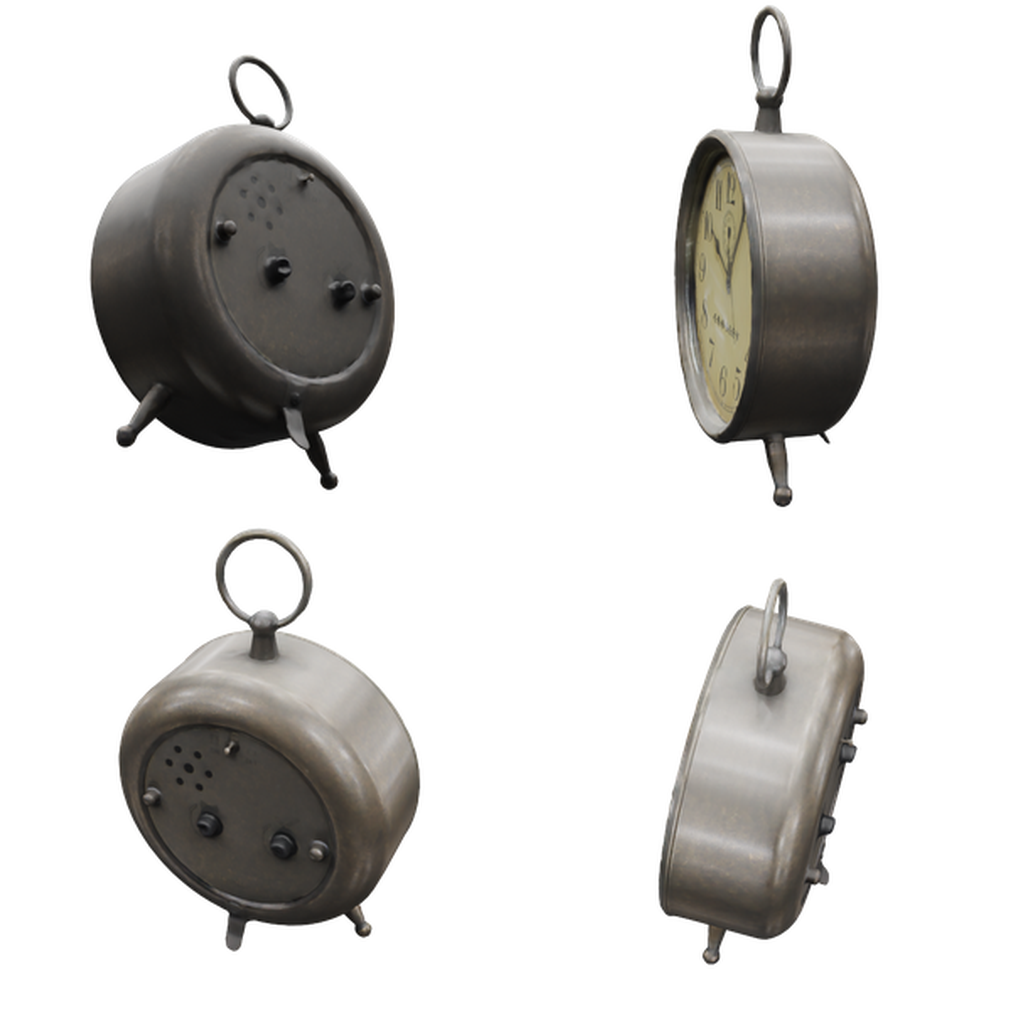} & \includegraphics[width=\compgimsize,height=\compgimsize]{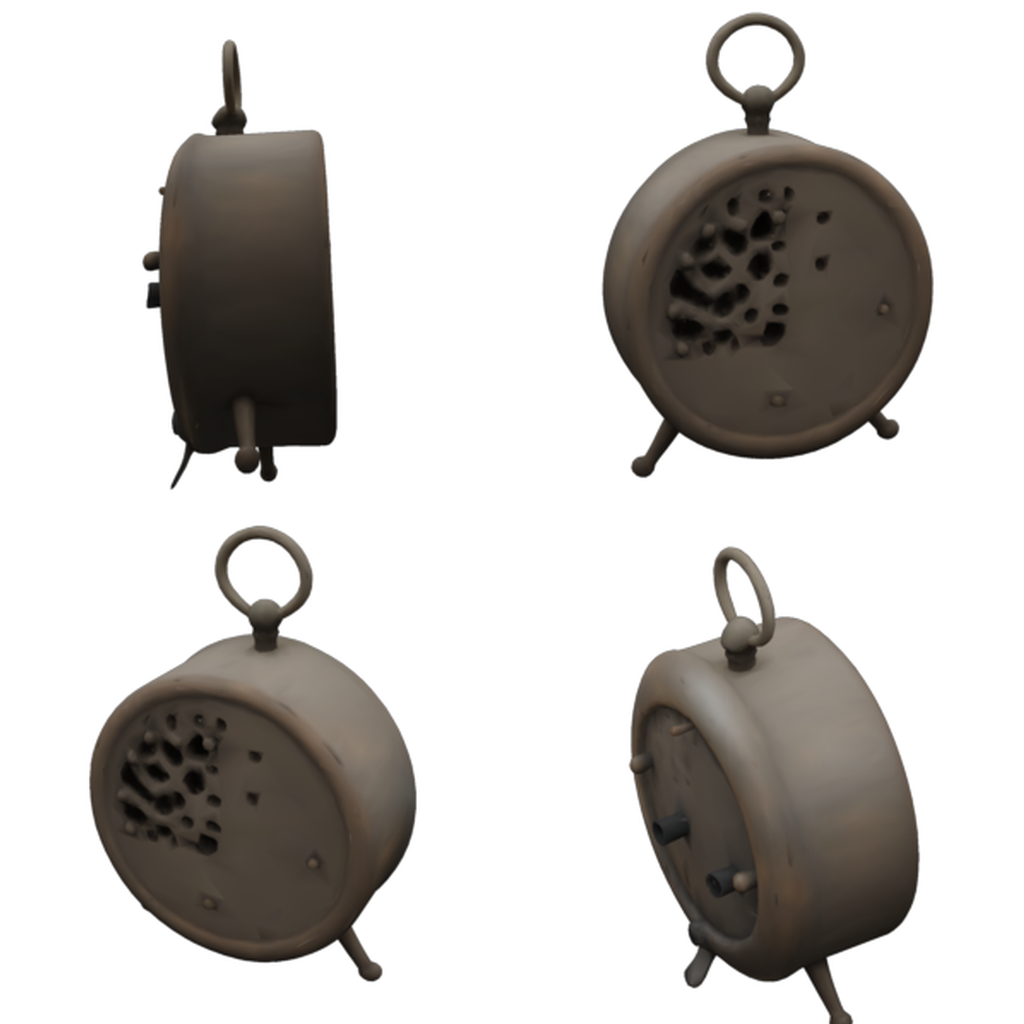} & \includegraphics[width=\compgimsize,height=\compgimsize]{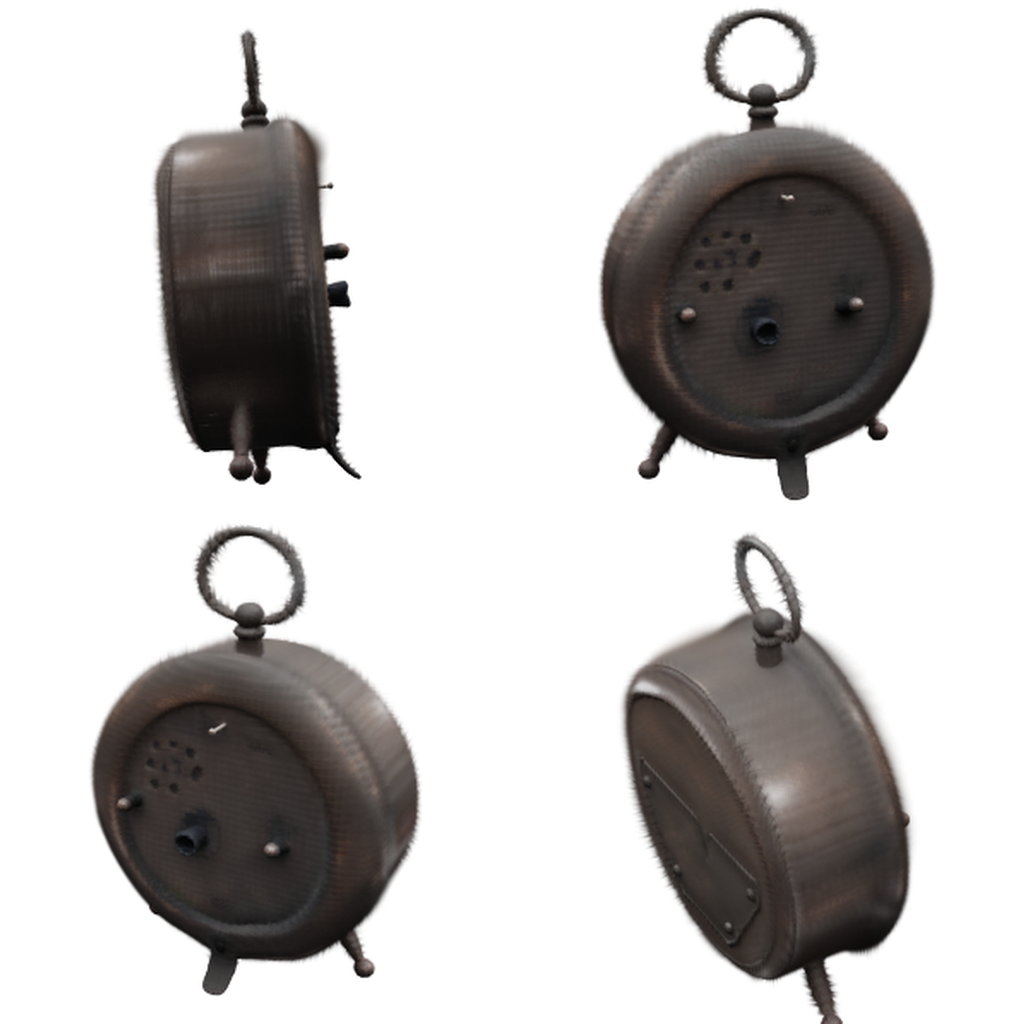} & \includegraphics[width=\compgimsize,height=\compgimsize]{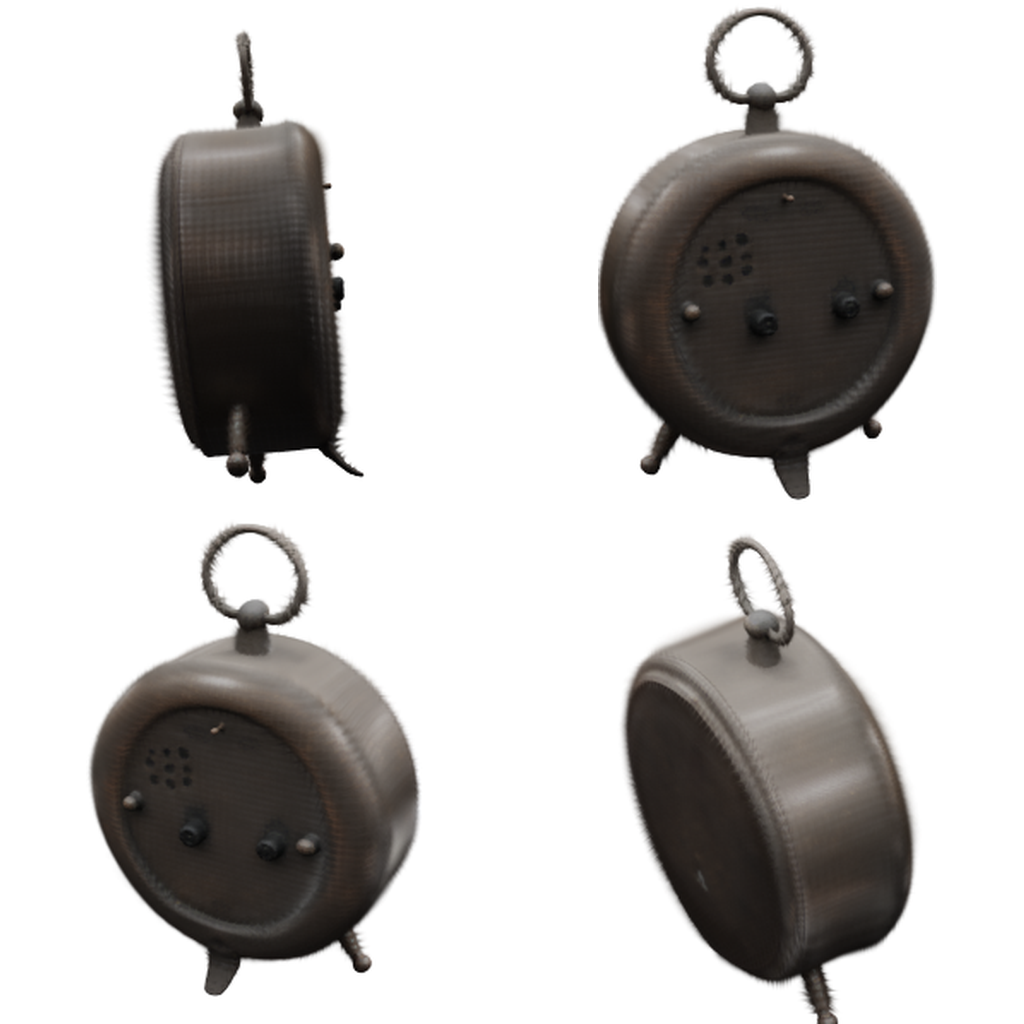} & \includegraphics[width=\compgimsize,height=\compgimsize]{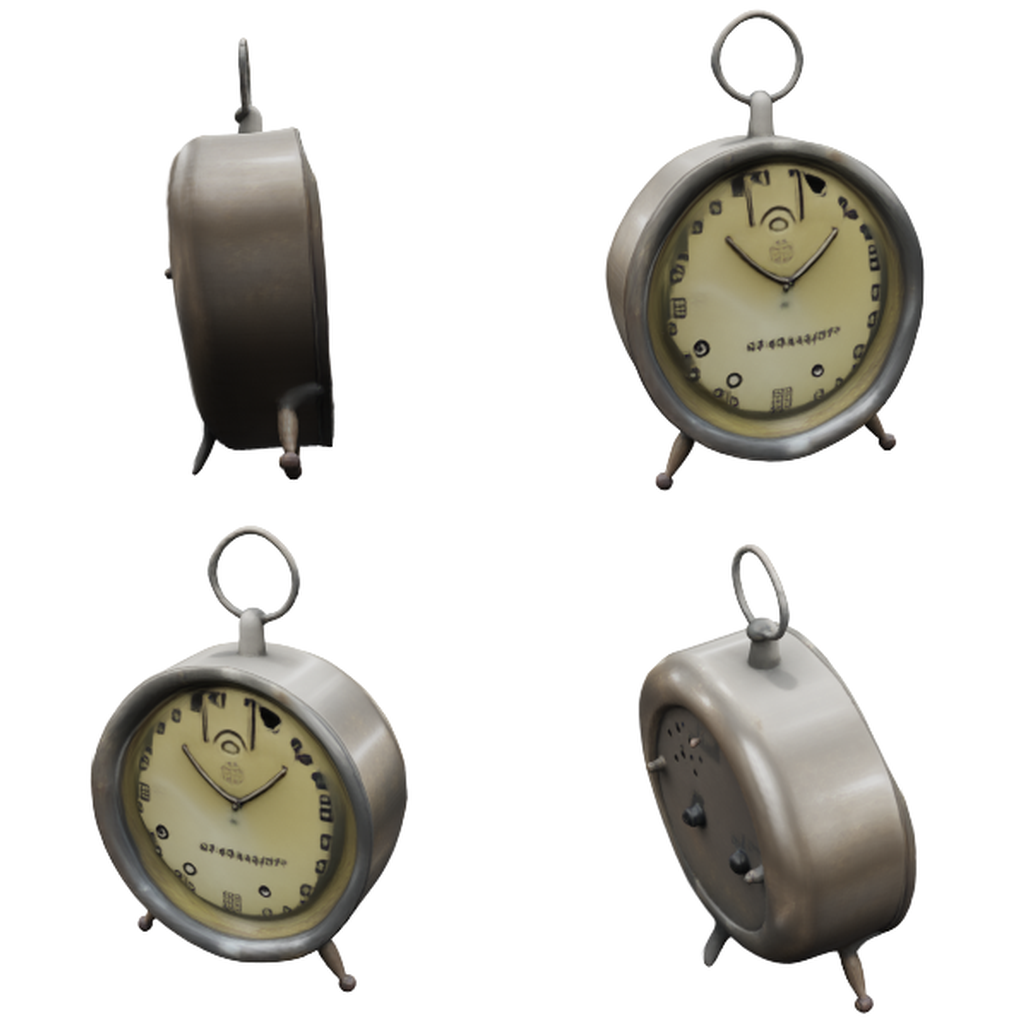} & \includegraphics[width=\compgimsize,height=\compgimsize]{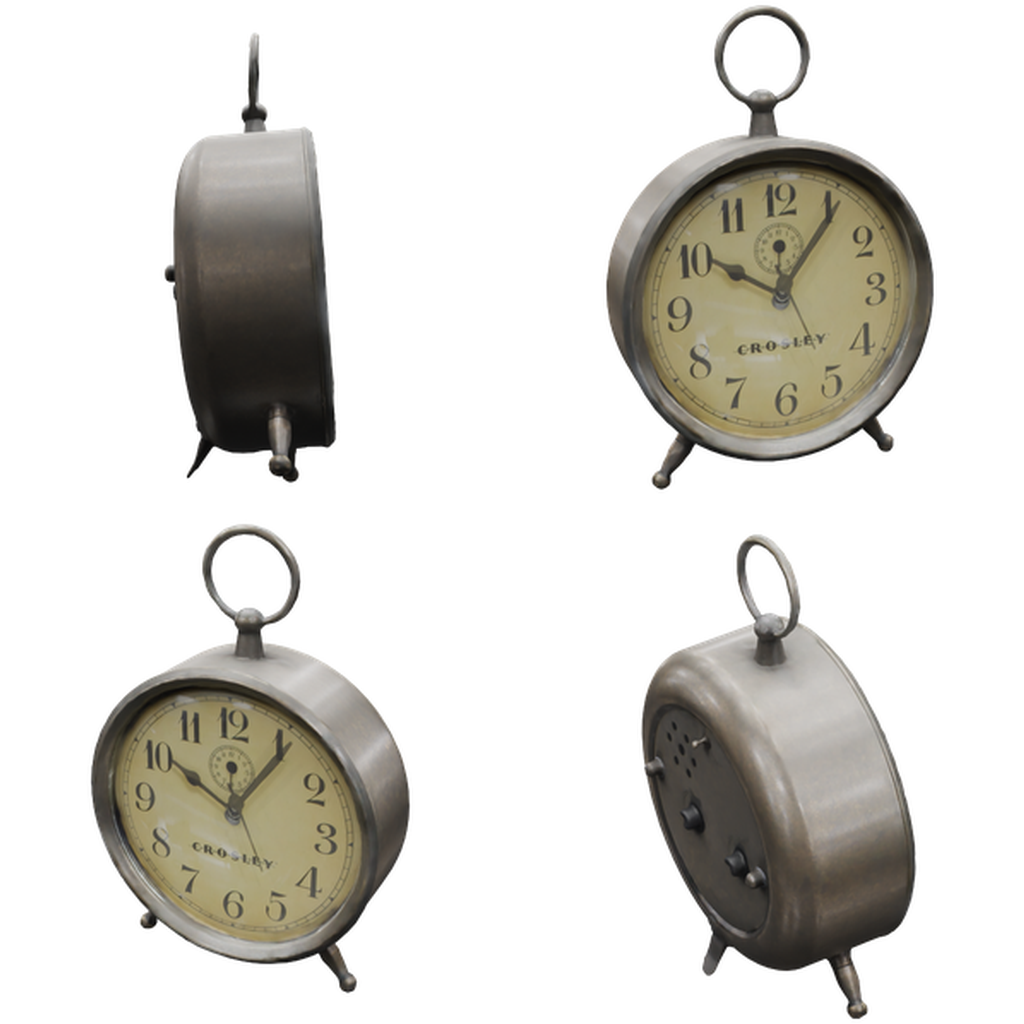} \\
\end{tabular}

\caption{
\textbf{Qualitative comparison with 3D generation methods} on Toys4k (top row) and GSO (bottom three). All non-input columns render the same held-out novel view in the GT camera frame after the similarity ICP alignment of Tab.~\ref{tab:main_gen}.
}
\label{fig:qualitative_gen}
\end{figure*}

\begin{table}[t]
\renewcommand{\arraystretch}{1.08}
\setlength{\tabcolsep}{4pt}
\centering
\caption{
\textbf{Camera-pose prediction accuracy.}
We report relative rotation error (RRE) with Acc.@$\{5^{\circ},15^{\circ}\}$ thresholds, relative translation error (RTE; the angular error between predicted and GT relative-translation directions), and mean absolute FOV error (RFOV; intrinsic-only); see supplementary for the full evaluation protocol. \ours{} reports two rows: known intrinsics (GT FOV fed in; RFOV not applicable, marked \xmark) and predicted intrinsics. Angular pose metrics use raw, gauge-invariant pairwise relative poses and the same ambiguity-tolerant score for all methods.
}
\label{tab:main_pose}
\resizebox{\linewidth}{!}{
\begin{tabular}{lcccccccccc}
\toprule
\multirow{2}{*}{\textbf{Method}} &
\multicolumn{5}{c}{\textbf{Toys4k}} &
\multicolumn{5}{c}{\textbf{GSO}} \\
\cmidrule(lr){2-6}\cmidrule(lr){7-11}
&
\textbf{RRE}$\downarrow$ & \textbf{Acc.@5°}$\uparrow$ & \textbf{Acc.@15°}$\uparrow$ & \textbf{RTE}$\downarrow$ & \textbf{RFOV}$\downarrow$ &
\textbf{RRE}$\downarrow$ & \textbf{Acc.@5°}$\uparrow$ & \textbf{Acc.@15°}$\uparrow$ & \textbf{RTE}$\downarrow$ & \textbf{RFOV}$\downarrow$ \\
\midrule
FreeSplatter~\cite{xu2024freesplatter} &
26.45 & 0.2438 & 0.4996 & 32.32 & 12.51 &
29.10 & 0.2327 & 0.4646 & 32.81 & 13.13 \\
ReconViaGen~\cite{chang2025reconviagen} &
16.54 & 0.2044 & 0.6881 & 19.56 & 9.54 &
14.48 & 0.2260 & 0.7343 & 16.16 & 8.95 \\
VGGT~\cite{wang2025vggt} &
10.29 & 0.5969 & 0.8406 & 18.61 & 37.66 &
9.33 & 0.6310 & 0.8702 & 17.69 & 36.08 \\
\rowcolor{tableRow}
\ours{} (predicted intrinsics) &
1.60 & 0.9723 & 0.9913 & 5.51 & 2.90 &
1.27 & 0.9821 & 0.9985 & 5.08 & 2.86 \\
\midrule
\ours{} (known intrinsics) &
1.44 & 0.9798 & 0.9910 & 2.38 & \xmark &
1.13 & 0.9892 & 0.9981 & 2.11 & \xmark \\
\bottomrule
\end{tabular}
}
\end{table}

\paragraph{Comparison with feed-forward reconstruction methods.}
Tab.~\ref{tab:main_recon} and Fig.~\ref{fig:qualitative_recon} compare against feed-forward reconstruction baselines under input-view pointmap LSQ alignment.
\ours{} outperforms every baseline on every metric and on both datasets, with the gap widening at the strict $\text{F@}.01$ threshold and qualitatively in the form of alignment artifacts and surface noise. Feed-forward methods compound any disagreement between their predicted geometry and predicted poses, while a generative 3D-native model refines it away.

\paragraph{Comparison with 3D generation methods.}
Tab.~\ref{tab:main_gen} and Fig.~\ref{fig:qualitative_gen} compare against 3D generation models that produce a canonical, pose-free output, under similarity-ICP alignment.
\ours{} leads on every metric, with the largest margins on the strict-threshold geometry metrics ($\text{F@}.01$, CD-med). Pose-free baselines hallucinate plausible but mismatched detail on unseen sides; the two strong multi-view baselines (\trellis-M's stochastic per-step view conditioning and ReconViaGen's feed-forward reconstruction with generative refinement) still trail \ours{} by large margins, indicating the gain comes from joint shape--pose denoising rather than late-stage view sampling.

\paragraph{Pose prediction accuracy.}
Tab.~\ref{tab:main_pose} reports camera-pose accuracy on baselines that predict input-view extrinsics. \ours{} achieves substantially lower rotation and translation errors, with Acc.@$5^{\circ}$ saturating; baselines commonly fail on symmetric objects whose orientation depends on low-level visual cues. Our model also recovers intrinsics accurately (known-intrinsics row shown for reference); VGGT, not trained on varying-FOV synthetic data, incurs large RFOV despite strong extrinsics, whereas \ours{} keeps RFOV under $3^{\circ}$ on both datasets while saturating Acc.@$5^{\circ}$. Thus, RFOV confirms reliable intrinsic recovery under varying FOV, but its margin over VGGT also reflects training-distribution coverage, not architecture alone.

\subsection{Ablation Study}
\label{sec:exp_ablation}

\begin{table}[t]
\renewcommand{\arraystretch}{1.08}
\setlength{\tabcolsep}{6pt}
\centering
\caption{
\textbf{Ablation study on Toys4k.}
Rows add components cumulatively, ending with the full model. Stage~1 uses a single-view-trained fine decoder extended to multi-view at inference; Stage~2 replaces it with a natively multi-view-trained decoder. All rows use the input-view pointmap alignment from Table~\ref{tab:main_recon}; RRE and Acc.@$5^{\circ}$ use a similarity fit from the final pointmap, so the fine decoder can shift them without changing raw coarse cameras. CD is the median across objects.
}
\label{tab:ablation}
\resizebox{\linewidth}{!}{
\begin{tabular}{llccccccc}
\toprule
\textbf{Stage} & \textbf{Variant} &
\textbf{PSNR}$\uparrow$ & \textbf{LPIPS}$\downarrow$ & \textbf{CD-med}$\downarrow$ & \textbf{F@.01}$\uparrow$ & \textbf{F@.05}$\uparrow$ & \textbf{RRE}$\downarrow$ & \textbf{Acc.@5°}$\uparrow$ \\
\midrule
\multirow{4}{*}{Stage 1} & Base (per-view cross-attention)             & 27.00 & 0.0383 & 0.2811 & 0.6078 & 0.9677 & 5.62 & 0.8653 \\
                        & ~~~~+ full cross-attention                  & 27.49 & 0.0359 & 0.2568 & 0.6324 & 0.9748 & 5.20 & 0.9069 \\
                        & ~~~~+ visual conditioning                   & 27.65 & 0.0353 & 0.2488 & 0.6346 & 0.9783 & 4.59 & 0.9228 \\
                        & ~~~~+ register tokens                       & 28.23 & 0.0327 & 0.2204 & 0.6670 & 0.9824 & 3.55 & 0.9478 \\
\midrule
\rowcolor{tableRow}
Stage 2 & ~~~~+ native multi-view training (\textbf{Full model})       & 28.61 & 0.0308 & 0.2184 & 0.6701 & 0.9823 & 3.65 & 0.9416 \\
\bottomrule
\end{tabular}
}
\end{table}


\newcommand{\ablqualcols}{%
  \makebox[0.16667\linewidth][c]{\footnotesize Base}%
  \makebox[0.16667\linewidth][c]{\footnotesize +cross}%
  \makebox[0.16667\linewidth][c]{\footnotesize +visual}%
  \makebox[0.16667\linewidth][c]{\footnotesize +reg}%
  \makebox[0.16667\linewidth][c]{\footnotesize \textbf{Full}}%
  \makebox[0.16667\linewidth][c]{\footnotesize GT}%
}

\begin{figure}[!t]
\centering
\begin{minipage}{0.9\linewidth}
\centering
\ablqualcols\par
\includegraphics[width=\linewidth]{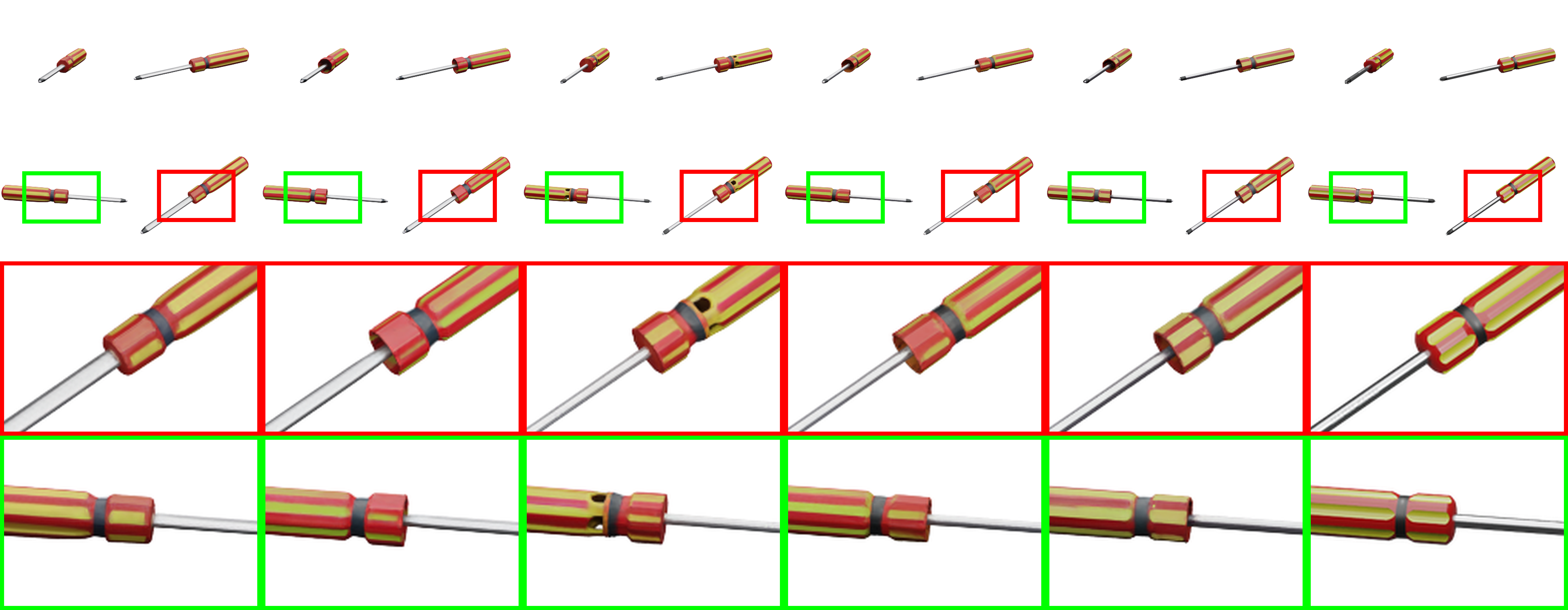}
\end{minipage}
\par\medskip
\begin{minipage}{0.9\linewidth}
\centering
\ablqualcols\par
\includegraphics[width=\linewidth]{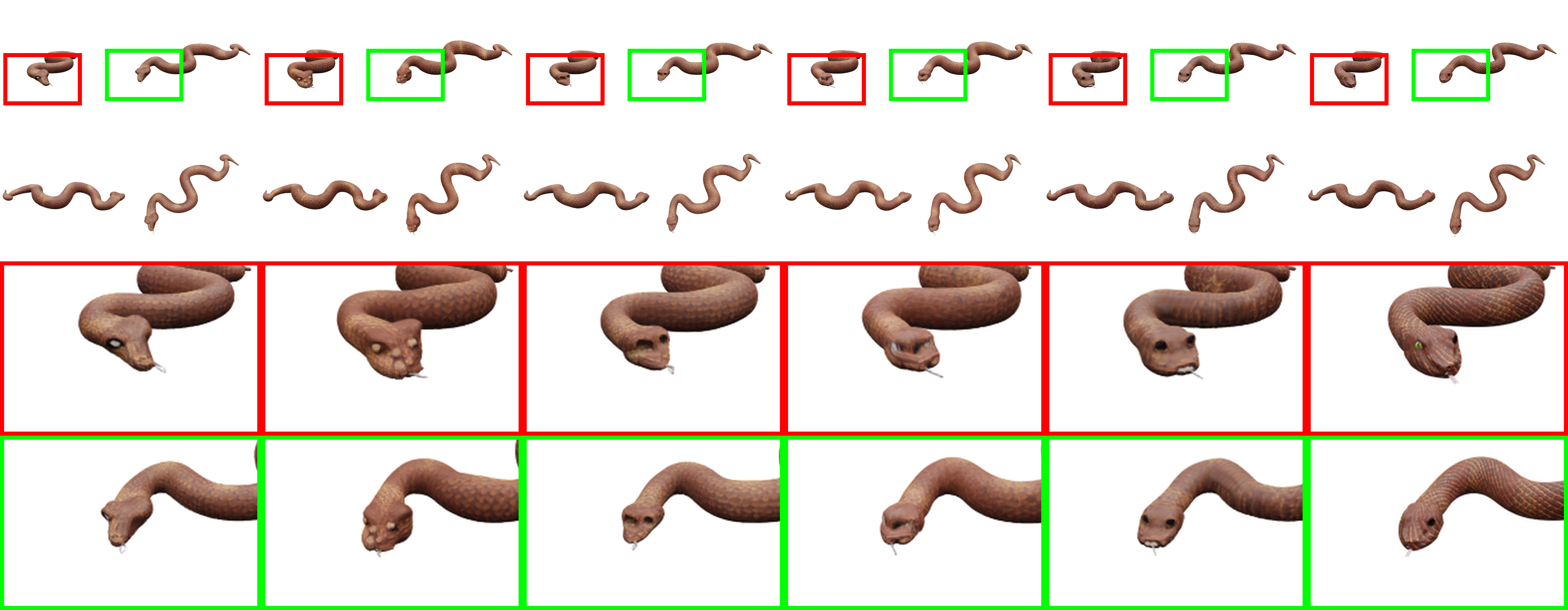}
\end{minipage}
\caption{\textbf{Qualitative ablation on Toys4k.}
For each instance we show the four held-out test views as a $2{\times}2$ grid (top), with two zoomed-in regions of largest cross-variant disagreement below (red / green borders).
Columns left-to-right correspond to the cumulative ablation rows of Table~\ref{tab:ablation}: Base, $+$cross, $+$visual, $+$reg, the full model, and ground truth.
Each added component sharpens local geometry and texture, with the full model best matching GT.
}
\label{fig:abl_qualitative}
\vspace{-4pt}
\end{figure}

We ablate each design choice on Toys4k in Tab.~\ref{tab:ablation}, with rows added cumulatively. In Stage~1, full cross-attention enables cross-view information sharing and improves all metrics; visual feature conditioning sharpens photometric fidelity and pose accuracy with only marginal geometry gains, suggesting appearance cues primarily help rendering and pose grounding; shared register tokens aggregate global information and contribute most to pose accuracy and the strict $\text{F@}.01$ threshold. In Stage~2, a fine decoder trained natively on multi-view data sharpens photometric fidelity, while 3D geometry is already saturated. The decoder swap leaves raw coarse-stage cameras unchanged but shifts geometry-aligned pose metrics because their similarity fit depends on the final pointmap.
\Figref{fig:abl_qualitative} corroborates these trends qualitatively: surface details (e.g., the screwdriver cap, the snake's head) sharpen left-to-right and the full model best matches GT.

\section{Conclusion}
\label{sec:conclusion}

We presented \ours, which recasts a native 3D diffusion model as the primary inference mechanism for sparse-view object reconstruction, jointly denoising shape and per-view camera poses in a single multi-modal diffusion transformer. Three choices keep the formulation tractable: a UV-volume pose latent compressed to a compact token grid, a unified multi-modal sequence that places shape and per-view pose tokens in a single self-attention, and a pose-aware fine stage that conditions each voxel on features sampled at its pose-aligned image projection. On Toys4k and GSO, \ours outperforms feed-forward, 3D-generation, and pose-prediction baselines on photometric, geometric, and pose metrics simultaneously, recovering both extrinsics and intrinsics to high accuracy without any external pose estimator or post-hoc optimization.

\paragraph{Limitations and future work.}
Our results are restricted to object-centric inputs at $\Nview{=}4$; extending to larger view counts, scene-level inputs, and in-the-wild captures with looser segmentation is a natural next step. Under distribution shift, large pose/calibration errors may misregister fine-stage image features, degrading local refinement while the coarse 3D prior preserves global completion. The model also inherits the object distribution and resolution limits of its \trellis backbone, so scaling the underlying generator should directly improve sparse-view recovery.

\begin{ack}
This work is partially supported by funding and computing resources provided by the HKU-Shanghai Intelligent Computing Research Center. This work was supported by the General Research Fund of the Research Grants Council of Hong Kong under Grant No.~17200725 and the JC STEM Lab funded by The Hong Kong Jockey Club Charities Trust.
\end{ack}


\medskip
{
    \small
    \bibliographystyle{unsrt}
    \bibliography{neurips_2026}
}


\appendix
\section{Technical appendices and supplementary material}
\label{sec:suppl}


\subsection{Alignment protocol}
\label{sec:suppl_alignment}

Our baselines differ in what they natively output (input-view pointmaps, canonical pose-free 3D, or input-camera poses), so a single shared alignment would unfairly discard the structure each method already provides. We therefore align each prediction with the protocol best matched to its output family before evaluating, and \ours{} is evaluated under each applicable family using that family's protocol so all comparisons within a table are like-for-like.

\paragraph{Input-view pointmap least squares (Table~\ref{tab:main_recon}).}
This protocol applies to feed-forward reconstruction methods (LucidFusion, FreeSplatter, ReconViaGen), which already place geometry in a noisy estimate of the input-camera frame. For each input view, we back-project the per-pixel predicted depth at the predicted camera into 3D, alpha-mask the result, randomly subsample to at most 50K points per view, and pair it with the GT pointmap obtained by back-projecting GT depth at the GT camera. Stacking the per-view pairs with equal aggregate weight per view yields one set of 3D-to-3D correspondences per object, which we fit with a closed-form Umeyama similarity~\citep{umeyama2002least}. The resulting transform is applied once to the full predicted 3D before all photometric and geometric evaluation; we log the per-object LSQ residual as an alignment-fidelity diagnostic. Note that this protocol exercises both halves of feed-forward reconstruction: the matched correspondences require the predicted geometry to align with the predicted cameras, so a method with weak geometry, weak pose recovery, or any mismatch between the two will produce noisy correspondences and a loose Umeyama fit, inflating all downstream metrics.

\paragraph{Similarity ICP (Table~\ref{tab:main_gen}).}
This protocol applies to 3D generation methods (TRELLIS-S, TRELLIS-M, SAM 3D), which output a canonical pose-free 3D, and to \ours{} when evaluated under the same family. To obtain the prediction-side point cloud uniformly across output formats (whether the prediction is a mesh or a set of 3D Gaussians), we render depth from a set of orbiting random views around the prediction and back-project the depth maps into 3D, which yields a well-covered surface point cloud without surface-sampling biases specific to the underlying representation. The GT-side cloud is sampled directly on the GT mesh surface. We initialize alignment by sweeping the 24 orientation-preserving signed axis permutations of the PCA-aligned bounding box and keep the candidate with the lowest post-ICP residual; we then refine with similarity ICP and use the resulting transform to map the prediction into the GT frame for evaluation.

\subsection{Pose metric protocol}
\label{sec:suppl_pose_metrics}

\paragraph{Pose metric definitions.}
Different pose-predicting baselines fix their world frame in different ways (some anchor to the first input view, some to an internal canonical frame, others through their own internal similarity fit), so absolute predicted extrinsics cannot be directly compared to GT extrinsics. To bypass this gauge ambiguity entirely, all metrics in Table~\ref{tab:main_pose} are computed on \emph{unordered pairs} of input views: for each pair, the relative pose between the two views cancels any global rotation, translation, and scale that affects all views jointly, leaving only the camera geometry intrinsic to the pair. We adopt this pairwise form uniformly for both rotation and translation rather than anchoring to one reference view, which would entangle every error with the chosen anchor. For each object, we average each metric over the $\Nview(\Nview{-}1)/2$ unordered pairs and then average across objects. \textbf{Relative Rotation Error (RRE)} is the geodesic angle (in degrees) between the predicted and GT relative rotations of the pair; \textbf{Acc.@$5^{\circ}$ / Acc.@$15^{\circ}$} is the fraction of pairs whose RRE falls below the corresponding threshold; \textbf{Relative Translation Error (RTE)} is the angular error (in degrees) between the predicted and GT relative-translation directions, an angular form chosen because absolute translation magnitude is itself gauge-dependent (a global similarity scale can be reabsorbed by any consistent rescaling of all camera centers); and \textbf{Field-of-View Error (RFOV)} is the mean absolute difference (in degrees) between predicted and GT FOV, computed directly on intrinsics without any pose alignment, since intrinsics are independent of the world frame. To score baselines with a coarse front/back chirality ambiguity on geometric correctness rather than chirality alone, all angular metrics (RRE, Acc.@$\{5^{\circ},15^{\circ}\}$, RTE) are read off after collapsing $R_{\text{pred}} R_{\text{gt}}^\top$ under the $\mathrm{diag}(\pm 1, \pm 1, -1)$ sign-flip equivalence class. The same rule is applied uniformly to all methods, including ours, to compare geometric recovery under a shared front/back equivalence. Chirality remains relevant for camera-dependent applications, so we explicitly note that this metric does not assess that failure mode.

\subsection{Pose VAE}
\label{sec:suppl_pose_vae}

\paragraph{Architecture and training.}
The pose VAE is a 3D-convolutional encoder--decoder over a $64^3$ voxel grid with a deliberately aggressive bottleneck: four $2{\times}$ spatial downsamples compress the input into a $4^3$ token grid with $8$ channels per token. This compression is feasible because each per-voxel target is a smooth function of voxel position and a low-DoF camera, so the latent does not need to absorb high-frequency content. Per voxel the encoder receives three input channels (a binary in-frustum indicator plus the two pixel coordinates obtained by projecting the voxel center under the input camera), and the decoder emits the same three channels through a classification head (visibility mask) and a regression head (pixel UV).

We use a dice loss on the visibility classifier and an L1 loss on the UV regression, with a small KL weight ($10^{-3}$) so the latent stays close to a standard Gaussian for the downstream diffusion stage. The training distribution matches inference: each minibatch sample takes a real camera pose from the multi-view renders of a sampled training object and projects it on the fly into the $64^3$ voxel grid. We optimize for $1$M steps with AdamW (lr $3{\times}10^{-4}$, weight decay $10^{-3}$) on a cosine schedule.

\paragraph{Out-of-frame voxels.}
Voxel centers can project outside the input image. We handle this by clamping target UVs to $[0,1]^2$ rather than dropping out-of-frame voxels, keeping the decoder output well-defined everywhere. The visibility channel encodes the same in-/out-of-frame information as a binary indicator with a small $1\%$ inset margin ($0.01 \le u,v \le 0.99$); the inference DLT solve uses this channel to restrict to in-frustum voxels, so saturated boundary UVs never contaminate the recovered pose. The UV reconstruction loss is applied uniformly to all voxels rather than only the in-frustum subset, keeping saturated-boundary regions consistent between encoder and decoder.

\paragraph{Pose autoencoding accuracy.}
\begin{table}[t]
\renewcommand{\arraystretch}{1.08}
\setlength{\tabcolsep}{4pt}
\centering
\caption{
\textbf{Pose VAE autoencoding accuracy on Toys4k across latent spatial resolutions.}
We sweep the latent grid in $\{4^3, 8^3, 16^3\}$ at a fixed $8$ channels per token (yielding $\{64,\,512,\,4096\}$ latent tokens respectively), encode each held-out $64^3$ ground-truth UV volume, decode, and recover camera pose via the DLT solve. Each cell reports \emph{median\,/\,mean\,/\,$95$\% quantile} over $\sim\!77$K (image, pose) pairs. UV distance is in normalized image coordinates ($\times 10^{-4}$); pose errors are in degrees.
}
\label{tab:pose_vae}
{\footnotesize
\begin{tabular}{@{}ccccc@{}}
\toprule
\textbf{Latent} & \textbf{UV Dist.}$\downarrow$ ($\times 10^{-4}$) & \textbf{RRE}$\downarrow$ & \textbf{RTE}$\downarrow$ & \textbf{RFOV}$\downarrow$ \\
\midrule
$4^3$  & 4.61\,/\,5.48\,/\,10.12 & 0.217\,/\,0.388\,/\,0.987 & 0.209\,/\,0.380\,/\,0.986 & 0.077\,/\,0.136\,/\,0.264 \\
$8^3$  & 7.06\,/\,7.85\,/\,12.41 & 0.202\,/\,0.373\,/\,0.919 & 0.200\,/\,0.366\,/\,0.924 & 0.063\,/\,0.121\,/\,0.234 \\
$16^3$ & 4.58\,/\,4.88\,/\,6.78  & 0.241\,/\,0.402\,/\,1.031 & 0.235\,/\,0.393\,/\,1.028 & 0.090\,/\,0.147\,/\,0.282 \\
\bottomrule
\end{tabular}}
\end{table}

Table~\ref{tab:pose_vae} reports the autoencoding fidelity of the trained pose VAE on the Toys4k held-out split, sweeping the latent spatial resolution. We encode each held-out $64^3$ ground-truth UV volume to the chosen latent grid, decode back, and recover camera pose via a DLT solve over in-frustum voxels. \textbf{UV Dist.}\ is the per-voxel L2 distance in normalized image coordinates; \textbf{RRE}, \textbf{RTE}, and \textbf{RFOV} are the standard rotation, translation-direction, and FOV errors in degrees.
All three latent resolutions deliver near-identical pose accuracy (within a few hundredths of a degree on RRE/RTE/RFOV), and even the worst row stays an order of magnitude below the downstream pose-prediction residuals in Table~\ref{tab:main_pose}, so the latent is not the bottleneck on overall pose accuracy. We adopt the $4^3$ latent in the main paper as the most compute-efficient operating point, since the larger grids carry $8\times$ and $64\times$ more tokens at no measurable pose gain.

\subsection{Image conditioning across stages}
\label{sec:suppl_image_cond}

\paragraph{Coarse stage: dual-stream encoder and token fusion.}
Each view is encoded by two complementary streams that fuse into a single per-view token sequence for the coarse DiT to cross-attend to. A frozen DINOv2-L/14 with registers at $518{\times}518$ produces $1374$ tokens per view (CLS + 4 registers + a $37{\times}37$ patch grid) at $1024$ channels, supplying object-level semantics. In parallel, a lightweight 4-level CNN at $296{\times}296$ ($5{\times}5$ convolutions, strides $\{1,2,2,2\}$, output channels $[6,6,6,8]$) is patchified down to the same $37{\times}37$ grid, with per-token contributions $[384,96,24,8]$ channel-concatenating into a $512$-dim local-texture feature. We append the $512$ local channels onto the $1024$ semantic channels at every patch token (non-patch tokens zero-pad on the local side), giving a uniform per-token width of $1536$ and a cross-attention key/value of length $\Nview \times 1374$. The two streams are deliberately complementary: semantic anchors pose to image content, while local texture restores the pixel-faithful appearance cues the semantic encoder has discarded, letting cross-attention recover accurate cameras without polluting the joint self-attention with raw image tokens.

\paragraph{Fine stage: visual pyramid and per-voxel projection.}
The fine stage refines up to $L \le 32{,}768$ active voxels at $64^3$ under the cameras already recovered by the coarse stage. Cross-attending every voxel to thousands of image tokens per view is intractable, so we replace cross-attention with per-voxel pixel-aligned feature sampling: each voxel projects to $\pixcoord_{j,i}$ in view $i$ under the predicted camera, and visual features are bilinearly sampled at that pixel. The visual encoder is a ConvNeXt-style pyramid run once per view at $256{\times}256$: four stages, each one ConvNeXt block with a $7{\times}7$ depthwise kernel at internal width $32$, with a $2{\times}$ downsample between stages. Each stage exposes a $4$-channel tap (LayerNorm + $1{\times}1$ Conv); the four taps sampled at $\pixcoord_{j,i}$ and channel-concatenated yield a $16$-channel multi-scale texture feature $\condfeat^{\tex}_j$. We pair it with an $8$-channel semantic feature $\condfeat^{\sem}_j$: the coarse-stage DINOv2 patch tokens (reused, no extra DINOv2 pass) are bilinearly sampled at the same per-voxel projection and re-encoded through the frozen SLat encoder, placing them on the SLat target's latent manifold and giving a compact $24$-channel per-voxel conditioning vector. This keeps fine-stage cost proportional to active voxels rather than view tokens, while the multi-scale taps let each voxel combine coarse contour cues with fine local texture without inflating the per-voxel width.

\subsection{Implementation details}
\label{sec:suppl_impl}

\paragraph{Pose VAE training.}
The pose VAE of \Appref{sec:suppl_pose_vae} is trained for $1$M steps on $8$ H200 GPUs with batch size $4$ per GPU, using AdamW (lr $3{\times}10^{-4}$, weight decay $10^{-3}$) under a cosine schedule with $1000$ warmup steps decaying to $10^{-5}$, EMA $0.9999$, and FP16 mixed precision. The same recipe trains the $4^3$, $8^3$, and $16^3$ latent variants of Tab.~\ref{tab:pose_vae}.

\paragraph{Coarse-stage shape-pose DiT training.}
The coarse stage is fine-tuned for $1$M steps on $8$ H200 GPUs at global batch size $64$ ($8$ per GPU, split into two gradient-accumulation microbatches), initialized from a prior single-view coarse-stage checkpoint of ours (itself from the pretrained \trellis-image-large SS DiT). Each instance contributes $\Nview{=}4$ random condition views uniformly sampled from the $24$ rendered views of the object; each input image is crop-resized at a $1.2{\times}$ tightness ratio around the alpha-mask bounding box and bilinearly resampled to $518{\times}518$ for the frozen DINOv2 ViT-L/14-with-registers encoder. The DiT operates on $4096$ shape, $256$ pose, and $128$ shared register tokens, with cross-attention over $\Nview \times 1374 = 5496$ image-condition tokens augmented by the convolutional pyramid of \Eqref{eq:image_cond}. We use AdamW (lr $10^{-4}$, no weight decay), CFG dropout $0.1$, the same cosine schedule and EMA as above, and timesteps drawn from a logit-normal distribution with mean $1.0$ and std $1.0$ ($\sigma_{\min}{=}10^{-5}$). At each step we randomly choose between two timestep-coupling modes: with probability $0.8$, the shape stream and all pose streams share a single timestep; with probability $0.2$, each stream draws an independent timestep. The shared mode dominates so the model stays well-calibrated to the synchronized inference schedule, while the per-pose mode supplies enough timestep diversity that the per-modality AdaLN modulation actually learns to handle independent noise levels.

\paragraph{Fine-stage SLat DiT training.}
The fine stage is fine-tuned independently for $1$M steps on $4$ H200 GPUs at global batch size $32$ ($8$ per GPU), initialized from \trellis's pretrained SLat DiT checkpoint. It denoises up to $L{\le}32{,}768$ active voxels at $64^3$; ground-truth cameras and occupancy are used during training so the fine stage learns from clean projections. Each instance again contributes $\Nview{=}4$ random condition views, with per-voxel features pooled across views by mean ($16$-channel ConvNeXt-pyramid texture and $8$-channel SLat-encoded semantic) and the cross-attention condition the per-view mean of DINOv2 patches. We use AdamW (lr $10^{-4}$, weight decay $10^{-3}$), CFG dropout $0.1$, and the same cosine schedule and logit-normal timestep distribution as the coarse stage.

\paragraph{Inference.}
The coarse stage runs a $25$-step flow-Euler sampler over the joint shape-pose sequence at classifier-free guidance strength $7.5$, active on the noise interval $\timestep \in [0.5, 1.0]$ with a guidance time-rescale factor of $3.0$~\cite{esser2024scaling}. We then decode each per-view pose latent through the pose VAE and recover $\projmat_i^{*}$ by the DLT solve of \Eqref{eq:dlt}. The fine stage runs a $25$-step flow-Euler sampler with sequential multi-view fusion at CFG strength $3.0$ on the same $[0.5, 1.0]$ interval and rescale factor, conditioned on the recovered cameras and on the per-voxel pose-sampled features of \Eqref{eq:slat_pose_cond}.

\subsection{Broader impacts}
\label{sec:suppl_impacts}

\ours produces a 3D representation of an object from a few uncalibrated images, jointly recovering shape and camera. The most direct positive applications are in robotics (object-level scene understanding from a small number of viewpoints), augmented and virtual reality (capturing personal items as 3D assets for placement in virtual environments), and content creation (3D meshes from reference photos as a starting point for downstream editing). Joint recovery of camera intrinsics and extrinsics with shape removes the calibration friction that has limited deployment of feed-forward reconstruction methods in casually-captured settings.

The negative considerations are those shared with image-conditioned 3D generation more broadly: the resulting 3D models could be used to fabricate convincing 3D content of real-world objects, including objects belonging to others, without their consent; coupled with downstream rendering pipelines, this could enable image-only digital reproduction at a scale that traditional photogrammetry has not previously allowed. The model is trained only on common, non-sensitive object categories from public 3D-asset datasets, and recovery quality degrades on out-of-distribution categories, which provides a partial natural mitigation. Released model checkpoints will carry an attribution requirement and an acceptable-use note as the safeguard accompanying our open release.




\end{document}